%% file: main.tex
\documentclass[10pt,twocolumn,letterpaper]{article}

\usepackage{iccv}
\usepackage{times}
\usepackage{epsfig}
\usepackage{graphicx}
\usepackage{amsmath}
\usepackage{amssymb}
\usepackage{wrapfig}

\usepackage{color}

\usepackage{multirow}
\usepackage{multicol}
\usepackage{caption}
\usepackage{subcaption}

\usepackage[breaklinks=true,bookmarks=false]{hyperref}

\usepackage[accsupp]{axessibility}

\iccvfinalcopy 

\def\iccvPaperID{5545} 

\ificcvfinal\fi

\title{FloorPlanCAD: A Large-Scale CAD Drawing Dataset for Panoptic Symbol Spotting}

\author{Zhiwen Fan$^{1}$$\dagger$\hspace{1cm}Lingjie Zhu$^{1}$$\dagger$\hspace{1cm}Honghua Li$^{1}$\hspace{1cm}Xiaohao Chen$^{1}$\hspace{1cm}Siyu Zhu$^{1}$\hspace{1cm}Ping Tan$^{1,2}$\\
${}^{1}$Alibaba Groups\hspace{1.5cm}${}^{2}$Simon Fraser University\hspace{1.5cm}${}^{\dagger}$Equal contribution\\
}

\begin{document}

    \maketitle
    \ificcvfinal\thispagestyle{empty}\fi
    
    \input{abs}
    \input{intro}
    \input{related}
    \input{task}

\input{dataset}
    \input{method}
    \input{exp}

    \input{conclusion}

\input{supplementary/supplementary}

    {\small
        \bibliographystyle{ieee_fullname}
        \bibliography{reference}
    }

\end{document}

%% file: abs.tex
\begin{abstract}
Access to large and diverse computer-aided design (CAD) drawings is critical for developing symbol spotting algorithms.
%
In this paper, we present FloorPlanCAD, a large-scale real-world CAD drawing dataset containing over 15,000 floor plans, ranging from residential to commercial buildings.
CAD drawings in the dataset are all represented as vector graphics, which enable us to provide line-grained annotations of 35 object categories. 
%
%
Equipped by such annotations, we introduce the task of \textbf{panoptic symbol spotting}, which requires to spot not only instances of countable things, but also the semantic of uncountable stuff.
Aiming to solve this task, we propose a novel method by combining Graph Convolutional Networks (GCNs) with Convolutional Neural Networks (CNNs), which captures both non-Euclidean and Euclidean features and can be trained end-to-end. 
The proposed CNN-GCN method achieved state-of-the-art (SOTA) performance on the task of semantic symbol spotting, and help us build a baseline network for the panoptic symbol spotting task. 
%
%
Our contributions are three-fold: 1) to the best of our knowledge, the presented CAD drawing dataset is the first of its kind; 2) the panoptic symbol spotting task considers the spotting of both thing instances and stuff semantic as one recognition problem; and 3) we presented a baseline solution to the panoptic symbol spotting task based on a novel CNN-GCN method, which achieved SOTA performance on semantic symbol spotting. 
We believe that these contributions will boost research in related areas. The dataset and code is publicly available at \url{https://floorplancad.github.io/}.
\end{abstract}

%% file: intro.tex
\begin{figure}[t]
    \centering
    \begin{subfigure}[b]{0.47\linewidth}
        \centering
        \includegraphics[width=\linewidth]{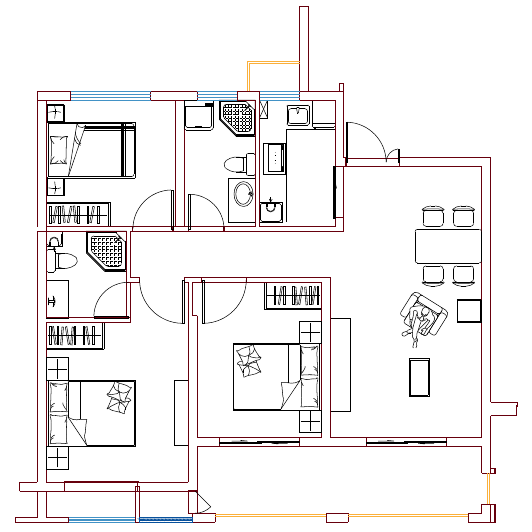}
        \caption{Floor plan}
    \end{subfigure}
    \begin{subfigure}[b]{0.485\linewidth}
        \centering
        \includegraphics[width=\linewidth]{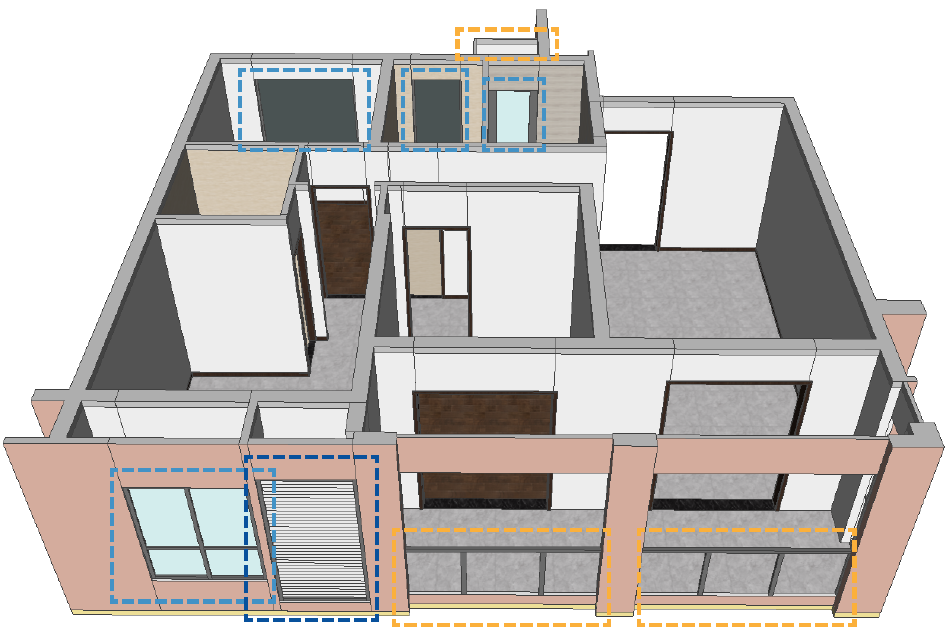}
        \includegraphics[width=\linewidth]{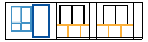}
        \caption{Facade and 3D model}
    \end{subfigure}
    \caption{The rich semantic, accurate location and detailed 3D shape (right top) of windows (light blue), blind windows (blue), railings (orange) and walls (dark red) are faithfully encoded in the CAD drawings of a floor plan (left) and its south facade (right bottom).}
    \label{fig:cad}
    \vspace{-2mm}
\end{figure}

\begin{figure}[ht]
    \centering
    \includegraphics[width=\linewidth]{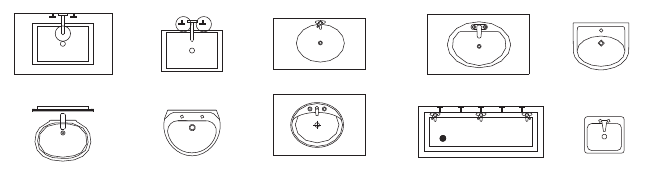}
    \caption{Various sink symbols from our FloorPlanCAD dataset. The style and appearance of a symbol depend on the producer of the drawing.}
    \label{fig:sinks}
    \vspace{-4mm}
\end{figure}

\begin{figure*}[ht!]
    \centering
    \begin{subfigure}[t]{0.45\textwidth}
        \centering
        \includegraphics[width=\textwidth]{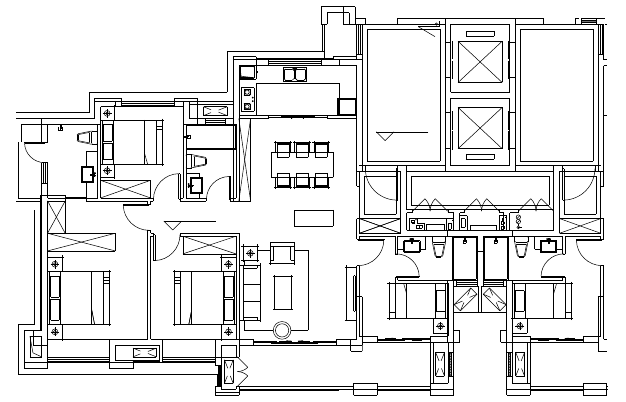}
        \caption{A raw floor plan drawing.}
        \label{fig:teaser-a}
    \end{subfigure}
    \begin{subfigure}[t]{0.525\textwidth}
        \centering
        \includegraphics[width=\textwidth]{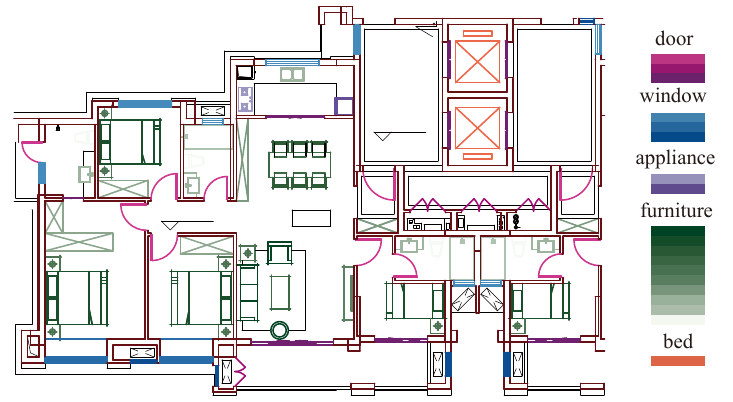}
        \caption{The drawing with instance and semantic annotations.}
        \label{fig:teaser-b}
    \end{subfigure}
    \caption{A snapshot of our FloorPlanCAD dataset. (a) Texts are removed to protect privacy and intellectual property. (b) The line-grained annotations is illustrated by colors.}
    \label{fig:teaser}
    \vspace{-4mm}
\end{figure*}

\section{Introduction}

The perception of 2D computer-aided design (CAD) drawings plays a crucial role for creating 3D prototypes, also known as ``digital twins", see Figure~\ref{fig:cad}, in architecture, engineering and construction (AEC) industries. 
A CAD drawing typically conveys accurate geometric and rich semantic information of a cross section of a 3D design. By integrating information from a group of CAD drawings, the according 3D model can be precisely reconstructed. 
For example, 3D buildings can be faithfully encoded by a bunch of 2D floor plans, which are detailed CAD drawings compose of line segments, arcs, curves, and texts, see Figure~\ref{fig:cad}. 
Automatic perception of CAD drawings will lead to efficient 3D modeling approaches, saving vast amount of labor work. That is especially true for architectures, which usually contain massive components and might cost months for creating detailed 3D models.

Symbol spotting refers to the recognition of graphic symbols embedded in the context of large digital drawings~\cite{rezvanifar2019symbol}. 
It is typically carried out as query-by-example approaches~\cite{nguyen2008symbol,nguyen2009symbol,rusinol2010symbol}, where candidate regions that might contain the given query symbol are obtained. These approaches are impractical in real-world scenarios, because symbols representing the same object might vary fiercely, see Figure~\ref{fig:sinks}. 
To cope with the variability of symbols, a recent work~\cite{rezvanifar2020symbol} attempted learning-based approach on real-world floor plan drawings, but they treat CAD drawings as pixel images, losing the accuracy of vector graphics and leading to possibly inaccurate annotations and predictions.

Traditional symbol spotting approaches focus on instance detection, therefore cannot deal with semantic of uncountable stuff. For example, these methods cannot detect the wall in CAD drawings, which is usually represented by a group of parallel lines with large span, see Figure~\ref{fig:teaser}. 
Following ideas in~\cite{kirillov2019panoptic}, we consider the instance spotting of countable things and the semantic detection of uncountable stuff as one visual recognition task, which is called \emph{panoptic symbol spotting}.

In practice, CAD drawings play the role of universal language among practitioners in AEC industries, including designers, engineers, constructors, who share a common knowledge set. This observation inspires us to adopt learning methods for recognition tasks on CAD drawings, which demands comprehensive annotated data for training and testing networks. 

We build a large-scale dataset of over 15,000 floor plans in the form of vector graphics.
These floor plans are collected from public repositories of real-world architecture projects across the industries.
To help researchers processing those drawings, we parse the initial proprietary ``.dwg'' file to the open standard ``.svg'' format and crop each floor plan into regular smaller blocks.
At the end, floor plan blocks in the dataset only contain geometric and structural information, as illustrated in Figure~\ref{fig:teaser-a}.
We select 35 object categories of our interest, and provide line-grained annotations, see Figure~\ref{fig:teaser-b}.

The property of vector graphics enable us to apply graph convolutional networks (GCNs), which is computational efficient due to its sparsity, and is good at extracting non-Euclidean features via topology connections. 
For each floor plan, we build a graph, whose nodes are graphic entities, e.g. straight segment, arc, and edges are created according to their adjacency. 

In our experiments, we found that Euclidean features captures by Convolutional Neural Networks (CNNs) can improve the performance. 
Therefore we propose a novel network combining GCNs and CNNs, which achieves state-of-the-art performance on the semantic symbol spotting task, and leads us to a baseline network for the panoptic symbol spotting task.

The goal of our research is to push the development of perception on the CAD drawing
by providing large-scale annotated dataset and a baseline algorithm. Our main contributions are:

\begin{itemize}
    \item We present a large-scale real-world dataset of over 15,000 CAD drawings with line-grained annotations, covering various types of buildings, e.g. residential towers, schools, hospitals, shopping malls and office buildings. To the best of our knowledge, it is the first of its kind.

    \item We introduce the task of \textit{panoptic symbol spotting}, which is a generalization of the traditional symbol spotting problem, considering the instance spotting of countable things and the semantic labeling of uncountable stuff as one recognition task. A panoptic metric is provided for evaluating the prediction quality of various methods.

    \item We propose a CNN-GCN method, which achieved state-of-the-art performance on the semantic symbol spotting task, and help us  build a unified baseline network for the panoptic symbol spotting task.
\end{itemize}


%% file: related.tex
\section{Related Work}

In this section we briefly summarize existing datasets and methods in related areas, including symbol spotting, sketch segmentation and panoptic segmentation.

\vspace{-4.5mm}
\paragraph{Existing datasets} The impact of a proper dataset for pushing the development in an area has been widely recognized.
For instance, ImageNet~\cite{russakovsky2015imagenet} for image recognition,
Matterport3D~\cite{Matterport3D} for RGB-D scene understanding
and ShapeNet~\cite{shapenet2015} for 3D shape perception.
SESYD~\cite{delalandre2010generation} is is a database of synthetical vectorized graphic documents
with corresponding ground truth including $1000$ floor plans.
FPLAN-POLY~\cite{rusinol2010relational} dataset contains $42$ floor plans converted from images
using a raster-to-vector algorithm~\cite{hilaire2006robust} implemented in the QGar~\cite{qgar} library.

\vspace{-4.5mm}
\paragraph{Symbol spotting}
Symbol spotting~\cite{rezvanifar2019symbol,rusinol2010symbol,santosh2018document} refers to the retrieval of graphical symbols embedded in larger images or documents~\cite{rezvanifar2020symbol}.
Symbol retrieval and recognition in technical documents remains a challenge in the document analysis community~\cite{nguyen2008symbol}.
Traditionally, hand-crafted symbol descriptors is designed to describe the shape~\cite{nguyen2008symbol,nguyen2009symbol,rusinol2010symbol}.
Then the query symbol and the document is matched with the sliding window or information retrieval techniques.
Graph representation and matching~\cite{dutta2011symbol,dutta2013near,dutta2013symbol}
is also used but more sensitive to the noise and topology error.
These methods work well on isolated symbols but fail significantly when symbols are embedded in documents.
Recently, an image based deep learning method is proposed~\cite{rezvanifar2020symbol} and achieves
the best result on existing public datasets~\cite{delalandre2010generation,rusinol2010relational}.

\vspace{-4.5mm}
\paragraph{Semantic sketch segmentation}
Semantic sketch segmentation aims to label pixels into semantic groups on freehand sketched line images~\cite{sun2012free,huang2014data,li2018fast,yang2020sketchgcn}.
Sun et al.~\cite{sun2012free} segment the cluttered sketch into multiple parts first,
and then detect semantically meaningful objects by leveraging a web-scale clipart database.
Huang et al.~\cite{huang2014data} formulate the problem as a mixed integer programming problem,
and present a data-driven solution.
By treating the input as a 2D point set and encode the stroke structure information into graph,
Yang et al.~\cite{yang2020sketchgcn} predict the per-point labels with a Graph Convolutional Networks (GCNs).

\vspace{-4.5mm}
\paragraph{Panoptic segmentation}
In computer vision, countable things are referred to as instances, such as doors, windows,
and tables~\cite{he2017mask,lin2017feature,lin2017focal}.
Uncountable stuff that extends in amorphous regions of similar texture or material has no instance and only semantic,
such as sky, road, and wall~\cite{chen2017rethinking,chen2018encoder,sandler2018mobilenetv2}.
Kirillov et al.~\cite{kirillov2019panoptic} take instance and semantic segmentation into
one visual recognition task and coin the phrase panoptic segmentation.
It tries to assign stuff pixels with semantic label and detect each object with a bounding box
or segmentation mask at the same time~\cite{kirillov2019panoptic,kirillov2019panoptic.2,xiong2019upsnet}.
BANet~\cite{chen2020banet} introduces a bidirectional path between semantic and instance segmentation to improve the panoptic performance.
The BGRNet~\cite{wu2020bidirectional} incorporates graph structure to mine the intra-modular and inter-modular relations.

%% file: task.tex
\label{sec:panoptic_symbol_spotting}
\section{Panoptic Symbol Spotting}
\label{sec:panoptic}

\begin{figure}
    \centering
    \begin{subfigure}[b]{0.48\linewidth}
        \centering
        \includegraphics[width=\linewidth]{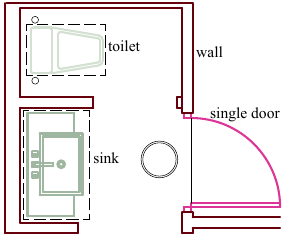}
        \caption{Ground truth}
    \end{subfigure}
    \begin{subfigure}[b]{0.48\linewidth}
        \centering
        \includegraphics[width=\linewidth]{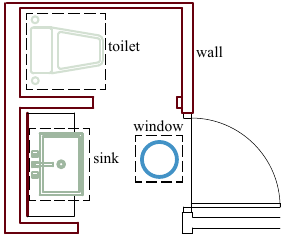}
        \caption{Prediction}
    \end{subfigure}
    \begin{subfigure}[b]{0.38\linewidth}
        \centering
        \includegraphics[width=0.48\linewidth]{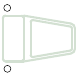}
        \includegraphics[width=0.48\linewidth]{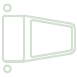}
        \caption{}
    \end{subfigure}
    \begin{subfigure}[b]{0.25\linewidth}
        \centering
        \includegraphics[width=0.48\linewidth]{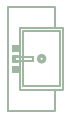}
        \includegraphics[width=0.48\linewidth]{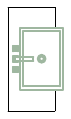}
        \caption{}
    \end{subfigure}
        \begin{subfigure}[b]{0.12\linewidth}
        \centering
        \includegraphics[width=\linewidth]{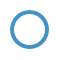}
        \caption{}
    \end{subfigure}
    \begin{subfigure}[b]{0.2\linewidth}
        \centering
        \includegraphics[width=\linewidth]{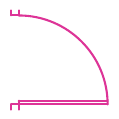}
        \caption{}
    \end{subfigure}
    \caption{In CAD drawings, a generalized symbol represents either an instance of things (e.g. sink, toilet, single door), or a particular stuff (e.g. wall).  
    Given the ground truth symbols (a), the predicted symbols (b) can be classified as true positive $TP$ (c-d) and false positive $FP$ (e), while missing ground truth symbols are considered as false negative $FN$ (f).
    %
    }
    \label{fig:metric}
    \vspace{-3mm}
\end{figure}

\begin{figure}
    \centering
    \includegraphics[width=0.9\linewidth]{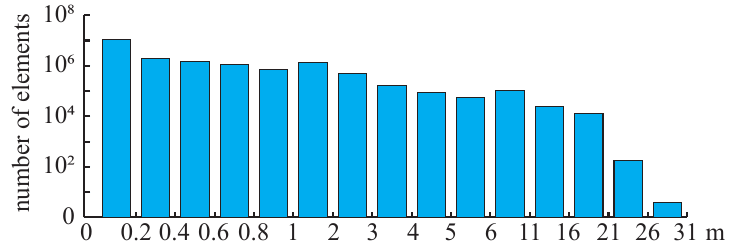}
    \caption{Length histogram of elements in the CAD drawing. Most of them are shorter than $0.2m$.}
    \label{fig:hist}
    \vspace{-3mm}
\end{figure}

Traditional symbol spotting techniques ~\cite{rusinol2010symbol,santosh2018document} mostly study \textit{things} -- countable symbols such as windows, doors, tables~\cite{rezvanifar2019symbol}.
\textit{Stuff} -- amorphous symbols of similar objects or material such as vegetation, wall, road -- is barely discussed.
Similar to the panoptic segmentation in~\cite{kirillov2019panoptic}, we propose and study a task named \textit{panoptic symbol spotting}, which considers semantic detection of stuff as well as instance spotting of things.

\vspace{-3mm}
\paragraph{Generalized symbol}
Symbols are graphical entities that hold a semantic meaning in a specific domain, such as logos, silhouettes and musical notes~\cite{rusinol2010symbol.2}.
In the setting of CAD drawings, a generalized symbol is a set of graphical entities, representing either a thing instance, e.g. toilet, sink, door, or a particular stuff, e.g. wall, see Figure~\ref{fig:metric}.
We denote a graphical entity $e_i=(l_i,z_i)$ by a semantic label $l_i$ and an instance index $z_i$, thus a symbol is defined as $s_j =\{e_i \mid l_i=l_j,z_i=z_j \}$. For short, we demote a symbol as $s_j=(l_j, z_j)$.
Things of the same class share the same semantic label, while instance of things can be distinguished by their instance indices. Notably, we ignore the instance index of entities belonging to stuff.
The selection of which classes are stuff or things is a design choice left to the creator of the dataset.

\vspace{-4mm}
\label{task.task_formation}
\paragraph{Task formation}
Given a CAD drawing represented by a set of graphical entities $\{e_k\}$, the \textit{panoptic symbol spotting} task requires a map $F_p: e_k\mapsto (l_k, z_k) \in\mathcal{L}\times\mathbf{N}$, where $\mathcal{L}:=\{0,\ldots,L-1\}$ is a set of predetermined set of object classes, and $\mathbf{N}$ is the number of possible instances.
The semantic label set $\mathcal{L}$ can be partitioned into stuff and things subsets, namely $\mathcal{L}=\mathcal{L}^{st}\cup\mathcal{L}^{th}$ and $\mathcal{L}^{st}\cap\mathcal{L}^{th}=\emptyset$.
By ignoring the instance indices, the task is degraded to a \textit{semantic symbol spotting} task $F_s: e_k\mapsto l_k \in\mathcal{L}$.
If we only focus on the thing classes $\mathcal{L}^{th}$, the panoptic symbol spotting is degraded to an \textit{instance symbol spotting} task $F_i: e_k\mapsto (l_k, z_k) \in\mathcal{L}^{th}\times\mathbf{N}$. 


\vspace{-4mm}
\paragraph{Panoptic metric} 
A predicted symbol $s_p=(l_p, z_p)$ is matched to a ground truth symbol $s_g=(l_g, z_g)$ if $l_p = l_g$ and $\text{IoU}(s_p, s_g)>0.5$, where the intersection over union (IoU) between two symbols are computed based on arc length $L(\cdot)$,
\begin{equation}\label{eq:iou}
    \text{IoU}(s_p, s_g) = \frac{\Sigma_{e_i \in s_p \cap s_g } log(1 + L(e_i)) }
    {\Sigma_{e_j \in s_p \cup s_g } log(1 + L(e_j))}.
\end{equation}
Here logarithm is adopted to degrade the influence of lines with very large span.
Figure~\ref{fig:hist} shows the distribution of entity length, which spans from several millimeters to tens of meters.
%
As proved in~\cite{kirillov2019panoptic}, this strategy produces a unique mapping: there can be at most one predicted symbol matched with each ground truth symbol.
A predicted symbol is considered as true positive ($TP$) if it can be matched to a ground truth symbol, otherwise false positives ($FP$). Missing ground truth symbols are marked as false negatives ($FN$).


The widely used $F_1$ score is used to measure the recognition quality:
\begin{equation}
RQ=\frac{\vert TP\vert}{\vert TP\vert+\frac{1}{2}\vert FP\vert+\frac{1}{2}\vert FN\vert}.
\end{equation}
By averaging the IoUs of matched symbols, the segmentation quality is measured by:

\begin{equation}
SQ=\frac{\sum_{(s^p,s^g)\in TP} \text{IoU}(s^p,s^g)}{\vert TP\vert}.
\end{equation}
Similar to~\cite{kirillov2019panoptic}, our panoptic symbol spotting metric is defined as the multiplication of $RQ$ and $SQ$:

\begin{align}
\nonumber PQ &=RQ\times SQ \\
&=\frac{\sum_{(s^p,s^g)\in TP} \text{IoU}(s^p,s^g)}{\vert TP\vert+\frac{1}{2}\vert FP\vert+\frac{1}{2}\vert FN\vert}.
\label{eq:metric}
\end{align}

Notably, this panoptic metric takes both thing and stuff symbols into account, proving a uniform quality measurement for evaluating panoptic symbol spotting methods. 


%% file: dataset.tex
\begin{figure}
    \centering
    \includegraphics[width=.9\linewidth]{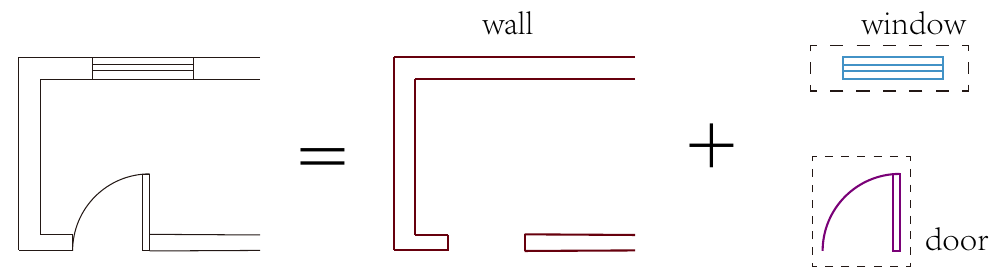}
    \caption{Illustration of layers in CAD drawings.}
    \label{fig:layers}
    \vspace{-4mm}
\end{figure}

\begin{figure*}[t]
    \centering
    \includegraphics[width=\textwidth]{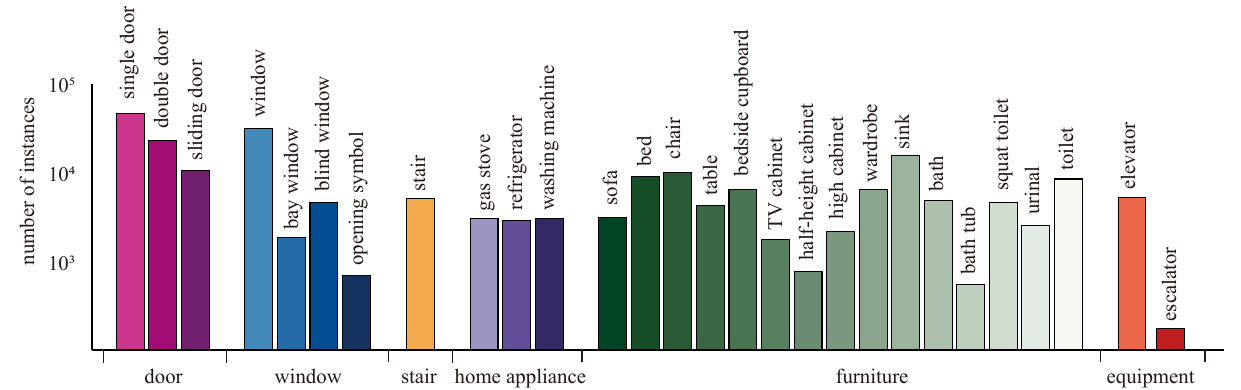}
    \caption{Number of finely annotated instances (y-axis) for 28 \textit{thing} classes and their associated categories (x-axis).}
    \label{fig:catalogue}
    \vspace{-4mm}
\end{figure*}

\begin{table*}[!t]
\begin{center}
\footnotesize
\begin{tabular}{c | c | c c | c c | c c c}
\hline
\multirow{2}{*}{Dataset} &\multirow{2}{*}{source} &\multicolumn{2}{c|}{scale} &\multicolumn{2}{c|}{image} &\multicolumn{3}{c}{annotation} \\
\cline{3-9}
& &\#classes &\#size &raster &vector &instance &semantic &vector \\
\hline
SESYD~\cite{delalandre2010generation}   &synthetic    &16   &1000   &\checkmark &\checkmark &\checkmark &           & \\
FPLAN-POLY~\cite{rusinol2010relational} &internet     &38   &48     &\checkmark &           &\checkmark &           & \\
BRIDGE~\cite{goyal2019bridge}           &internet     &-    &13000  &           &           &\checkmark &           & \\
FloorPlanCAD                            &industry     &35   &15663  &\checkmark &\checkmark &\checkmark &\checkmark &\checkmark \\
\hline
\end{tabular}
\end{center}

\vspace{-4mm}
\caption{Properties and statistics of existing datasets. BRIDGE~\cite{goyal2019bridge} still is not publically available by the time this paper is submitted.
Note that our FloorPlanCAD is the only one that gives panoptic annotation while retaining the characteristics of vector graphics throughout the process.}
\label{tab:datasets}
\vspace{-4mm}
\end{table*}



\section{The FloorPlanCAD dataset}

We collect large-scale CAD drawings from public repositories of real-world architecture projects across the industries.
Since the data is from different sources, it shows much more varieties in style and appearance of the objects.
To the date of this revision, we have obtained over 100 projects including residential buildings, schools, hospitals, and large shopping malls with complicated structures.

\subsection{Preproccess}
\label{subsec.data.preprocess}
In practice, architects tend to organize floors and related components of several buildings into one file, and a project usually has more than one file.
Before getting started, we need to clip out every floor plan from the project files.
The number of floor plans in a project is usually between 10 to 50, and now we have $2,500$ files of individual floor plans.


Generally, a floor plan drawing consists of dozens of layers distinguished by the functionality of the elements.
The layer name does not necessarily explains its content, and the layer content might be quite messy because there is no restrictions on what should be grouped as a layer.
The original multi-layer floor plan is firstly split into separate layers as shown in the Figure~\ref{fig:layers}.
The following annotation is performed on each layer that is much less cluttered, and boosts the annotation efficiency and effect significantly.

\subsection{Annotation}
We choose 35 object classes of our interest including 30 thing classes in this revision.
Some of them are listed in Figure~\ref{fig:catalogue}, please visit our website for more details.
Two stuff classes, \textit{wall} and \textit{parking}, are included because they are the very fundamental and dominant elements in floor plans.
The \textit{wall} serves as the basic structure for windows, doors, beams, etc. The \textit{wall} and \textit{parking} entities together accounts for about $27\%$ of the total entities (Table 6 in supplementary material), therefore is sufficient for studying the symbol spotting problem and solving the related task.
%
%
We make \textit{parking} as a semantic class, even though each parking slot can be treated as an instance.
However they always appear side by side and span a huge area, it is reasonable to view them as an a whole parking space to reduce the burden in annotation.

11 specialists spend over 1,000 hours on creating line-grained annotations.
To protect the privacy of the data owners, data obfuscation is conducted.
First of all, fields that are classified as identifiable, personally or commercially sensitive are removed.
Each floor plan is cut into squared blocks with dimension of $10m \times 10m$ and only $30\%$ of the blocks are kept in our dataset.

\subsection{Properties}

There are three key characteristics that make our dataset unique and valuable: \textit{large-scale}, \textit{real-world}, and \textit{vector graphics}.
It implies that studies on our dataset are more suitable to practical applications.
Table~\ref{tab:datasets} shows that we have much richer categories and larger amount of drawings comparing to SESYD~\cite{delalandre2010generation} and FPLAN-POLY~\cite{rusinol2010relational}.

The annotated dataset is split into two sets: 60 projects are randomly chosen for training and the remaining for testing.
Then our dataset consists of $10,161$ training and $5,502$ testing drawings with line-grained annotations.
In the reported experiments, $800$ random CAD drawings are split from the training set for validation to mitigate over-fitting.




%% file: method.tex
\begin{figure*}[ht!]
    \centering
    \includegraphics[width=0.95\linewidth]{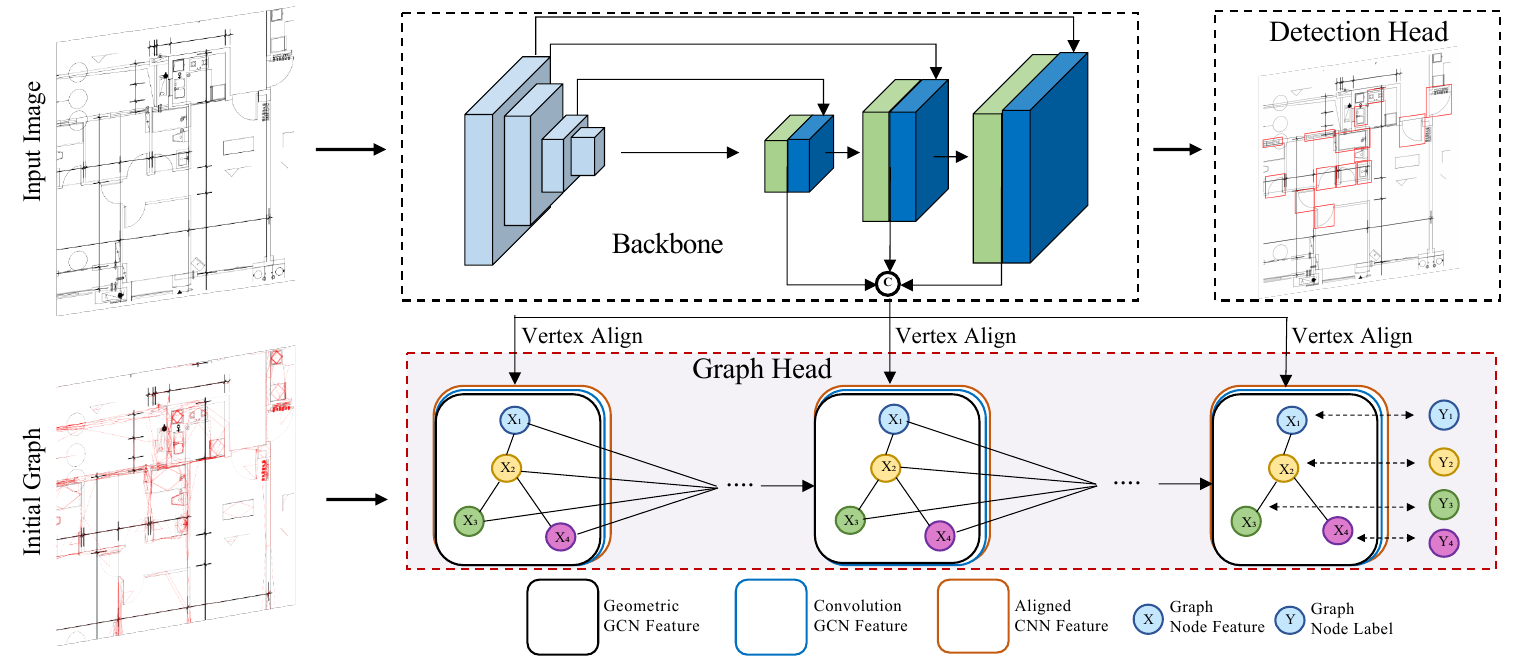}
    \caption{Network architecture of the proposed panoptic symbol spotting model (PanCADNet).}
    \label{fig:framework}
    \vspace{-4mm}
\end{figure*}

\section{PanCADNet}

To tackle the panoptic symbol spotting problem, an end-to-end architecture named PanCADNet is proposed.
It consists of a CNN backbone, a graph convolution head, and detection head,
which are designed for semantic and instance symbol spotting respectively.

\vspace{-4.5mm}
\paragraph{Graph construction}
Given a CAD drawing, a graph $\mathcal{G}=(\mathcal{V},\mathcal{E})$ is constructed, whose vertices are graphical entities $\mathcal{V} = \{e_i\}$.
An edge is created for two vertices if their distance are close enough, namely $D(e_i, e_j) < \epsilon$:
\begin{equation}
    D(e_i, e_j)= 
    \text{min}_{p \in \{e_i^s, e_i^t\}, q \in \{e_j^s, e_j^t\}} \|p - q \|,
\end{equation}
where $e^s$ and $e^t$ represent the start and end points of the graphical entity $e$ respectively, regardless of its shape.
For a pair of parallel lines, we slightly modify the distance as:
\begin{equation}
    D_\|(e_i, e_j)= \eta * \text{min}_{p \in e_i, q \in e_j} \|p - q \|, 
\end{equation}
\begin{wrapfigure}{r}{0.4\linewidth}
  \begin{center}
    \includegraphics[width=\linewidth]{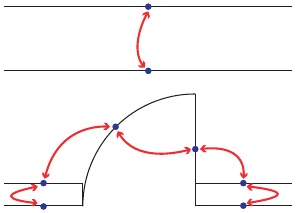}
  \end{center}
\end{wrapfigure}
where $\eta$ is a scale factor for establishing parallelism connection for wall and windows.
To keep the graph complexity low, at almost $K$ edges are allowed for every node by random dropping. 
In our dataset, we have $\epsilon = 100\text{mm}, \eta = 0.2, K = 3$.
Figure on the right illustrates how a graph is constructed, where black dots represent nodes and red curves stands for edges. 

\vspace{-5mm}
\paragraph{Geometric features} 
Graphical entities in a symbol usually share properties such as scale, location, and types.
We encode the length $f_i^l$ and the normalized center location $f_i^p$ of each vertex $e_i$ as its \textit{spatial features}.
In addition, the type (segment, circle or curve) of each vertex is encoded as a one-hot array $f_i^s$, namely its \textit{type feature}. 

\vspace{-5mm}
\paragraph{Textural features}
CNN is capable of extracting multi-scale textural features from an image.
To utilize the power of CNN, we follow Pixel2Mesh~\cite{wang2018pixel2mesh} to obtain the visual feature $f_i^{cnn}$ for each entity.
Specifically, given the 2D coordinate of an entity $e_i$, CNN features is fetched from the aligned position from feature
pyramid networks~\cite{lin2017feature}.
Concatenating with the manually designed geometric features, the node features become $f_i=\{f_i^l,f_i^p,f_i^{cnn}\}$.

\vspace{-5mm}
\paragraph{Graph convolution}
Applying the graph convolution network~\cite{kipf2016semi} on $\mathcal{G}$, the vertex aggregates information by propagating information from neighboring vertices,
\begin{equation}
f_{i}^{'} = ReLU(W_0f_i + \sum_{e_j \in \mathcal{N}(e_i)} W_1 f_j),
\end{equation}
where $\mathcal{N}(e_i)$ are neighbor vertices of graph vertex $e_i$, and $W_0$ and $W_1$ are learnable parameters of the model.
In our experiment, three graph convolution layers are used.

\vspace{-5mm}
\paragraph{Detection head}
Previous methods~\cite{goyal2019bridge, rezvanifar2020symbol} apply object detection models to retrieve instance indices of symbols.
Similarly, we build a two-stage detectors head, namely Faster R-CNN, for instance symbol spotting.
It takes the pyramid features from our backbone network as input and outputs a bounding box with category label and confidence for each detected symbol instance. The loss term in the detection head $Loss_{Detection}$ follows~\cite{ren2016faster}.

\vspace{-4.5mm}
\paragraph{Loss function}
As indicated in Figure~\ref{fig:catalogue}, class imbalance in the training set is evident.
As a result, we design a class weighted loss for GCN head. Here, we adopted weights for different classes by the number of entities in each class:
\begin{equation}
    Loss_{GCN}=
    -\sum_{e_i \in \mathcal{V}}
    \sum_{j = 1}^{\|\mathcal{L}\|} 
    w_j * l_i^g * log(P(l_i^j)),
\end{equation}
where $w_j=\vert\{e_i\vert GT(e_i) = l_j\}\vert/\vert\{e_i\}\vert$,
$l_i^g$ and $l_i^j$ are the ground truth and predicted labels of vertex $e_i$, respectively. $P(\cdot)$ describes the probability of a prediction.
%
%
By integrating loss terms of detection head and GCN head, we are able to train the whole system end-to-end,
\begin{equation}
Loss_{Total}= \lambda * Loss_{GCN} + Loss_{Detection},
\end{equation}
where $\lambda = 3$ in our experiments.


%% file: exp.tex
\begin{figure*}[t!]
    \centering
    \begin{subfigure}{.378\textwidth}
        \centering
        \includegraphics[width=0.85\linewidth]{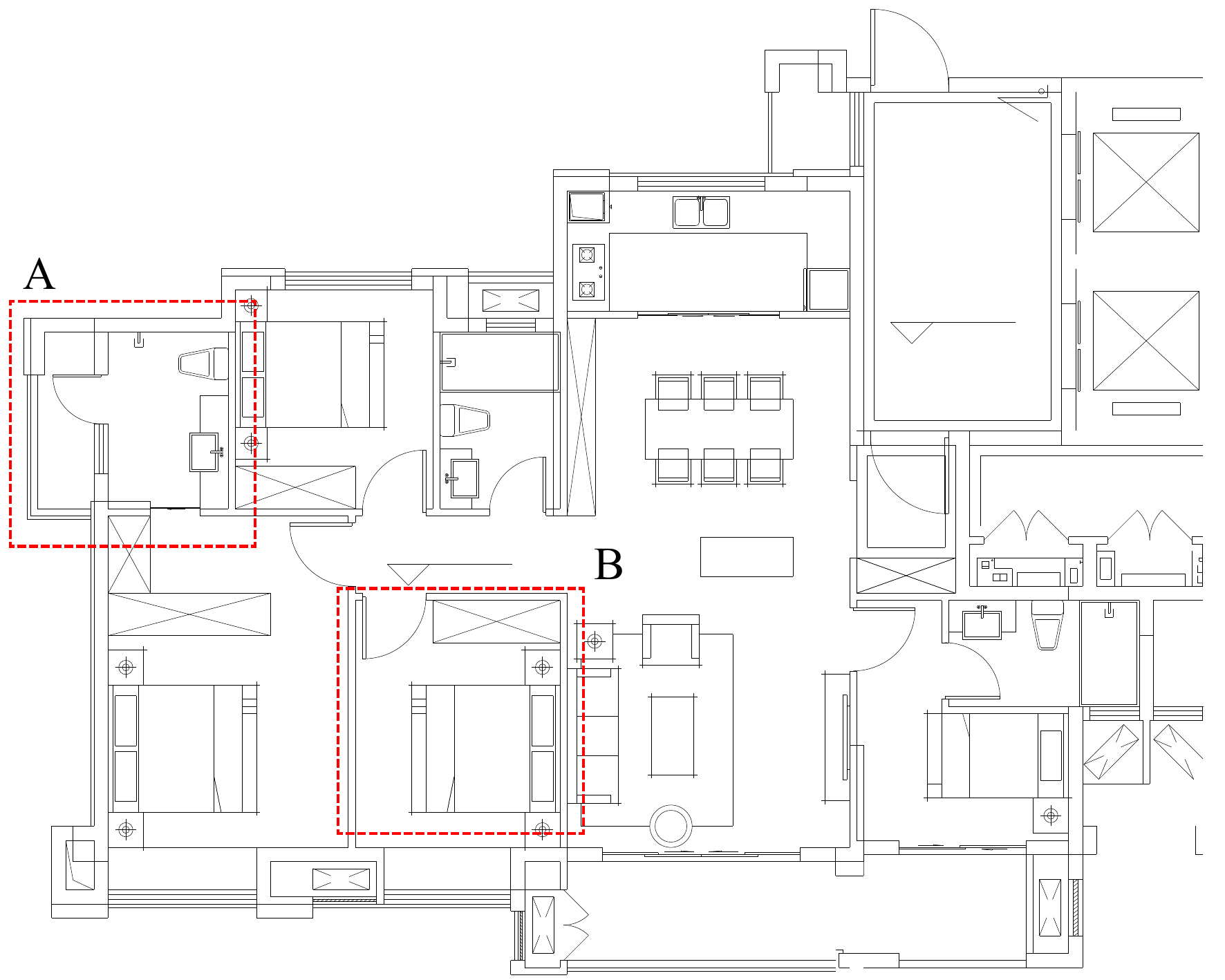}
        \caption{Input CAD drawing with region A and B}
    \end{subfigure}
    \begin{subfigure}{.15\textwidth}
        \centering
        \includegraphics[width=0.85\linewidth]{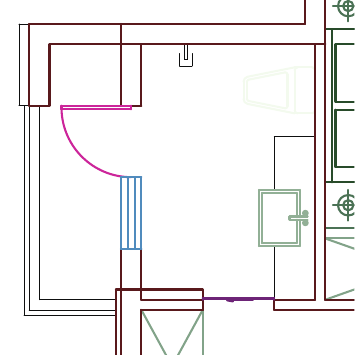}
        \includegraphics[width=0.85\linewidth]{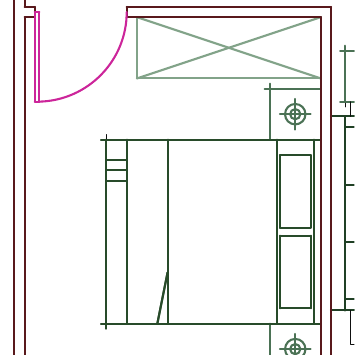}
        \caption{GT}
    \end{subfigure}
    \begin{subfigure}{.15\textwidth}
        \centering
        \includegraphics[width=0.85\linewidth]{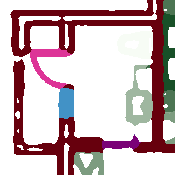}
        \includegraphics[width=0.85\linewidth]{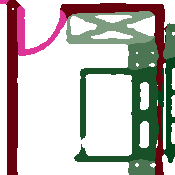}
        \caption{DeepLabv3~\cite{chen2018encoder}}
    \end{subfigure}
    \begin{subfigure}{.15\textwidth}
        \centering
        \includegraphics[width=0.85\linewidth]{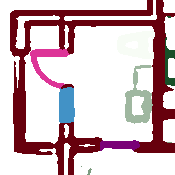}
        \includegraphics[width=0.85\linewidth]{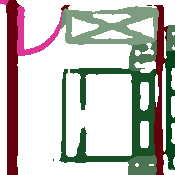}
        \caption{HRNetV2~\cite{wang2020deep}}
    \end{subfigure}
    \begin{subfigure}{.15\textwidth}
        \centering
        \includegraphics[width=0.85\linewidth]{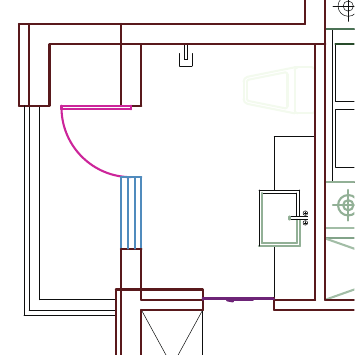}
        \includegraphics[width=0.85\linewidth]{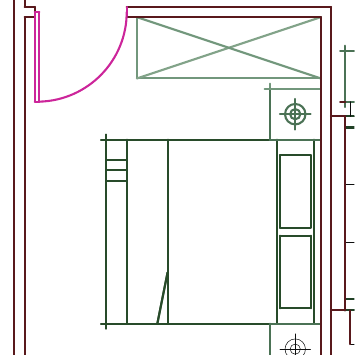}
        \caption{Ours}
    \end{subfigure}
    \vspace{-3mm}
    \caption{Qualitative results on the test set on the \textit{semantic symbol spotting} task. }
    \label{fig:results-sss}
    \vspace{-2mm}
\end{figure*}

\section{Experiments}

We first evaluate the proposed GCN head on semantic symbol spotting task with various image segmentation models to validate the effectiveness of the proposed module. 
Then, we test several detection-based methods and traditional symbol spotting methods on instance symbol spotting task. 
In the panoptic symbol spotting task, we apply the proposed model shown in Figure.~\ref{fig:framework} with tuned modules. 

\subsection{Semantic Symbol Spotting}
We utilize the backbone and GCN head shown in Figure~\ref{fig:framework} as a network for semantic symbol spotting and validate the effectiveness of each component.

\vspace{-4.5mm}
\paragraph{Datasets}
In order to apply semantic segmentation models, annotation masks are generated by projecting graphical entities onto the background canvas with line width of 5 pixels during training.


\vspace{-4.5mm}
\paragraph{Implementation}
HRNetsV2~\cite{wang2020deep} is adopted to extract rich high resolution representations.
We use Adam optimizer with $\beta_1 = 0.9$ and $\beta_2 = 0.999$ for the proposed GCN-based method.
The training is done for 40k iterations with an initial learning rate of 0.0001, we scheduled the learning rate with cosine annealing~\cite{loshchilov2016sgdr}. To push the performance ahead, we adopt AM-Softmax~\cite{wang2018additive} loss relacing the origin softmax loss.
We train our method with 8 Nvidia GTX 2080Ti GPUs with 1 training samples on each GPU. 
All semantic segmentation models are trained for 40k iterations which corresponds to the 40k schedule in the \textit{mmsegmentation}~\cite{mmseg2020}. 

\vspace{-4.5mm}
\paragraph{Benchmark Performance}
In order to compare with traditional image segmentation method on our dataset, their label prediction on pixels are converted to label on entities by sampling and voting:
\begin{equation}
PD(e_i)=\underset{l_i}{\mathrm{argmax}}\vert\{p_k\vert PD(p_k)=l_i,p_k\in e_i\}\vert,
\end{equation}
where $p_k$ indicates the sample point and $PD(p_i)$ is the predicted mask retrieved from the CNN output.

Quantitative results on semantic symbol segmentation task are shown in Table~\ref{tab:semantic}.
We can see that our method with GCN module outperforms both HRNetsV2~\cite{wang2020deep} and DeepLabv3+~\cite{chen2018encoder} significantly, with the improvement of $11.8\%$ on $F_1$ score and $8.4\%$ weighted $F_1$ score, whose weights are similarly defined as Equation~\ref{eq:iou}.
The qualitative results are shown in Figure~\ref{fig:results-sss}.
We can see that our method predicts noticeable better results, especially in class wall, where their lengths have statistical regularity and semantic segmentation methods may fail in the boundary areas.

\subsection{Ablation Study on Semantic Symbol Spotting}

\begin{table}
    \begin{center}
    \resizebox{0.9\linewidth}{!}{
    \scriptsize
    \begin{tabular}{c c c c c | c}
    \hline
    Spa. feat. &Type feat. &CNN feat. &Wei. loss &Am-softmax~\cite{wang2018additive} &wF1. \\
    \hline
    \checkmark &           &           &           &           &0.199 \\
    \checkmark &\checkmark &           &           &           &0.308 \\
    \checkmark &\checkmark &\checkmark &           &           &0.755 \\
    \checkmark &\checkmark &\checkmark &\checkmark &           &0.759 \\
    \checkmark &\checkmark &\checkmark &\checkmark &\checkmark &0.798 \\
    \hline
    \end{tabular}}
    \end{center}
    \vspace{-4.5mm}
    \caption{The quantitative comparisons between proposed GCN model and GCN model with/without its components. 
    }
    \label{tab:ablation}
\end{table}

Extensive ablation studies are performed to validate the effectiveness of our proposed method GCN network.
As we can see in Table~\ref{tab:ablation}, GCN itself with simple geometric features is the worst and strong visual CNN features from the backbone boost the model accuracy significantly.
With topology embedded in the graph convolutional layers, it connects image features and aggregates information from neighbour in a way that is almost impossible for the CNN-based methods. 
Moreover, weighted loss function and AM-softmax loss are both helpful to the proposed method.

\subsection{Instance Symbol Spotting}
In this section, thorough experiments are conducted to evaluate traditional and modern,
supervised and unsupervised methods on the sub-task, namely instance symbol spotting defined at Task formation~\ref{task.task_formation} of the proposed FloorPlanCAD dataset.

\vspace{-4.5mm}
\paragraph{Dataset} 
We uniformly sample points on each element of the annotated CAD drawing and calculate the outside bounding box for each instance. We convert the instance information to COCO-styled version~\cite{lin2014microsoft}.

\vspace{-4.5mm}
\paragraph{Implementation}
All models are trained for 100 epochs except YOLOv3 for 273 epochs. ResNet-101 with FPN is used as backbone for Faster-RCNN and FCOS, DarkNet-53 is used for YOLOv3. Experiments are implemented based on the latest release of mmdetection~\cite{mmdet}.
Two typical traditional methods, SCIP~\cite{nguyen2009symbol} and Graph Matching~\cite{graphmatching}, are implemented and evaluated on this task.

\vspace{-4.5mm}
\paragraph{Benchmark Performance}
Quantitative results on instance symbol detection task are shown in Table~\ref{tab:instance}.
We can see that Faster-RCNN and FCOS achieve comparable accuracy. Both traditional methods perform poorly on the proposed dataset since our data is from different vendors and do not have such standard templates to compare with.


\begin{table*}[t!]
    \begin{center}
    \resizebox{0.95\linewidth}{!}{
    \footnotesize
    \begin{tabular}{c | c c c c c c c c | c c}
    \hline
    Categories &Door &Window &Stair &Appliance &Furniture &Equipment &Wall &Parking lot &F1 &Weighted F1 \\
    \hline
    HRNetsV2 W18~\cite{wang2020deep}       &0.821   &0.620   &0.845   &0.597   &0.726   &0.880   &0.620   &0.610   &0.656   &0.683 \\
    HRNetsV2 W48~\cite{wang2020deep}       &0.811   &0.640   &0.847   &0.651   &0.754   &0.889   &0.624   &0.577   &0.666   &0.693 \\
    DeepLabv3+ R50~\cite{chen2018encoder}  &0.828   &0.659   &0.856   &0.684   &0.763   &0.895   &0.630   &0.664   &0.680   &0.705 \\
    DeepLabv3+ R101~\cite{chen2018encoder} &0.837   &0.666   &0.852   &0.725   &\textbf{0.780}   &0.895   &0.634   &\textbf{0.669}   &0.688   &0.714 \\
    \hline
    Ours                                   &\textbf{0.848}  &\textbf{0.709}   &\textbf{0.857}   &\textbf{0.769}   &0.764   &\textbf{0.926}   &\textbf{0.814}   &0.539   &\textbf{0.806}   &\textbf{0.798} \\
    \hline
    \end{tabular}}
    \end{center}
    \vspace{-2mm}
    \caption{Statistical results on the proposed dataset of different semantic segmentation models and our GCN-based method. Both HRNetsV2~\cite{wang2020deep} and DeepLabv3+~\cite{chen2018encoder} show that deeper networks produce better results.}
    \label{tab:semantic}
    \vspace{-2mm}
\end{table*}

\begin{table}
    \begin{center}
    \footnotesize
    \begin{tabular}{c | c c c c}
    \hline
    Methods &backbone &AP50 &AP75 &mAP \\
    \hline
    Faster R-CNN~\cite{ren2016faster}   &R101 &0.602   &0.510   &0.452 \\
    FCOS~\cite{tian2019fcos}            &R101 &0.624   &0.491   &0.453 \\
    YOLOv3~\cite{redmon2018yolov3}      &DarkNet53 &0.639   &0.452   &0.413 \\
    SCIP~\cite{nguyen2009symbol}        &-    &0.231   &0.151   &0.135 \\
    Graph Matching~\cite{le2012integer} &-    &0.137   &0.118   &0.102 \\
    \hline
    \end{tabular}
    \end{center}
    \vspace{-2mm}
    \caption{Quantitative results of different detection methods and symbol spotting methods on the symbol detection task.}
    \label{tab:instance}
\end{table}





\begin{figure}
    \centering
    \begin{subfigure}{.48\linewidth}
        \centering
        \includegraphics[width=\linewidth]{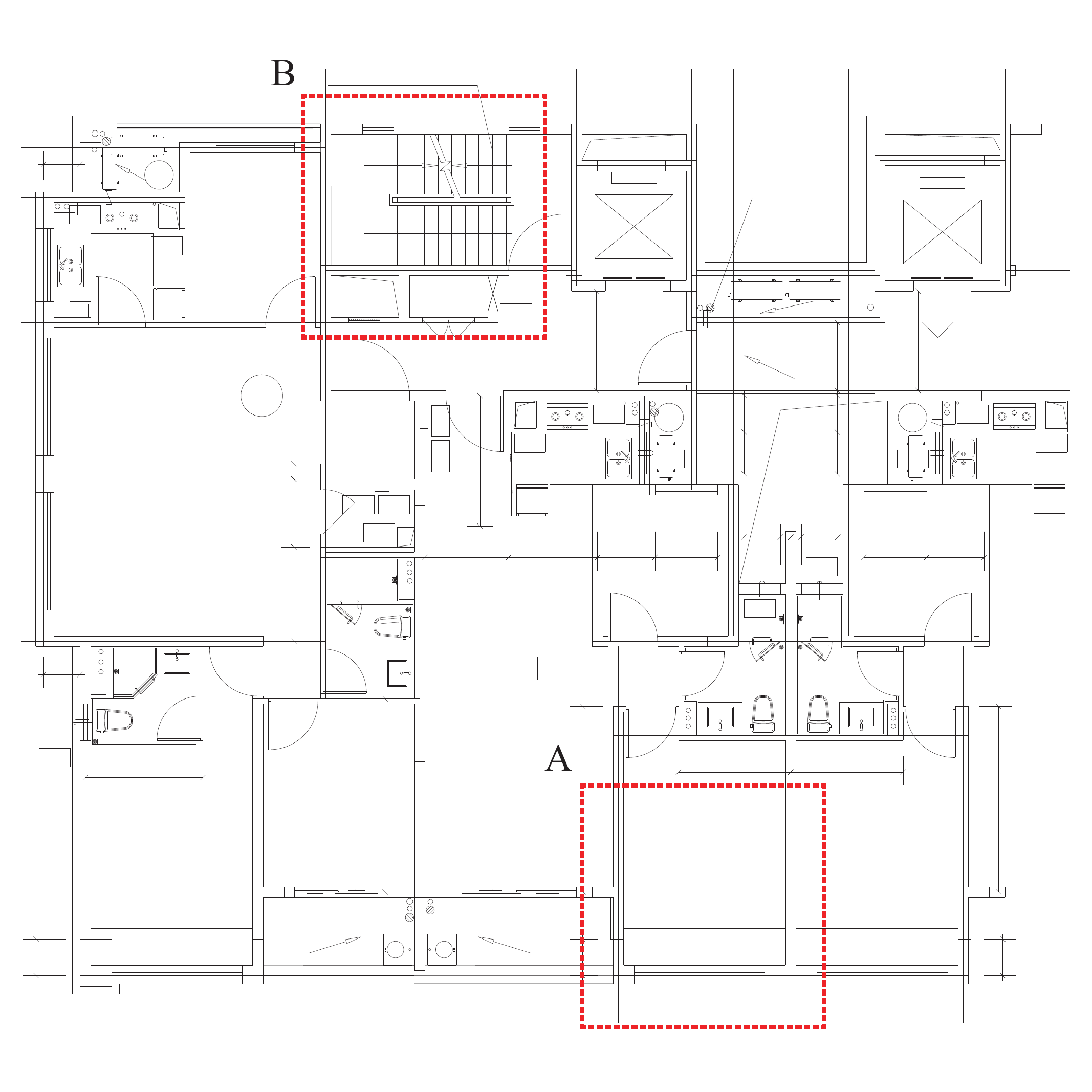}
    \end{subfigure}
    \hfill
    \begin{subfigure}{.493\linewidth}
        \centering
        \includegraphics[width=.48\linewidth]{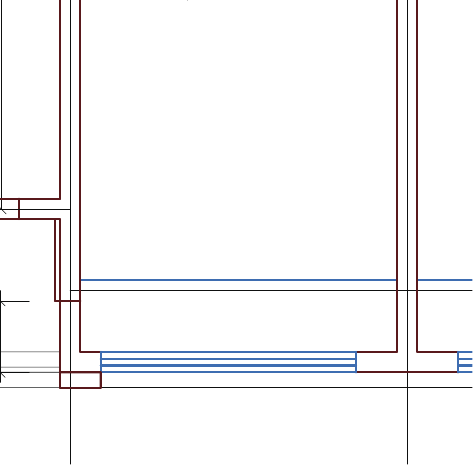}
        \hfill
        \includegraphics[width=.48\linewidth]{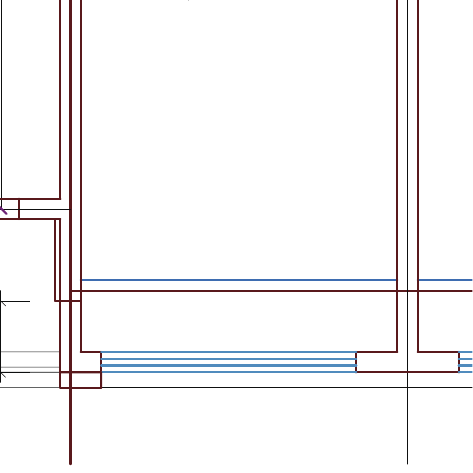}
        \includegraphics[width=.48\linewidth]{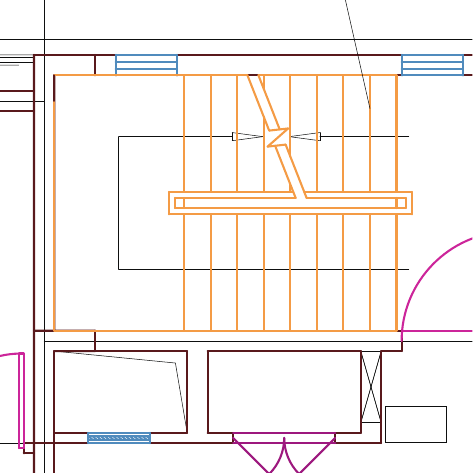}
        \hfill
        \includegraphics[width=.48\linewidth]{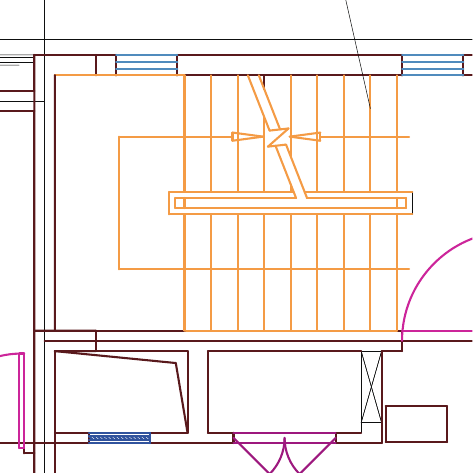}
    \end{subfigure}
    \begin{subfigure}{.48\linewidth}
        \centering
        \includegraphics[width=\linewidth]{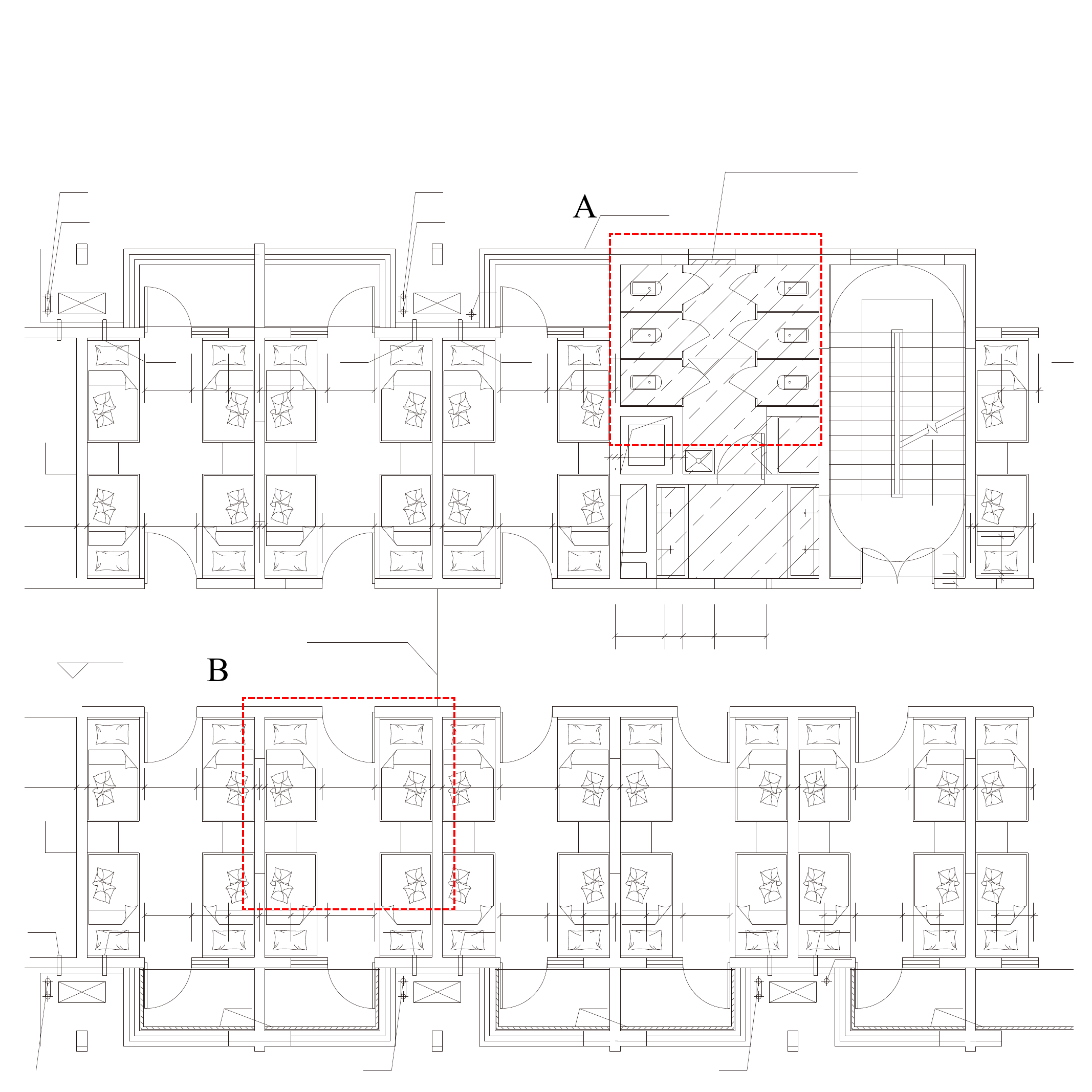}
    \end{subfigure}
    \hfill
    \begin{subfigure}{.493\linewidth}
        \centering
        \includegraphics[width=.48\linewidth]{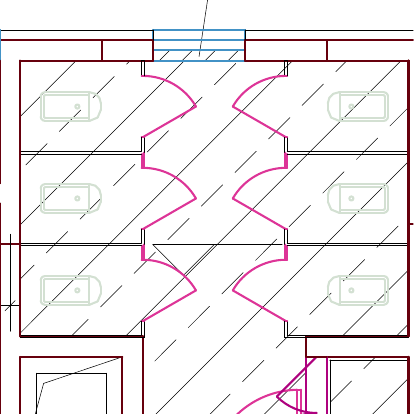}
        \hfill
        \includegraphics[width=.48\linewidth]{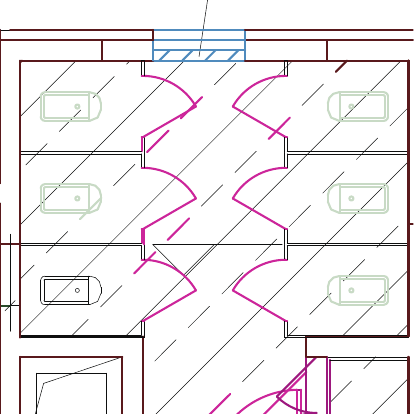}
        \includegraphics[width=.48\linewidth]{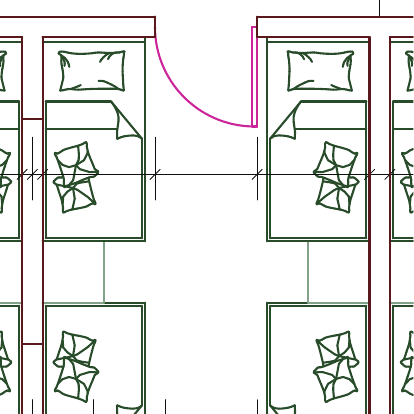}
        \hfill
        \includegraphics[width=.48\linewidth]{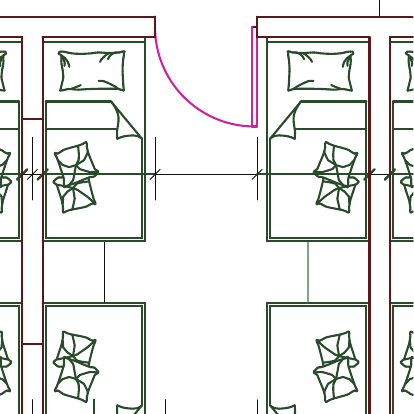}
    \end{subfigure}
    \begin{subfigure}{.48\linewidth}
        \centering
        \includegraphics[width=\linewidth]{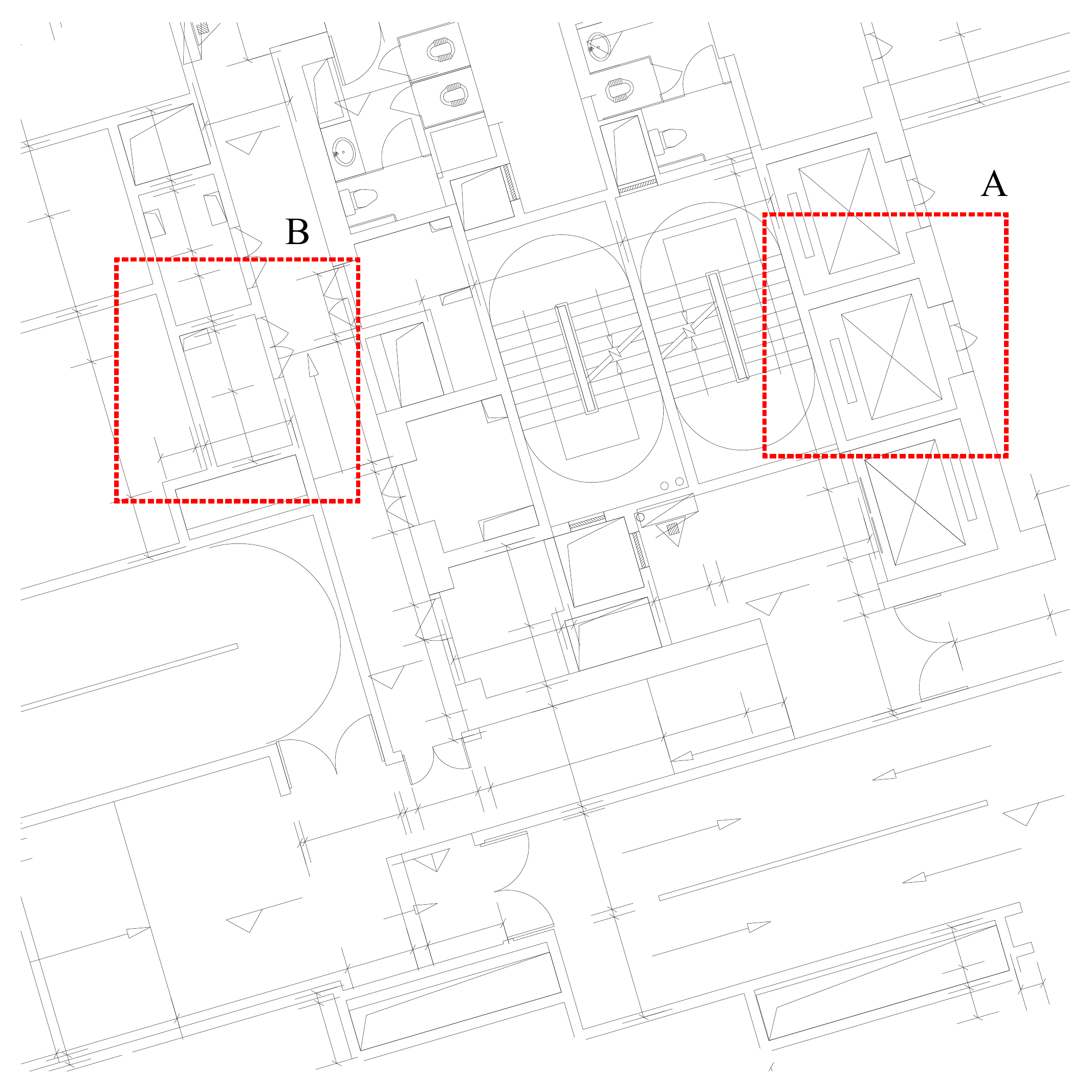}
        \caption{Input floor plan}
    \end{subfigure}
    \hfill
    \begin{subfigure}{.493\linewidth}
        \centering
        \includegraphics[width=.48\linewidth]{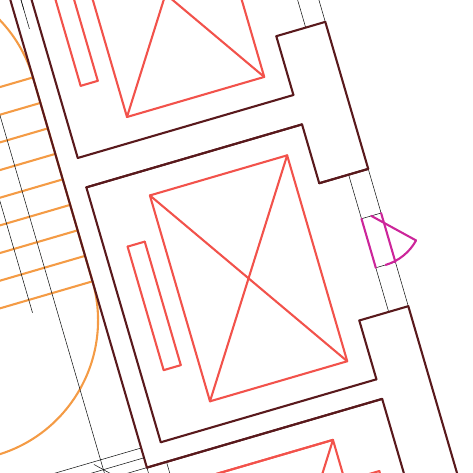}
        \hfill
        \includegraphics[width=.48\linewidth]{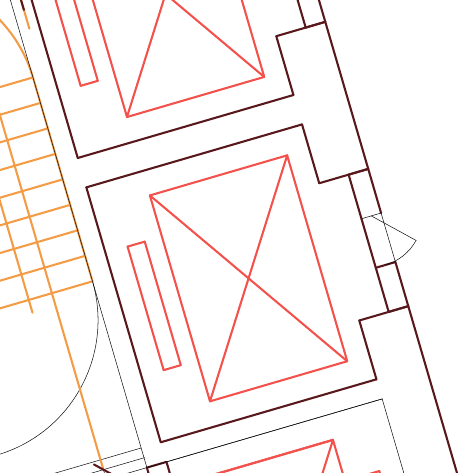}
        \includegraphics[width=.48\linewidth]{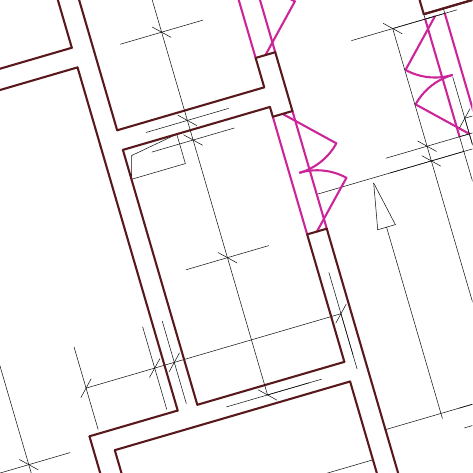}
        \hfill
        \includegraphics[width=.48\linewidth]{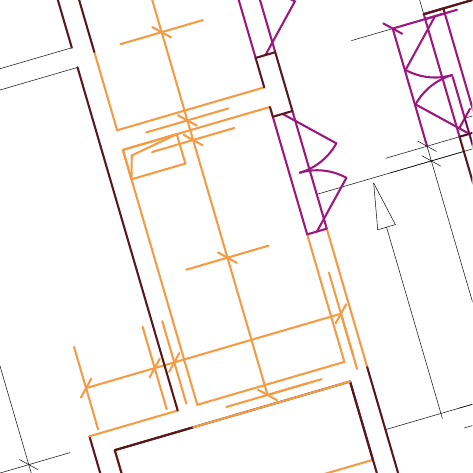}
        \caption{GT (L) and prediction (R)}
    \end{subfigure}
    \caption{The proposed panoptic symbol spotting results on various scenes in the FloorPlanCAD dataset. From top to down: apartment, dormitory and shopping mall.}
    \label{fig:results_panoptic}
    \vspace{-3mm}
\end{figure}

\subsection{Panoptic symbol spotting}
With the optimized GCN head and detection head, then we propose the PanCADNet shown in Figure~\ref{fig:framework}. To our knowledge, there is no other method conducting similar task on vector graphics. 

\vspace{-2mm}
\paragraph{Dataset}
For the panoptic symbol spotting task, we merge the $5$ semantic symbol and $30$ instance symbol classes for the classification graph head.
The box annotation for the detection head is the same to the previous section.

\vspace{-2mm}
\paragraph{Implementation}
We set the training scheme of PanCADNet similar to our semantic symbol spotting model except that we use ResNet-50 as backbone to balance the model efficiency and accuracy. By enforcing an overlap greater than 0.5 IoU, we are given a unique matching similar to~\cite{kirillov2019panoptic}.

\vspace{-2mm}
\paragraph{Benchmark Performance}
Table~\ref{tab:panoptic} lists the calculated panotic quality on the proposed panoptic symbol spotting task.
Visualizations of several typical scenes of panoptic results are shown in Figure~\ref{fig:results_panoptic}.
Bay window in the first row area A is correctly detected, which has only one more parallel line segment further away compared to the ordinary window symbol.
Several failure cases are visible in second row area A and thrid row area B.
All graphics shown in the experiment are vector graphics, please feel free to zoom in and check.





\begin{table}
    \begin{center}
    \footnotesize
    \begin{tabular}{c | c c c}
    \hline
    Dataset &Pan. Quality &Seg. Quality &Recog. Quality \\
    \hline
    FloorPlanCAD   &0.561 &0.838  &0.660    \\
    \hline
    \end{tabular}
    \end{center}
    \vspace{-2mm}
    \caption{Quantitative results of the proposed PanCADNet on the proposed FloorPlanCAD test set. Metrics for every class can be found in supplemental materials.}
    \label{tab:panoptic}
    \vspace{-2mm}
\end{table}



%% file: conclusion.tex
\section{Conclusion}

In this paper, we present a large-scale CAD drawing dataset of real-world floor plans with line-grained annotations.
Based on the characteristics of the vector graphics, we introduce the panoptic symbol spotting problem with evaluation metric.
By integrating CNN features into the GCN, we show the power of combining topological and geometric features.
Our dataset and code with be public available and the next version dataset is under construction.

%% file: supplementary/supplementary.tex
\begin{table*}
    \begin{center}
    \footnotesize
    \begin{tabular}{c | c | c c | c c c | c c c}
    \hline
    \multirow{3}{*}{Class}   
    &\multicolumn{1}{c|}{Property}
    &\multicolumn{2}{c|}{Semantic Symbol Spotting}
    &\multicolumn{3}{c|}{Instance Symbol Spotting}
    &\multicolumn{3}{c}{Panoptic Symbol Spotting}\\
    \cline{2-10}
          &\multirow{2}{*}{$\#$Entity($\times 10^{4}$)}          &\multicolumn{2}{c|}{weighted F1}  &\multicolumn{3}{c|}{mAP} &PQ &SQ &RQ
          \\ \cline{3-10}
          & &GCN-based  &DeepLabv3+~\cite{chen2018encoder}  &Faster R-CNN~\cite{ren2016faster} &FCOS~\cite{tian2019fcos} &YOLOv3~\cite{redmon2018yolov3} &\multicolumn{3}{c}{PanCADNet}
          \\\hline
    single door         &$301$   &0.885 &0.827   &0.843 &0.859 &0.829 &0.763 &0.878 &0.869
          \\\hline
    double door         &$285$   &0.796 &0.831   &0.779 &0.771 &0.743 &0.748 &0.845 &0.885
          \\\hline
    sliding door        &$122$   &0.874 &0.876   &0.556 &0.494 &0.481 &0.763 &0.895 &0.852
          \\\hline
          
    window              &$266$    &0.691 &0.603  &0.518 &0.465 &0.379 &0.459 &0.795 &0.577
          \\\hline
    bay window          &$15.1$   &0.050 &0.163   &0.068 &0.169 &0.062 &0.154 &0.595 &0.260
          \\\hline
    blind window        &$98.6$   &0.833 &0.856   &0.614 &0.520 &0.322 &0.706 &0.869 &0.813
          \\\hline
    opening symbol      &$2.68$   &0.451 &0.721   &0.496 &0.542 &0.168 &0.455 &0.945 &0.481
          \\\hline
         
    stairs              &$197$   &0.857 &0.853   &0.464 &0.487 &0.370 &0.608 &0.784 &0.775
          \\\hline
          
    gas stove           &$175$   &0.789 &0.847   &0.503 &0.715 &0.601 &0.743 &0.957 &0.776
          \\\hline
    refrigerator        &$55.0$   &0.705 &0.730   &0.767 &0.774 &0.723 &0.769 &0.888 &0.866
          \\\hline
    washing machine     &$115$   &0.784 &0.569   &0.379 &0.261 &0.374 &0.430 &0.719 &0.599
          \\\hline
          
    sofa                &$105$   &0.606 &0.674   &0.160 &0.133 &0.435 &0.252 &0.928 &0.272
          \\\hline
    bed                 &$1480$   &0.893 &0.908   &0.713 &0.738 &0.664 &0.805 &0.909 &0.886
          \\\hline
    chair               &$176$   &0.524 &0.543   &0.112 &0.087 &0.132 &0.481 &0.802 &0.600
          \\\hline
    table               &$77.9$   &0.354 &0.496   &0.175 &0.109 &0.173 &0.228 &0.811 &0.282
          \\\hline
    bedside cupboard    &$68.0$   &0.509 &0.770   &0.231 &0.363 &0.310 &0.600 &0.825 &0.727
          \\\hline
    TV cabinet          &$32.8$   &0.581 &0.611   &0.231 &0.187 &0.247 &0.426 &0.800 &0.533
          \\\hline
    half-height cabinet &$4.18$   &0.144 &0.179   &0.133 &0.108 &0.110 &0.009 &0.970 &0.009
          \\\hline
    high cabinet        &$20.1$   &0.325 &0.426   &0.271 &0.188 &0.296 &0.287 &0.820 &0.351
          \\\hline
    wardrobe            &$502$   &0.462 &0.426   &0.325 &0.354 &0.354 &0.433 &0.821 &0.527
          \\\hline
    sink                &$512$   &0.825 &0.844   &0.468 &0.470 &0.384 &0.778 &0.895 &0.870
          \\\hline          
    bath                &$254$   &0.540 &0.432   &0.422 &0.446 &0.430 &0.413 &0.720 &0.573
          \\\hline
    bath tub            &$45.8$   &0.476 &0.637   &0.259 &0.248 &0.215 &0.817 &0.856 &0.955
          \\\hline
    squat toilet        &$139$   &0.842 &0.904   &0.836 &0.821 &0.599 &0.901 &0.989 &0.911
          \\\hline
    urinal              &$118$   &0.866 &0.923   &0.780 &0.762 &0.622 &0.921 &0.981 &0.938
          \\\hline
    toilet              &$298$   &0.875 &0.864   &0.666 &0.599 &0.664 &0.831 &0.906 &0.917
          \\\hline
          
    elevator            &$78.7$   &0.948 &0.900   &0.767 &0.816 &0.750 &0.838 &0.897 &0.935
          \\\hline          
    escalator           &$10.0$   &0.744 &0.864   &0.115 &0.190 &0.129 &0.439 &0.718 &0.612
          \\\hline
          
    parking             &$163$   &0.529 &0.667   & -    & -    & -    &0.251 &0.661 &0.380
          \\\hline
    wall                &$1880$   &0.814 &0.634   & -    & -    & -    &0.451 &0.661 &0.682
          \\\hline 
    total               &$7600$     &0.798 &0.714   &0.452 &0.453 &0.413 &0.561 &0.838 &0.660
          \\\hline 
    \end{tabular}
    \end{center}
    \vspace{-2mm}
    \caption{Dataset entities number and quantitative results for semantic symbol spotting, instance symbol spotting and panoptic symbol spotting of each category. Entity length weighted F1 is used for semantic symbol spotting evaluation, mAP is used for instance symbol spotting evaluation, panoptic quality is used for panoptic symbol spotting evaluation. Note that, the results are reported based on V1 dataset which contains only 30 classes}
    \label{tab:per_class}
\end{table*}

\section{Appendix Introduction}
Due to space limitations in the paper, this supplemental
material contains more descriptions about the dataset and more quantitative and qualitative results of the proposed methods.

\section{Dataset}
\subsection{Entity Distribution among Classes}
In Table~\ref{tab:per_class}, we provide statistics of graphical entities of 30 object classes, including 28 thing classes and 2 stuff classes. We can see that the \textit{wall} class takes up a large portion of the whole dataset.

\subsection{Visualizations on Train Set}
\begin{figure*}
\centering
\begin{tabular}{c|c}
    \includegraphics[width=0.43\textwidth]{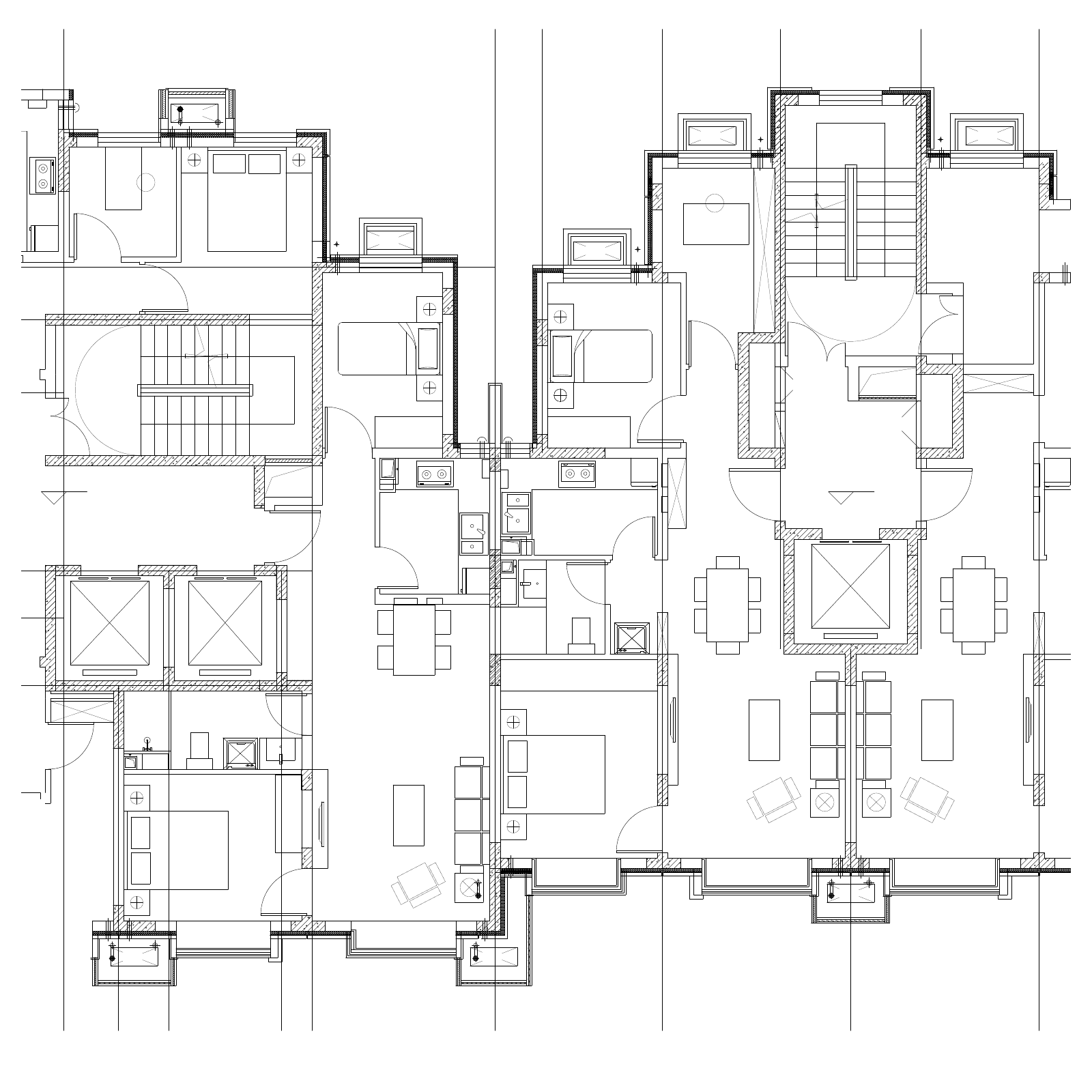} & \includegraphics[width=0.43\textwidth]{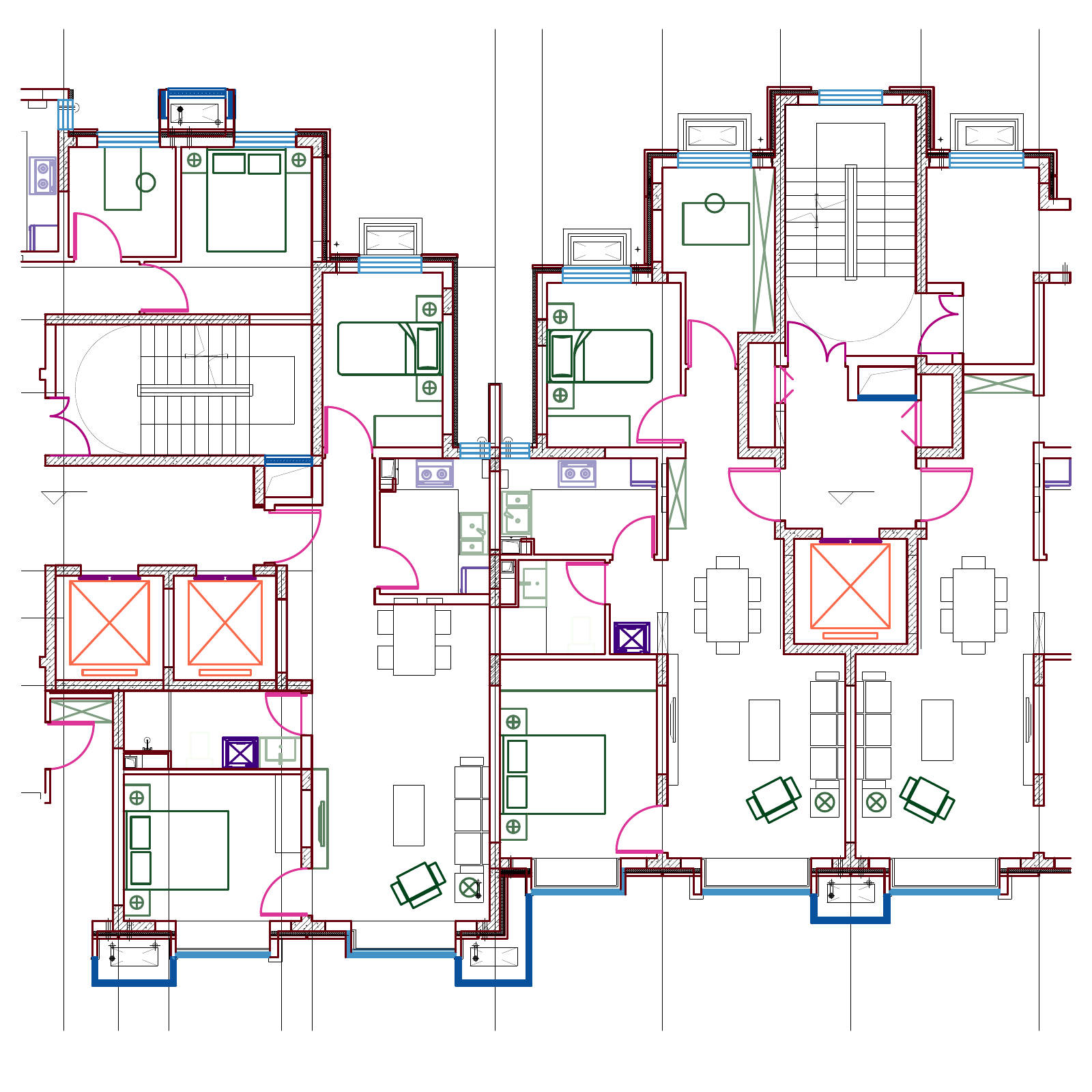} \\ \hline
    \includegraphics[width=0.43\textwidth]{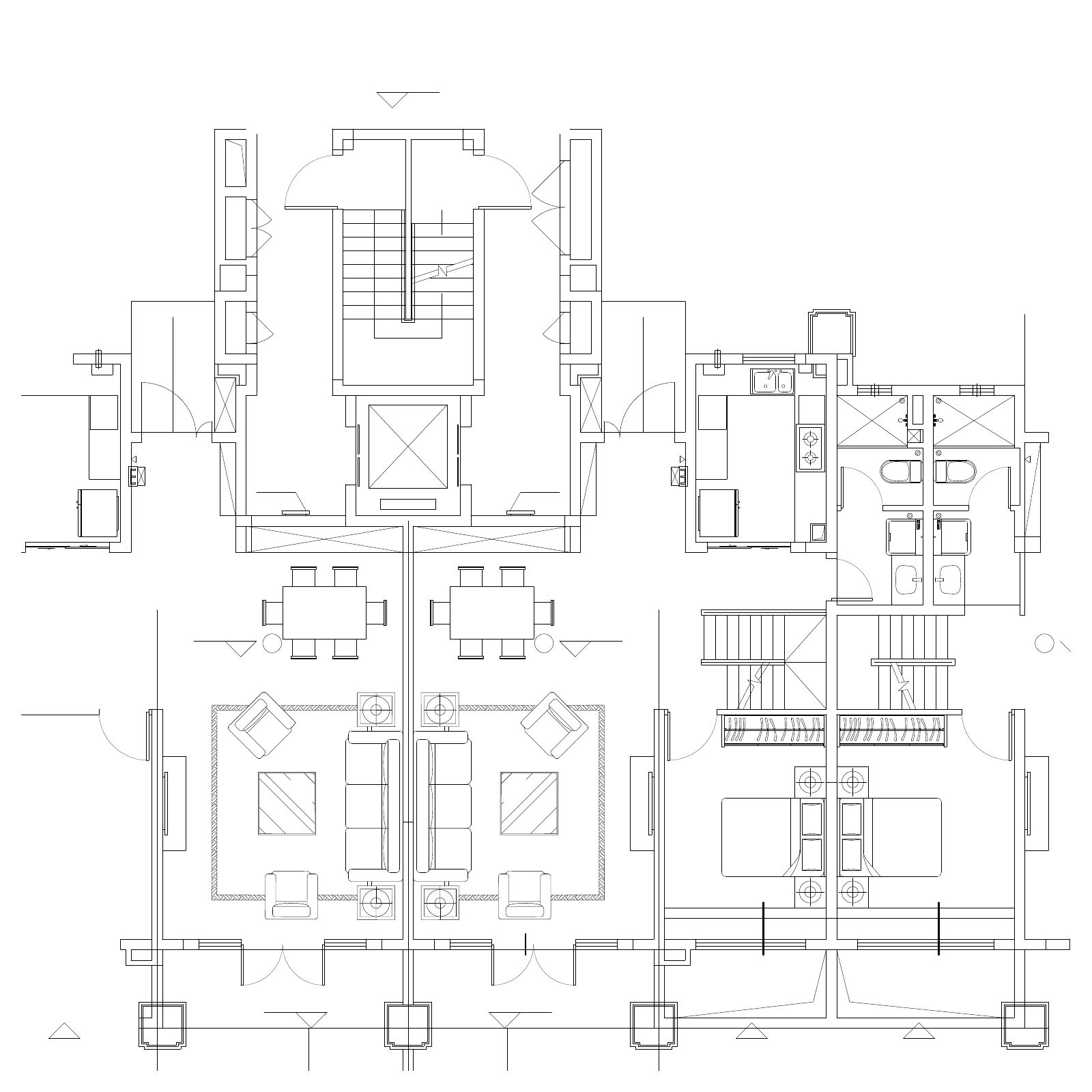} & \includegraphics[width=0.43\textwidth]{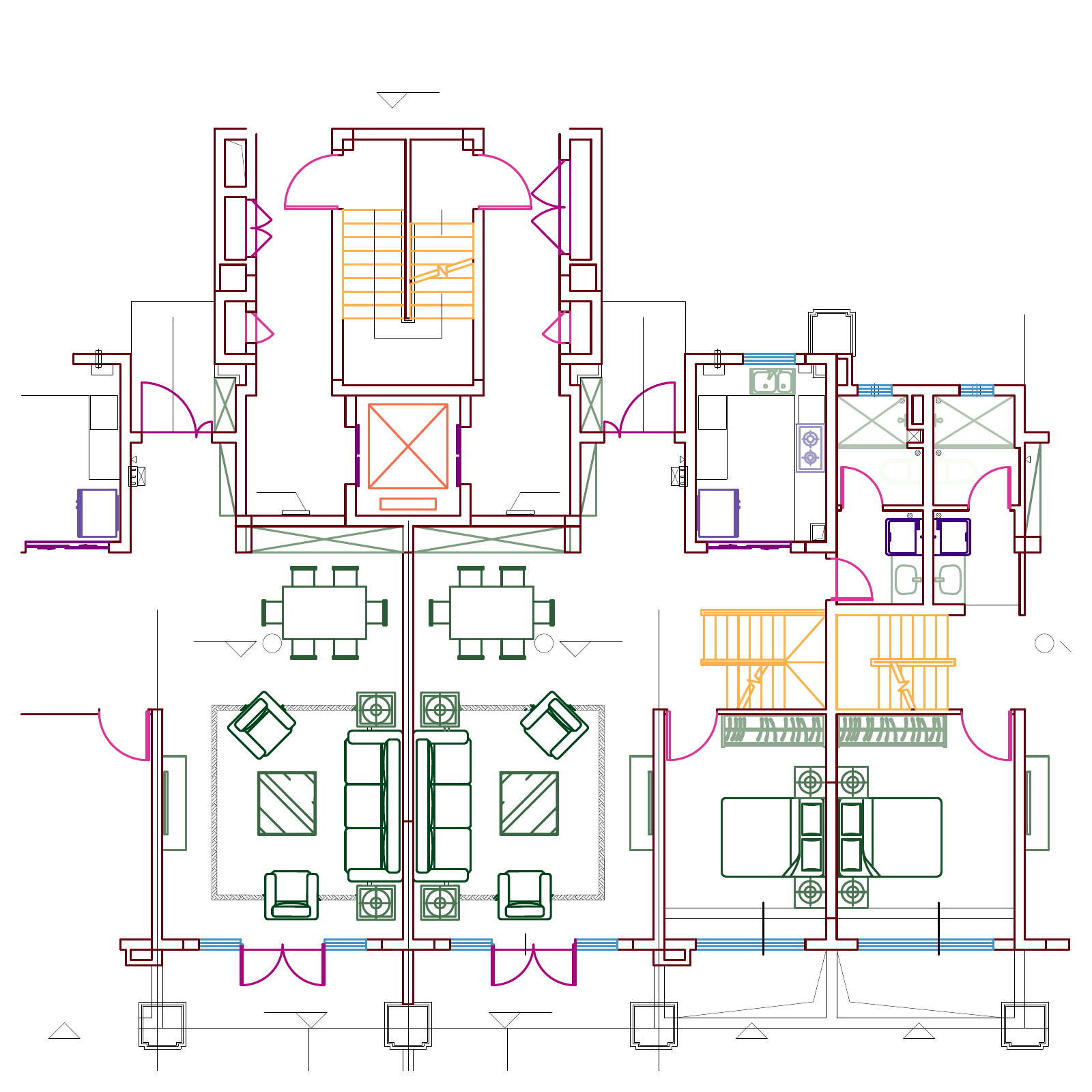} \\  \hline
    \includegraphics[width=0.43\textwidth]{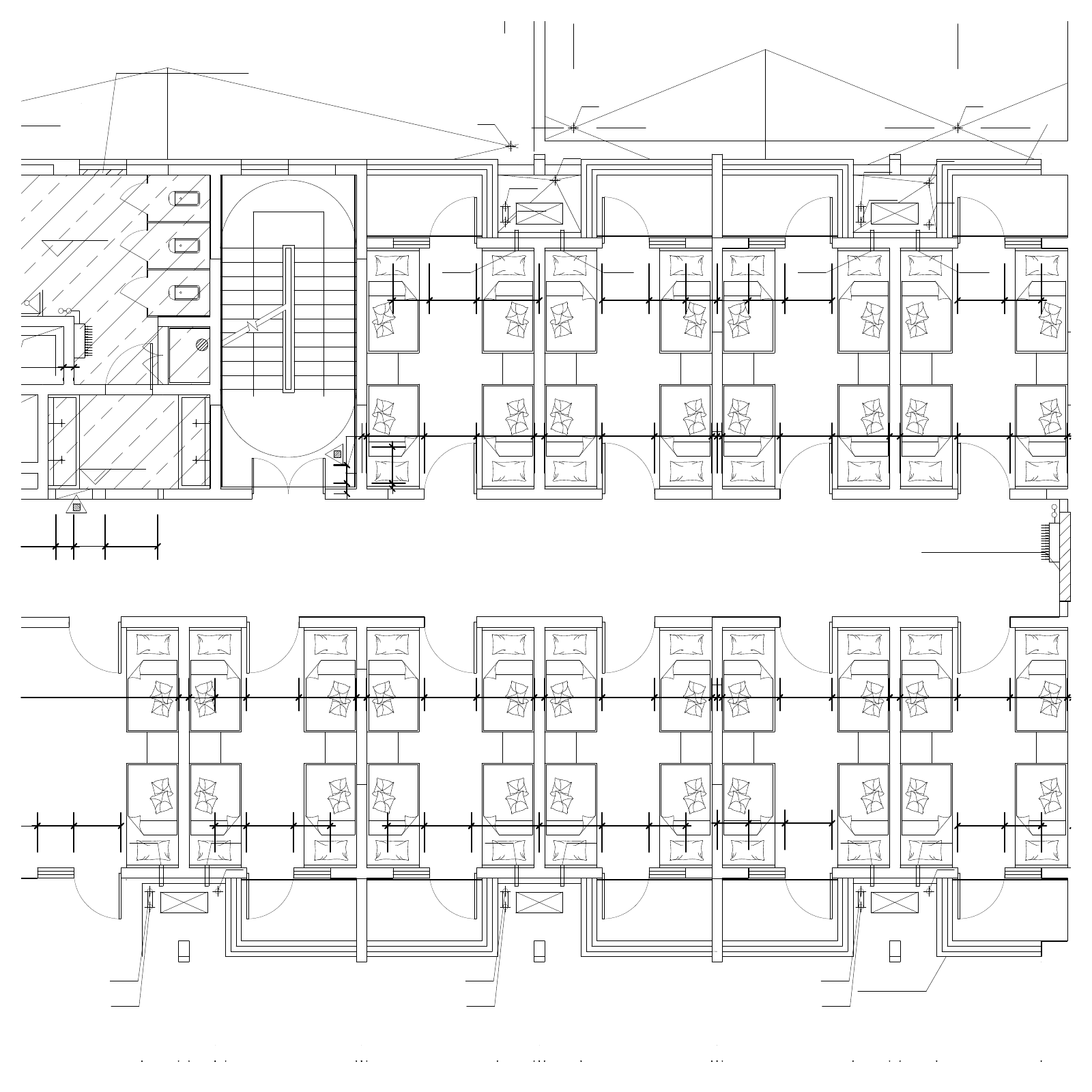} & \includegraphics[width=0.43\textwidth]{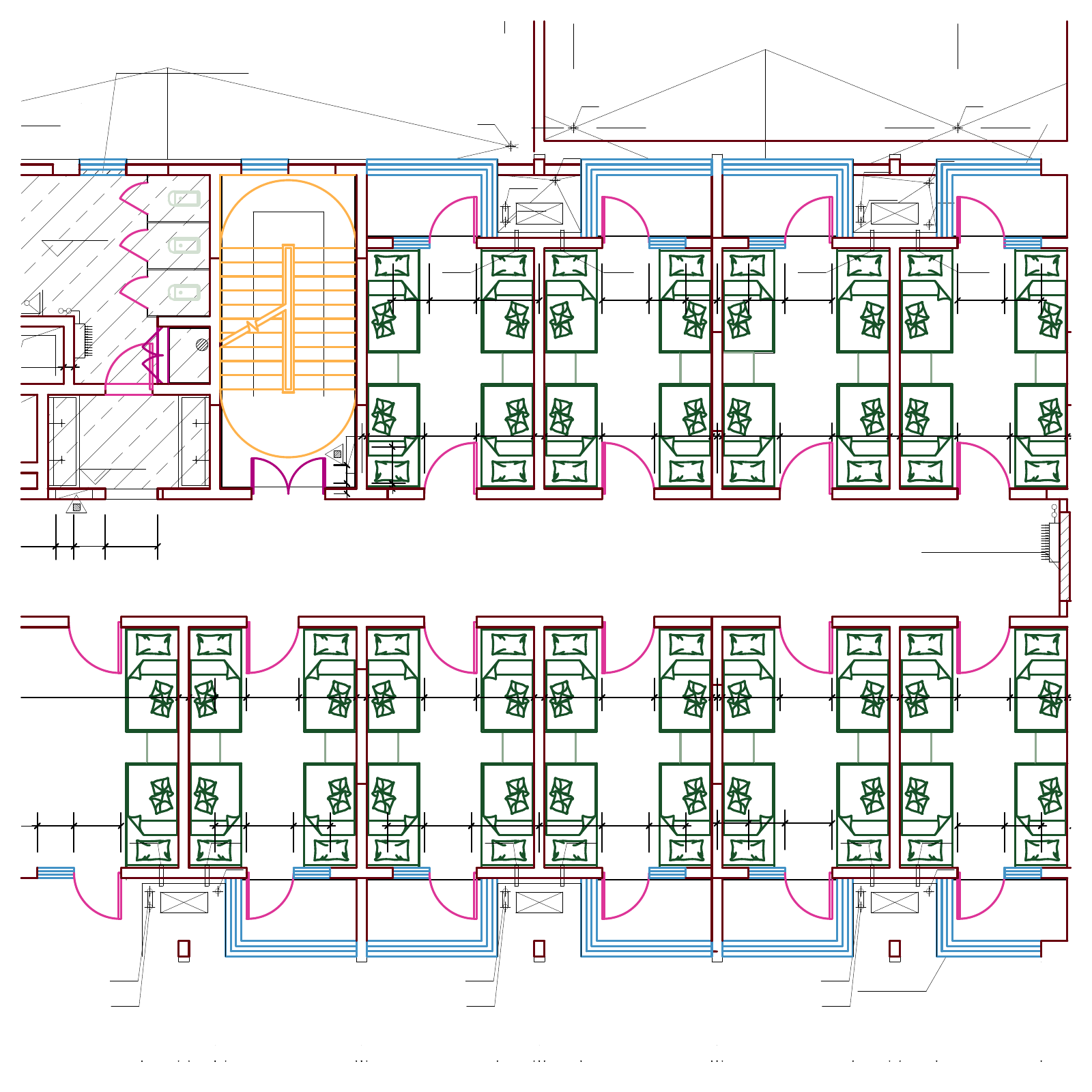} \\
    Raw Input & GT \\
\end{tabular}
\caption{Exemplary raw inputs and annotations in FloorPlanCAD, see the main manuscript for annotation details. The images are part of our \textit{train} set of \textit{residential building} and \textit{school} CAD drawings.}\label{fig:data_train_1}
\end{figure*}

\begin{figure*}
\centering
\begin{tabular}{c|c}
    \includegraphics[width=0.43\textwidth]{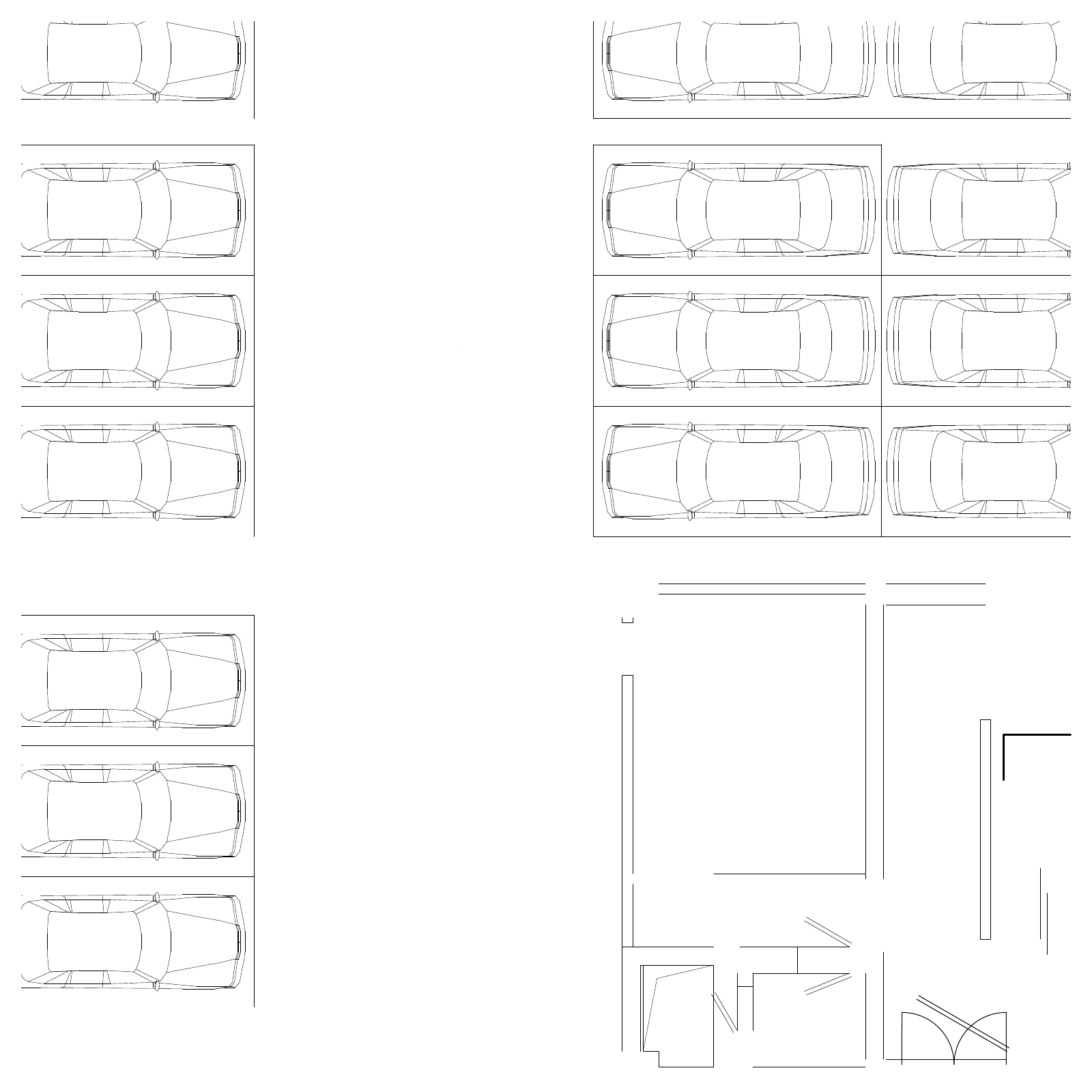} & \includegraphics[width=0.43\textwidth]{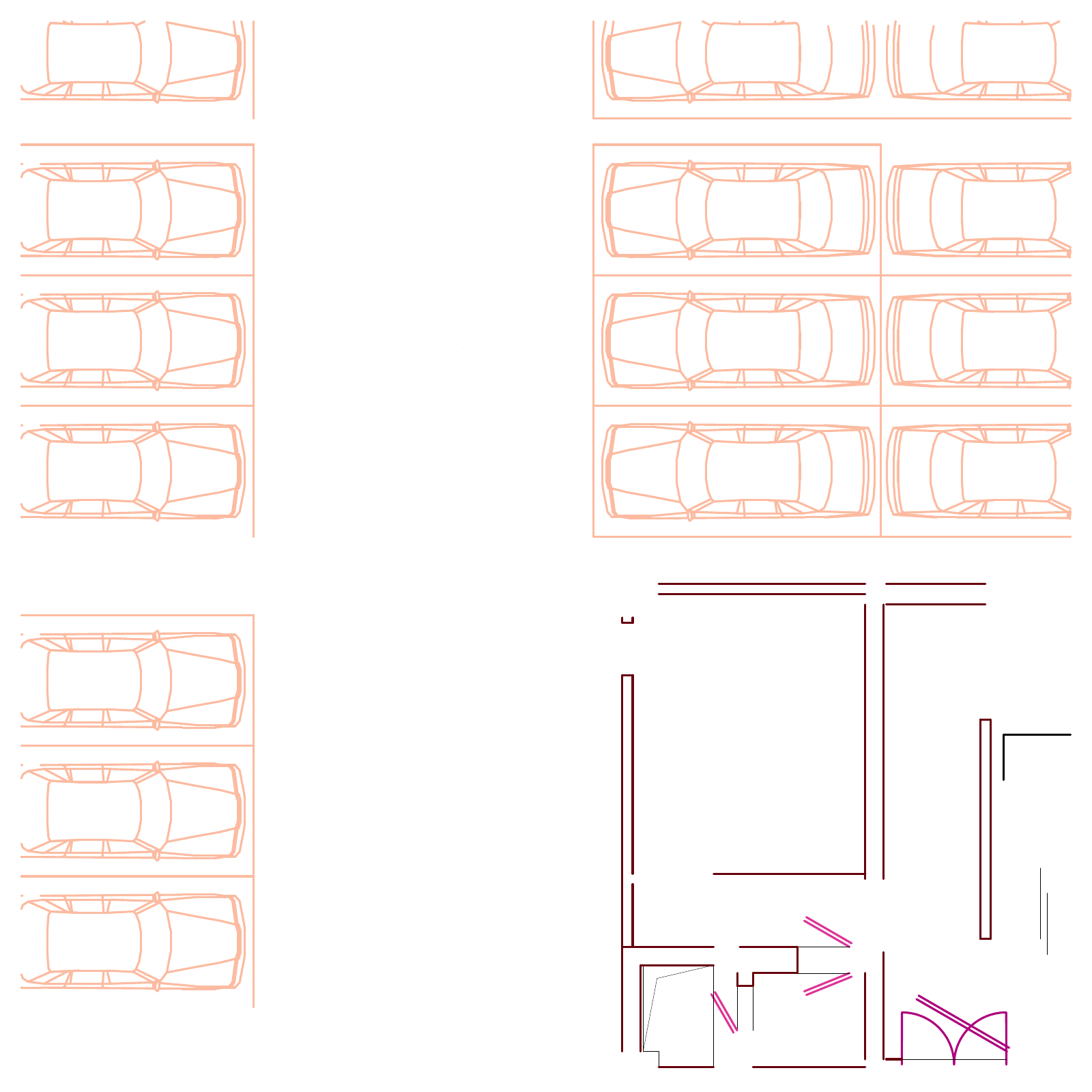} \\ \hline
    \includegraphics[width=0.43\textwidth]{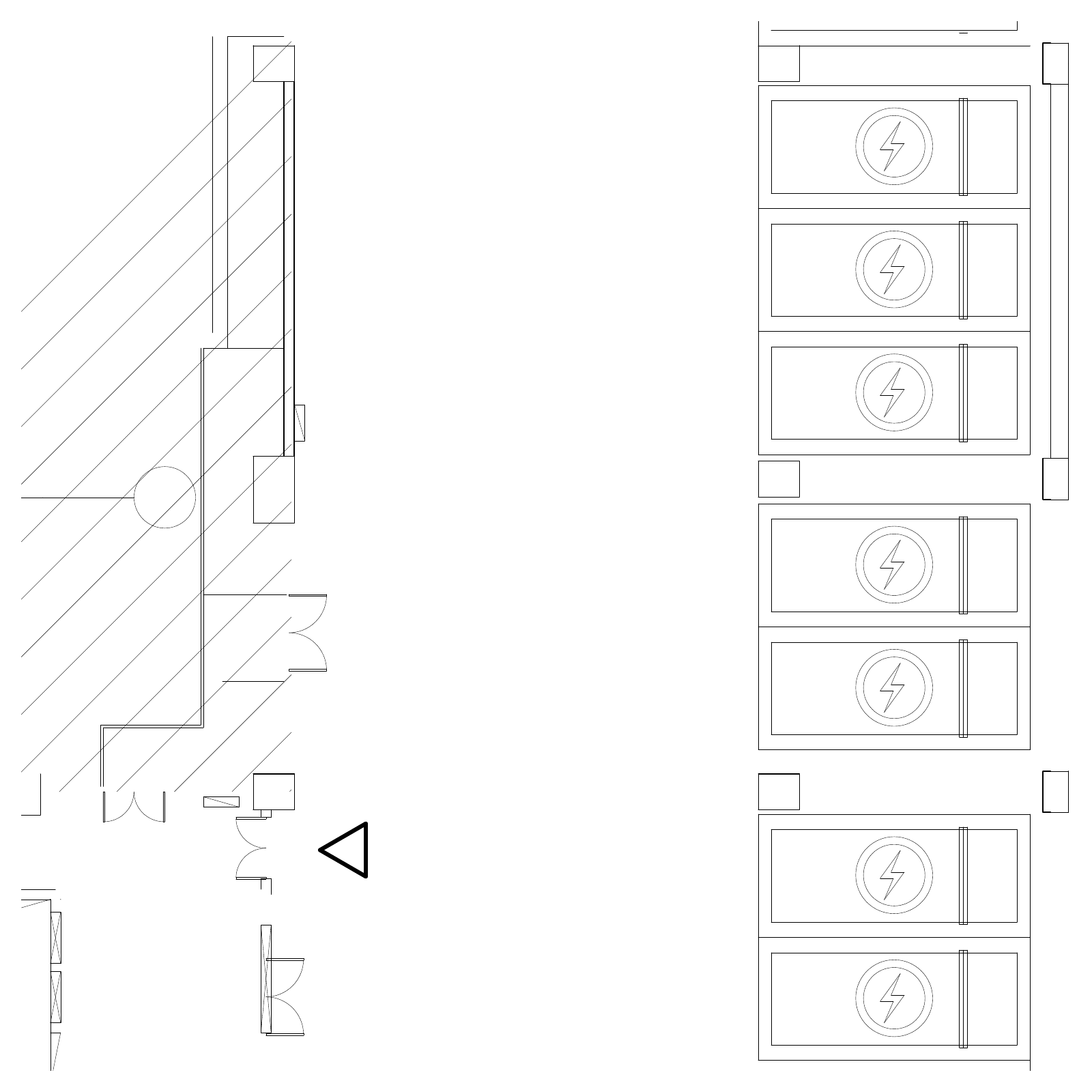} & \includegraphics[width=0.43\textwidth]{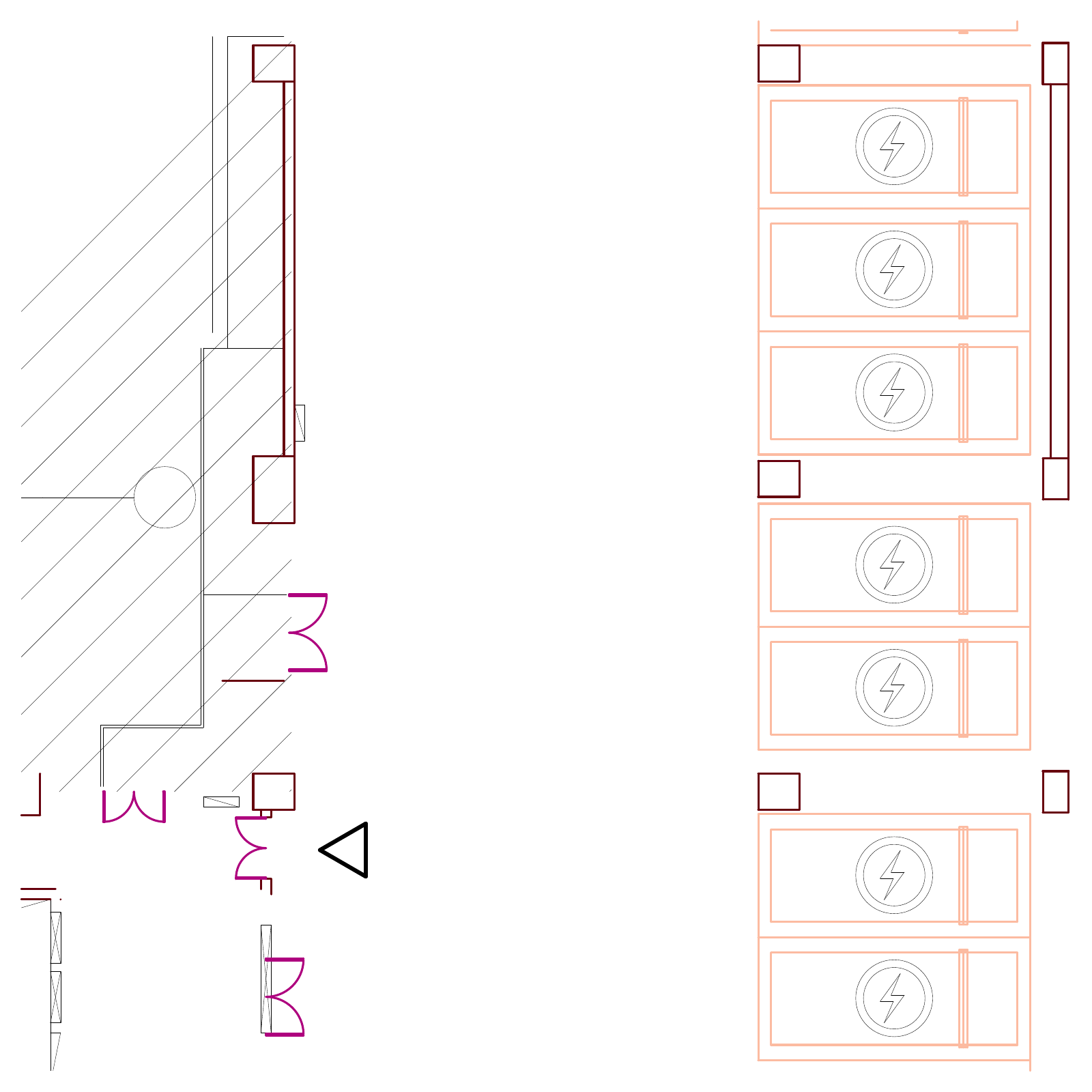} \\  \hline
    \includegraphics[width=0.43\textwidth]{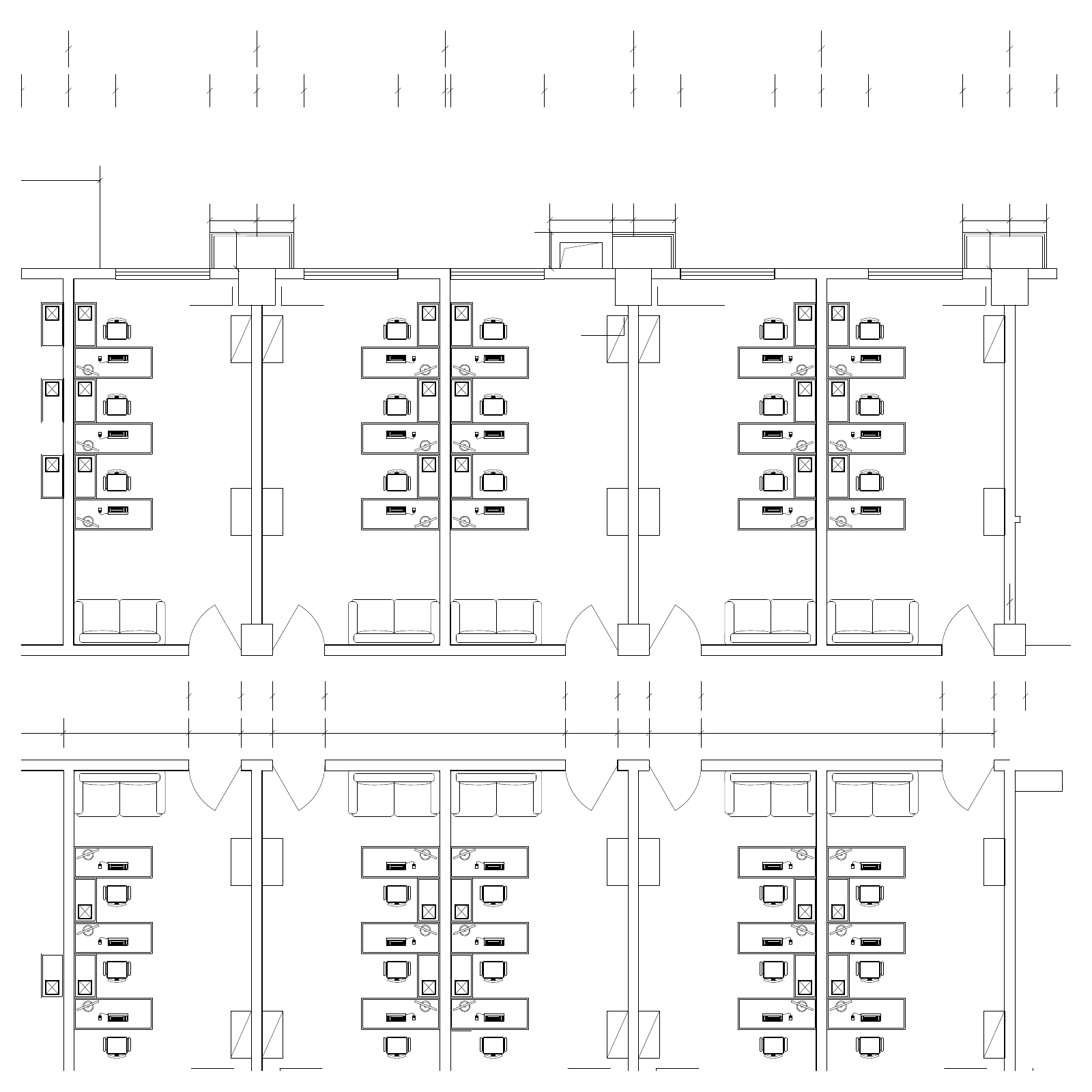} & \includegraphics[width=0.43\textwidth]{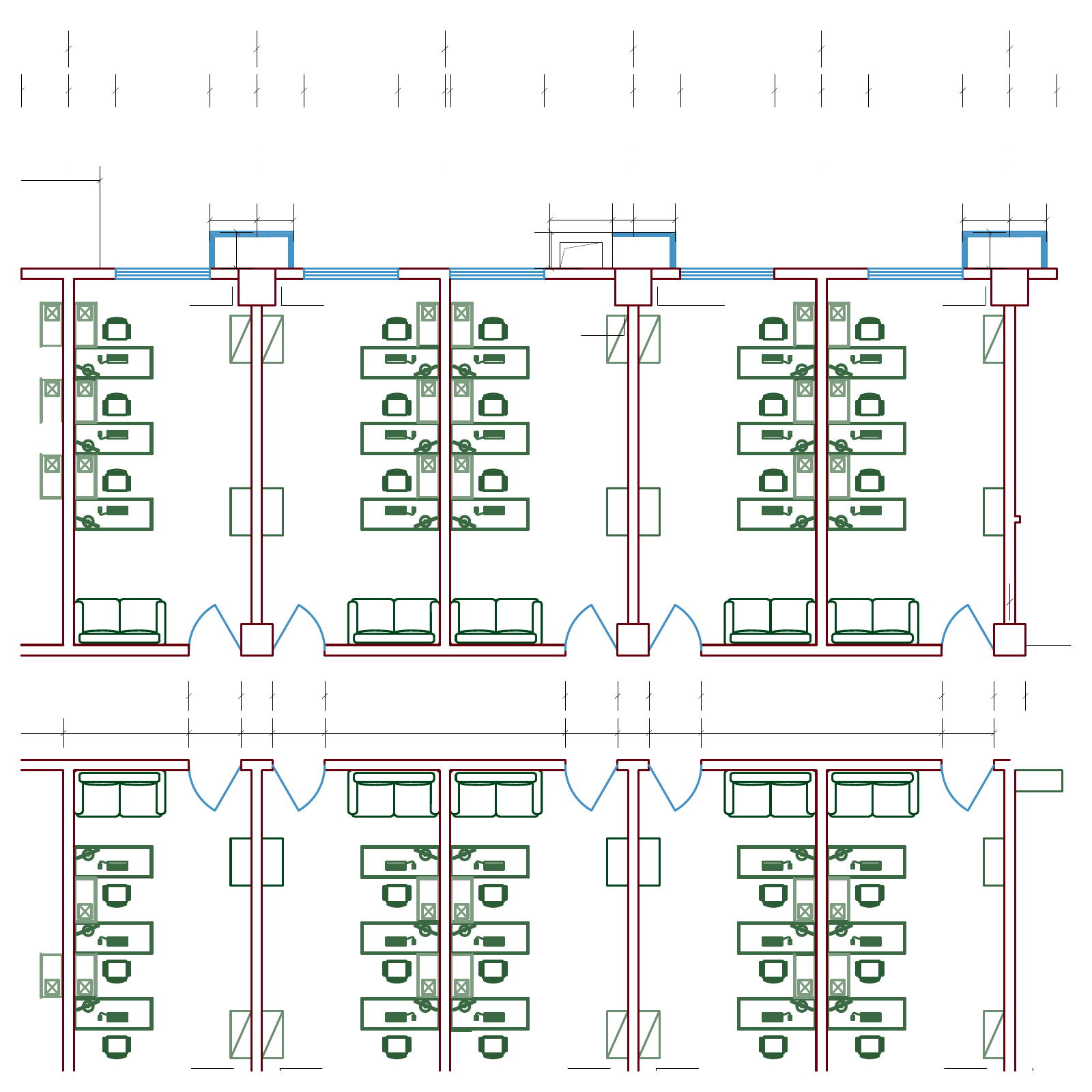} \\
    Raw Input & GT \\
\end{tabular}
\caption{Exemplary raw inputs and annotations in FloorPlanCAD, see the main manuscript for annotation details. The images are part of our \textit{train} set of \textit{underground parking lot} and \textit{office building} CAD drawings.}\label{fig:data_train_2}
\end{figure*}

We visualize several samples in train set of FloorPlanCAD dataset in Figure~\ref{fig:data_train_1} and Figure~\ref{fig:data_train_2} to demonstrate the variety of the proposed dataset.

\section{Per-Class Evaluation}

\subsection{Semantic Symbol Spotting}
The third column of Table~\ref{tab:per_class} shows semantic symbol spotting results of DeepLabv3+~\cite{chen2018encoder} and the proposed GCN-based method on all object classes. Here, we use \textit{weighted F1 score}  as the metric which use the entity length $\log(1 + L(e_{i}))$ to weight the \textit{TP}, \textit{FP} and \textit{FN}. We can see the GCN-based methods significantly outperforms DeepLabv3+ in \textit{wall} class since \textit{wall} class always mixes with thing classes.

\subsection{Instance Symbol Spotting}
We provide the class-wise mAP for Faster R-CNN~\cite{ren2016faster}, FCOS~\cite{tian2019fcos} and YOLOv3~\cite{redmon2018yolov3} in the fourth column of Table~\ref{tab:per_class}. The results includes 28 thing classes. We can notice Faster R-CNN is comparable with FCOS and both outperform YOLOv3 which may caused by our dataset contains various scenes, 28 possible symbol classes and complex background.

\subsection{Panoptic Symbol Spotting Results}
\begin{figure*}
\centering
\begin{tabular}{c|c|c}
    \includegraphics[width=0.3\textwidth]{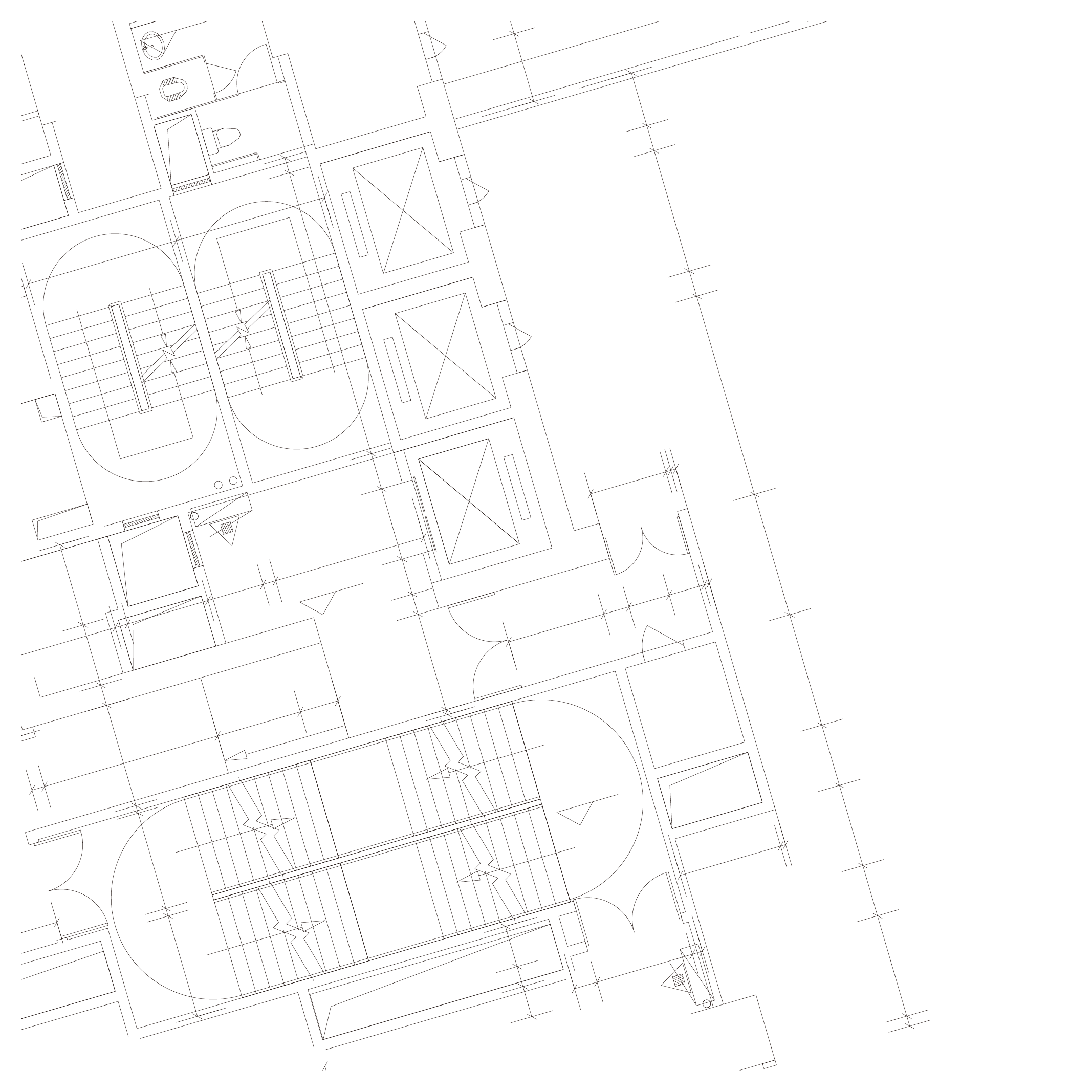} & \includegraphics[width=0.3\textwidth]{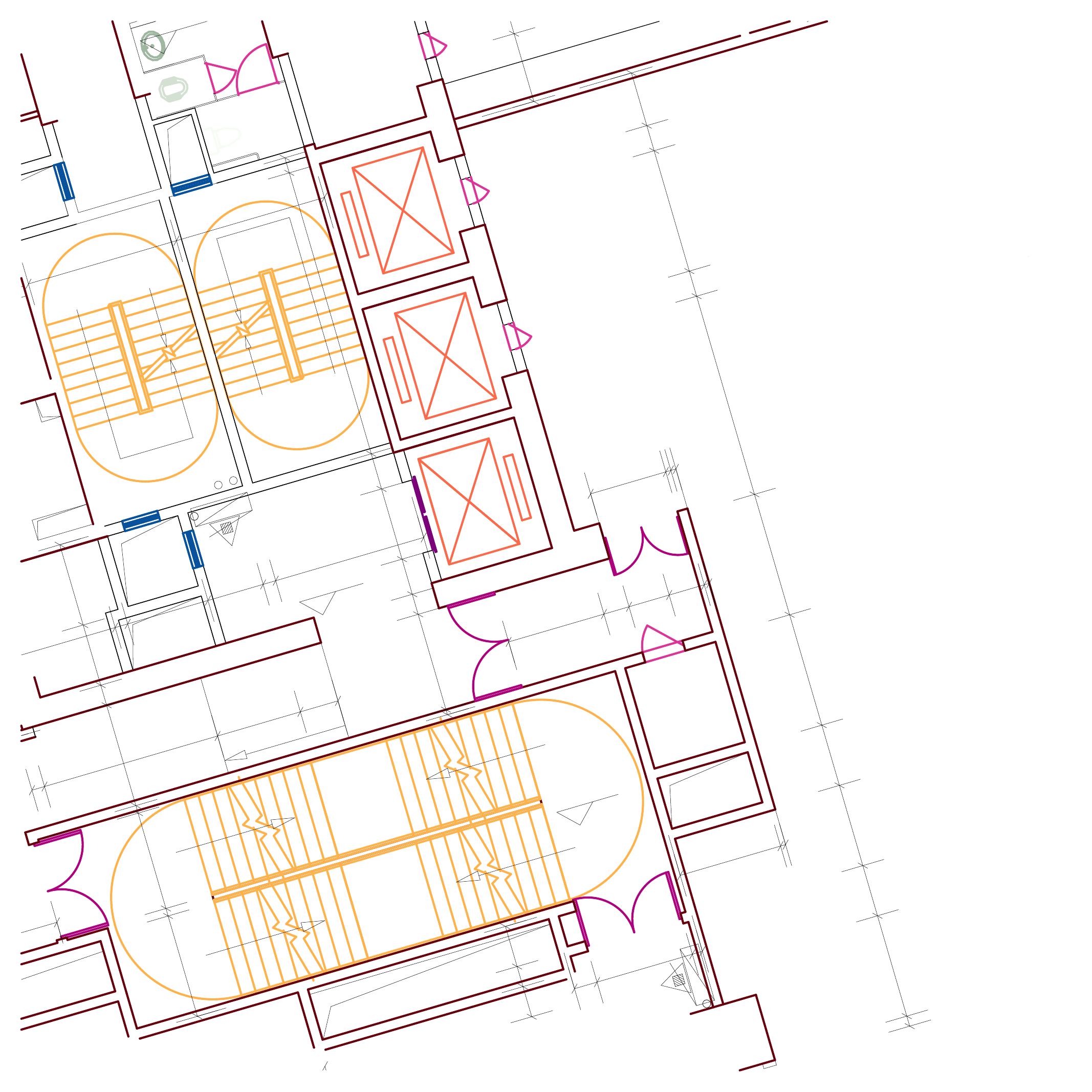} & \includegraphics[width=0.3\textwidth]{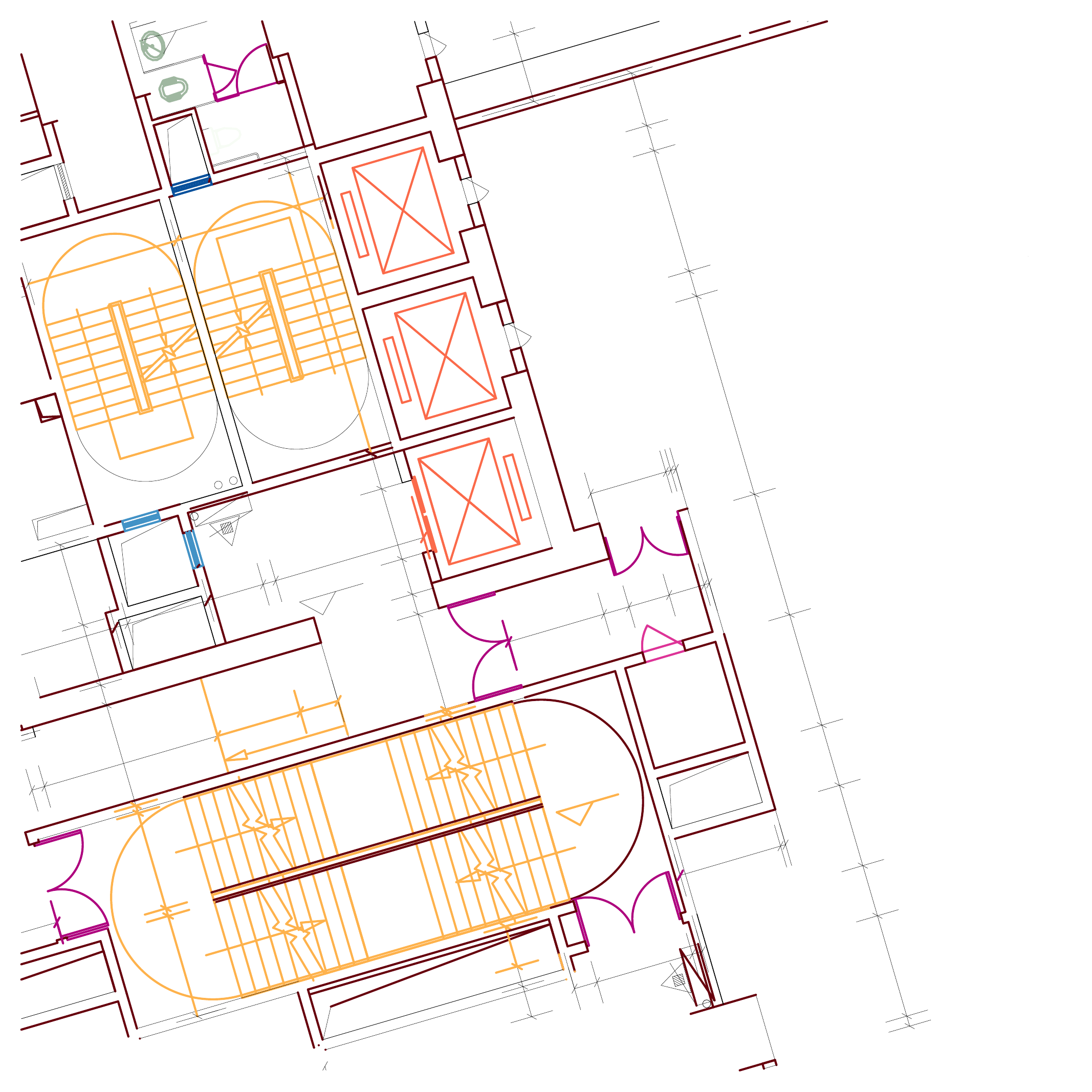} \\ \hline
    \includegraphics[width=0.3\textwidth]{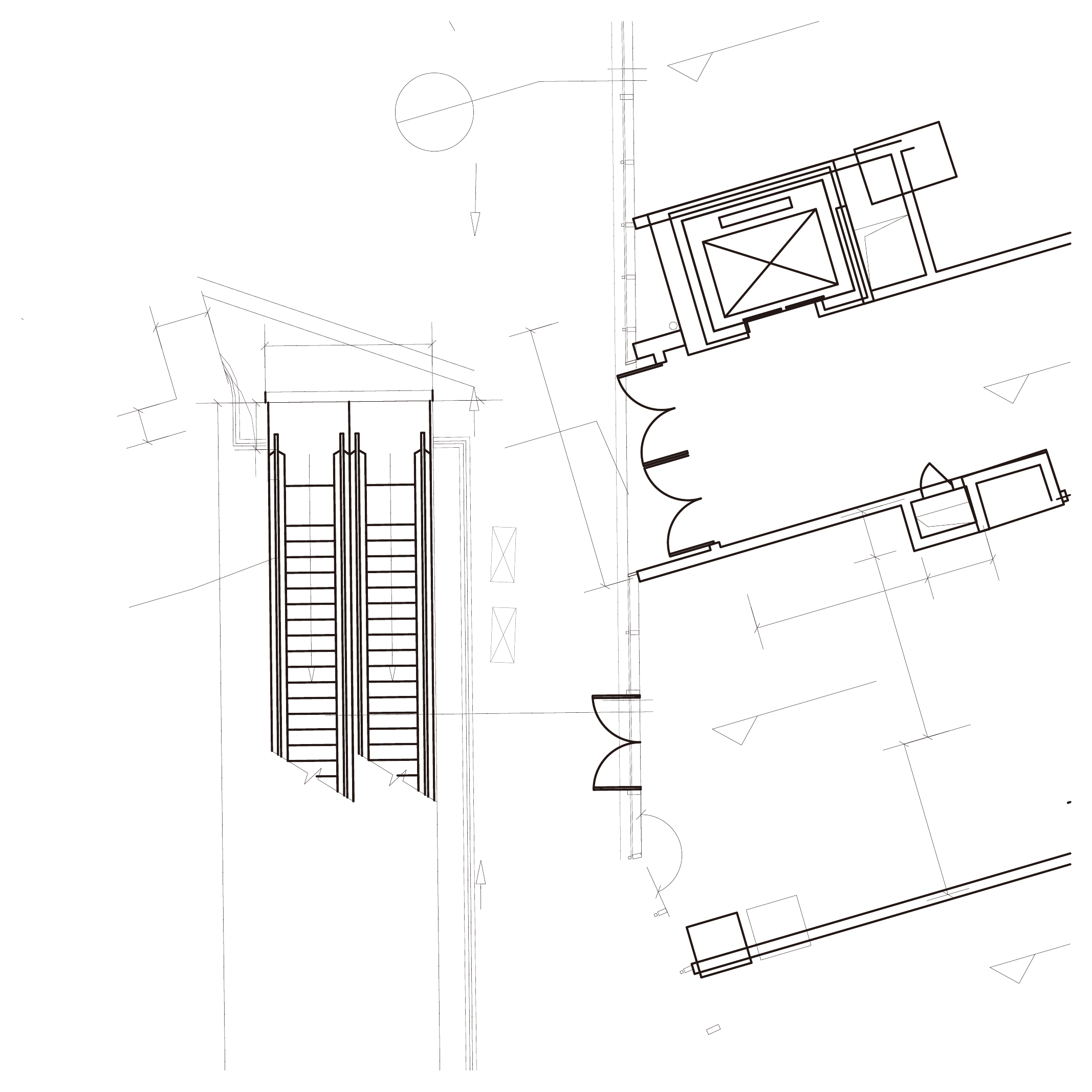} & \includegraphics[width=0.3\textwidth]{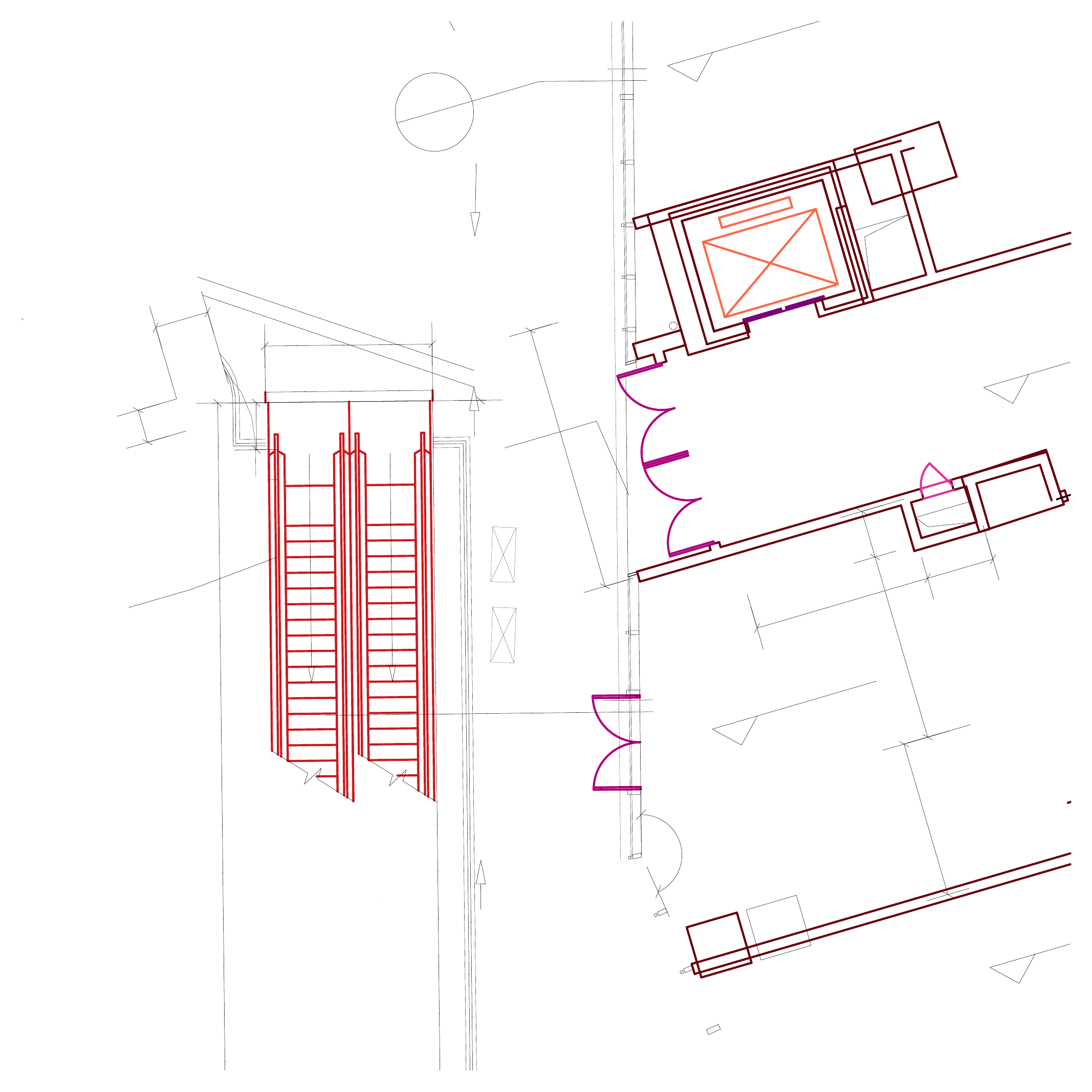} & \includegraphics[width=0.3\textwidth]{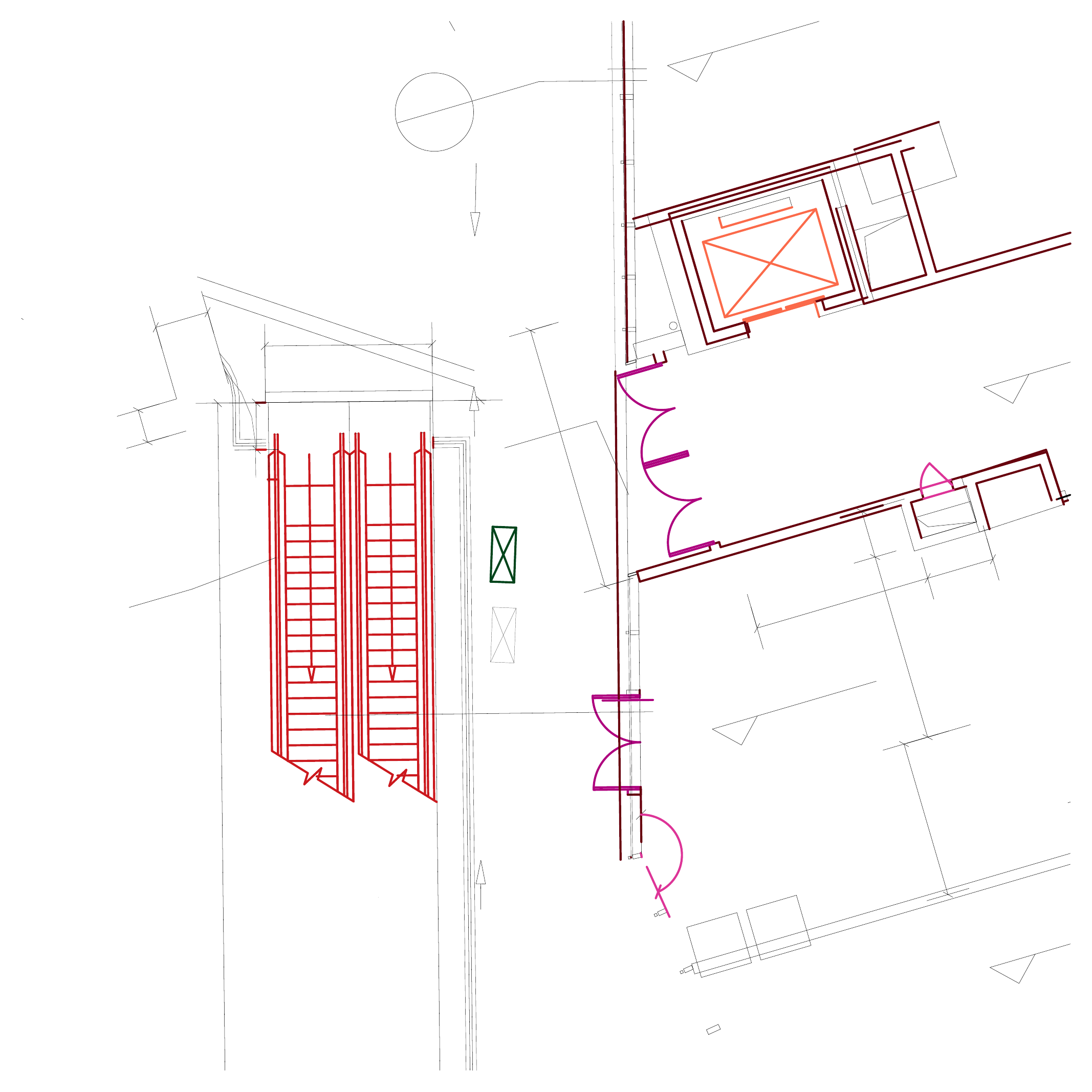} \\ \hline
    \includegraphics[width=0.3\textwidth]{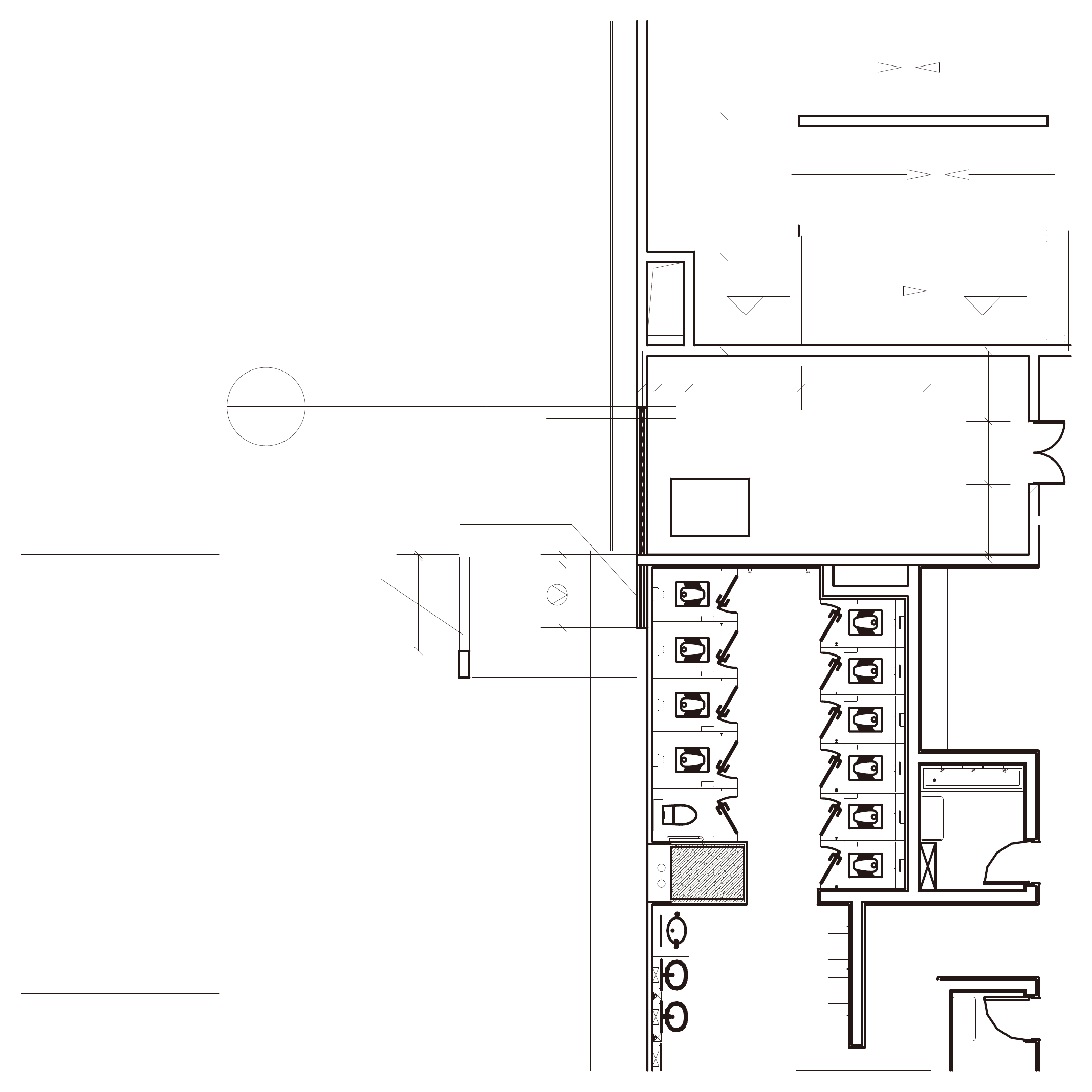} & \includegraphics[width=0.3\textwidth]{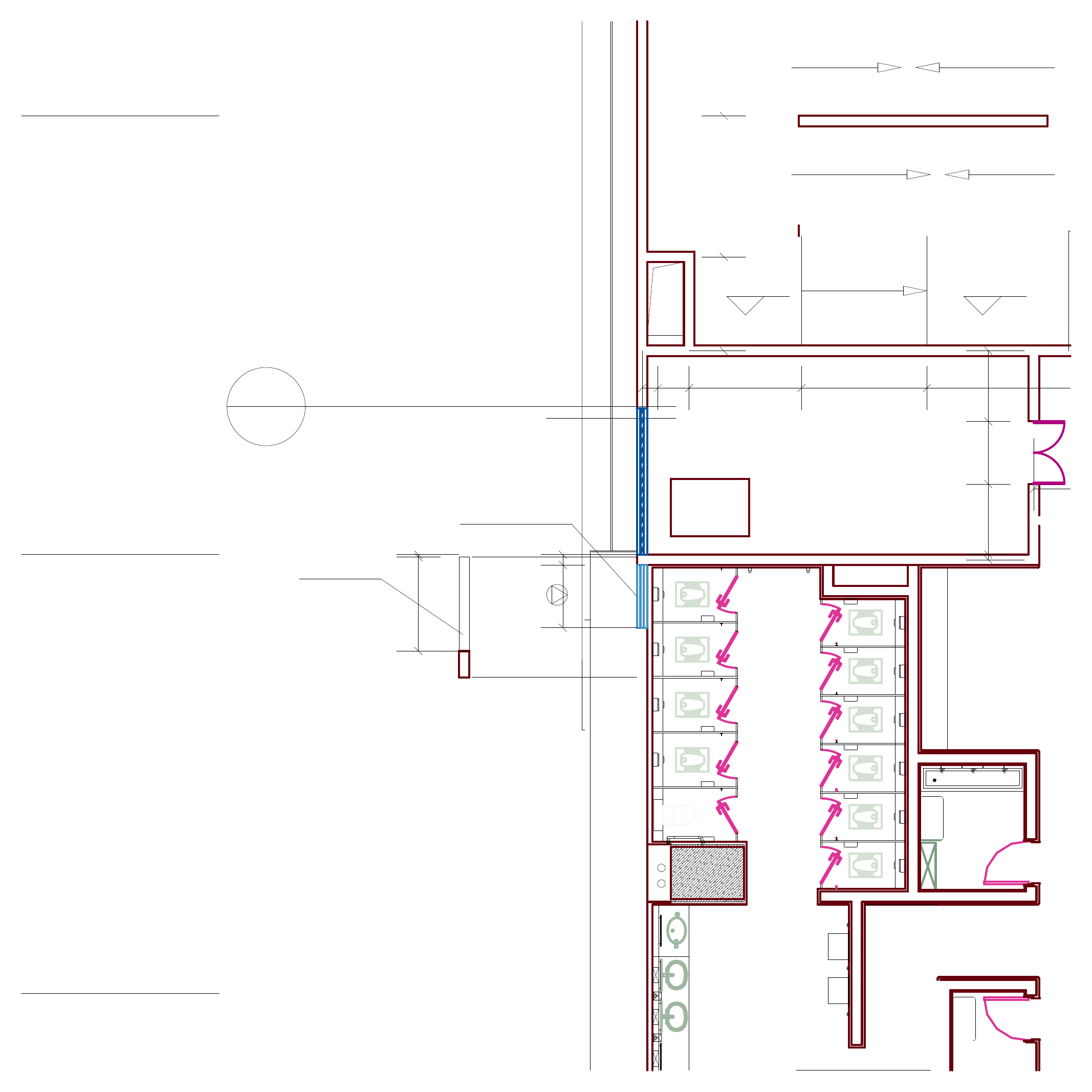} & \includegraphics[width=0.3\textwidth]{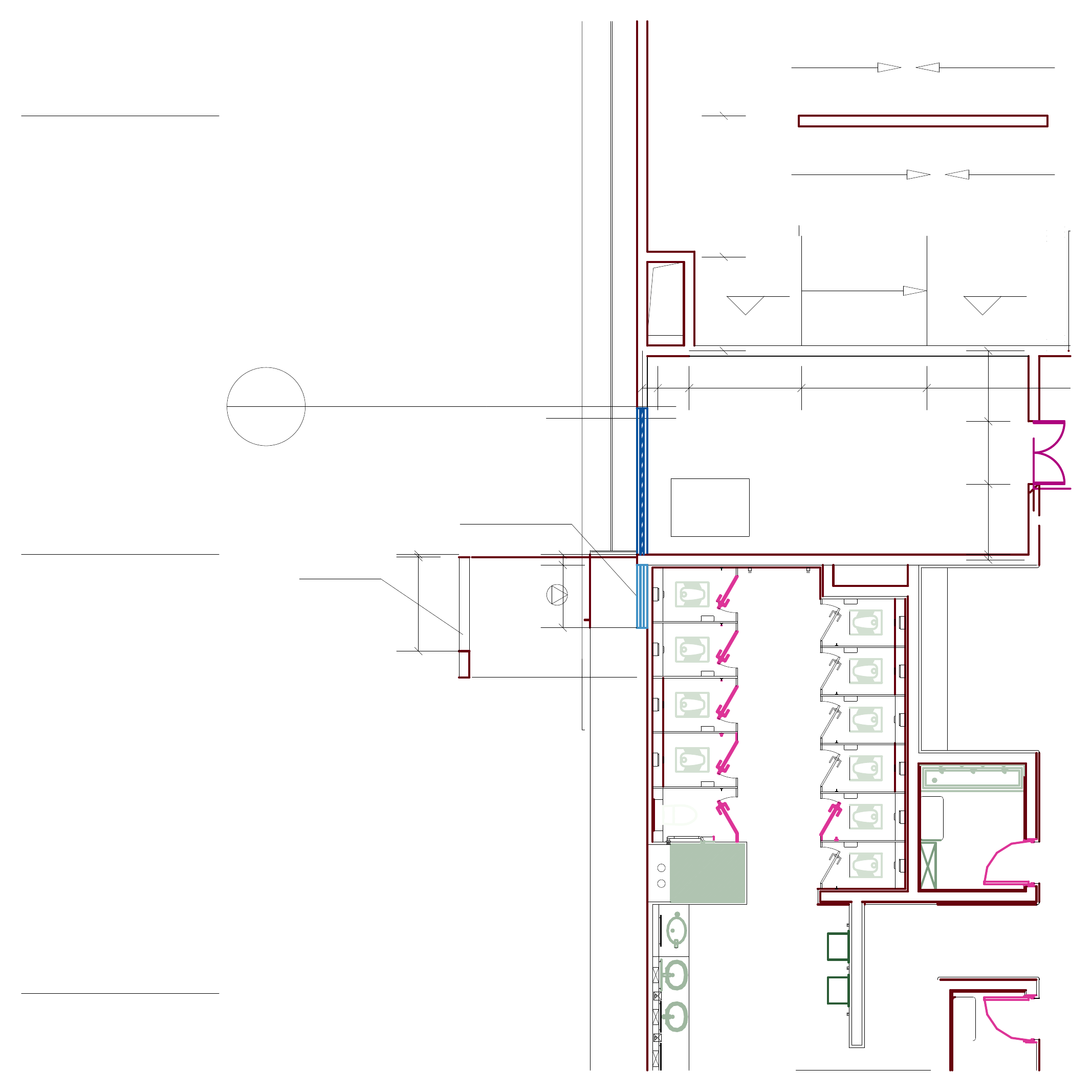} \\ \hline
    \includegraphics[width=0.3\textwidth]{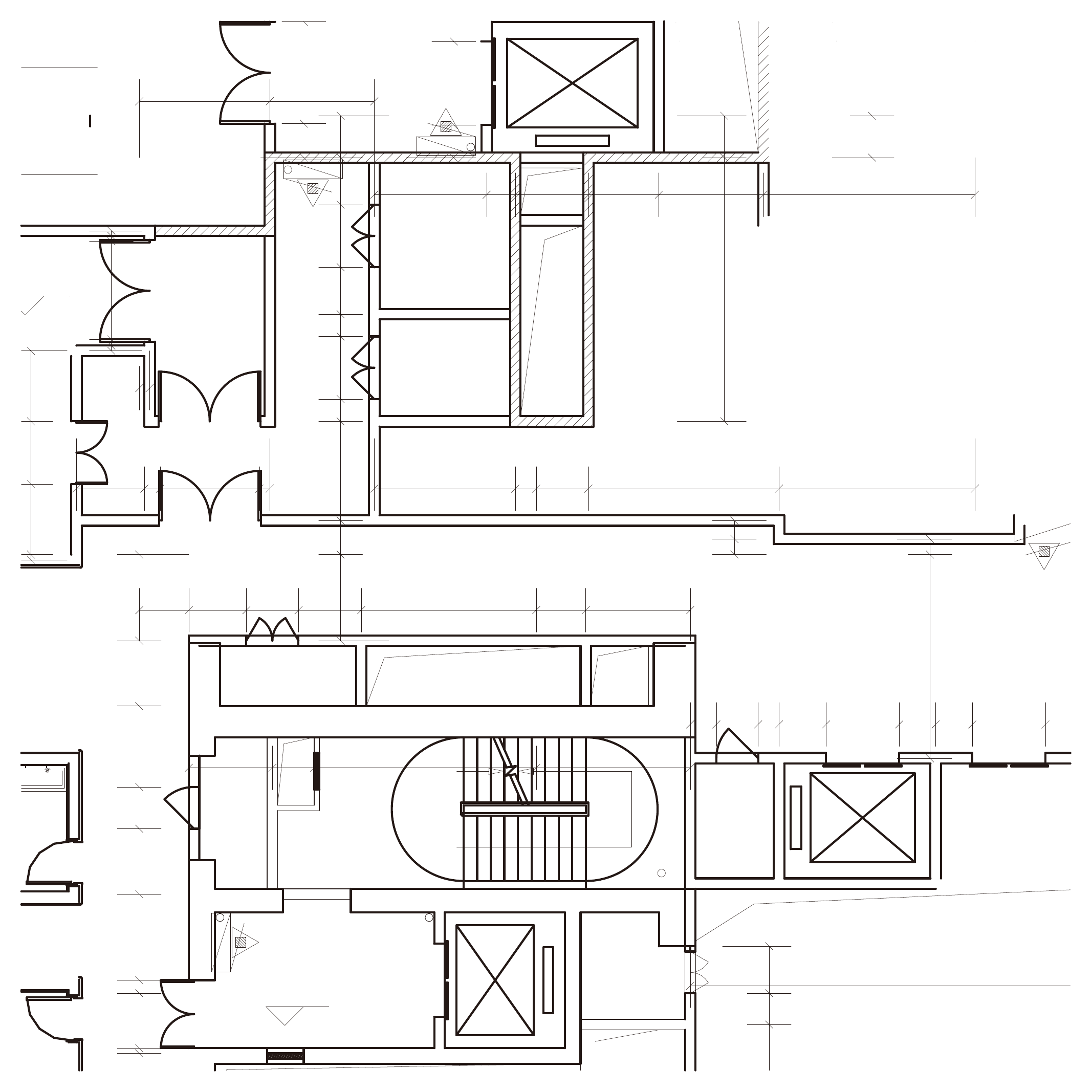} & \includegraphics[width=0.3\textwidth]{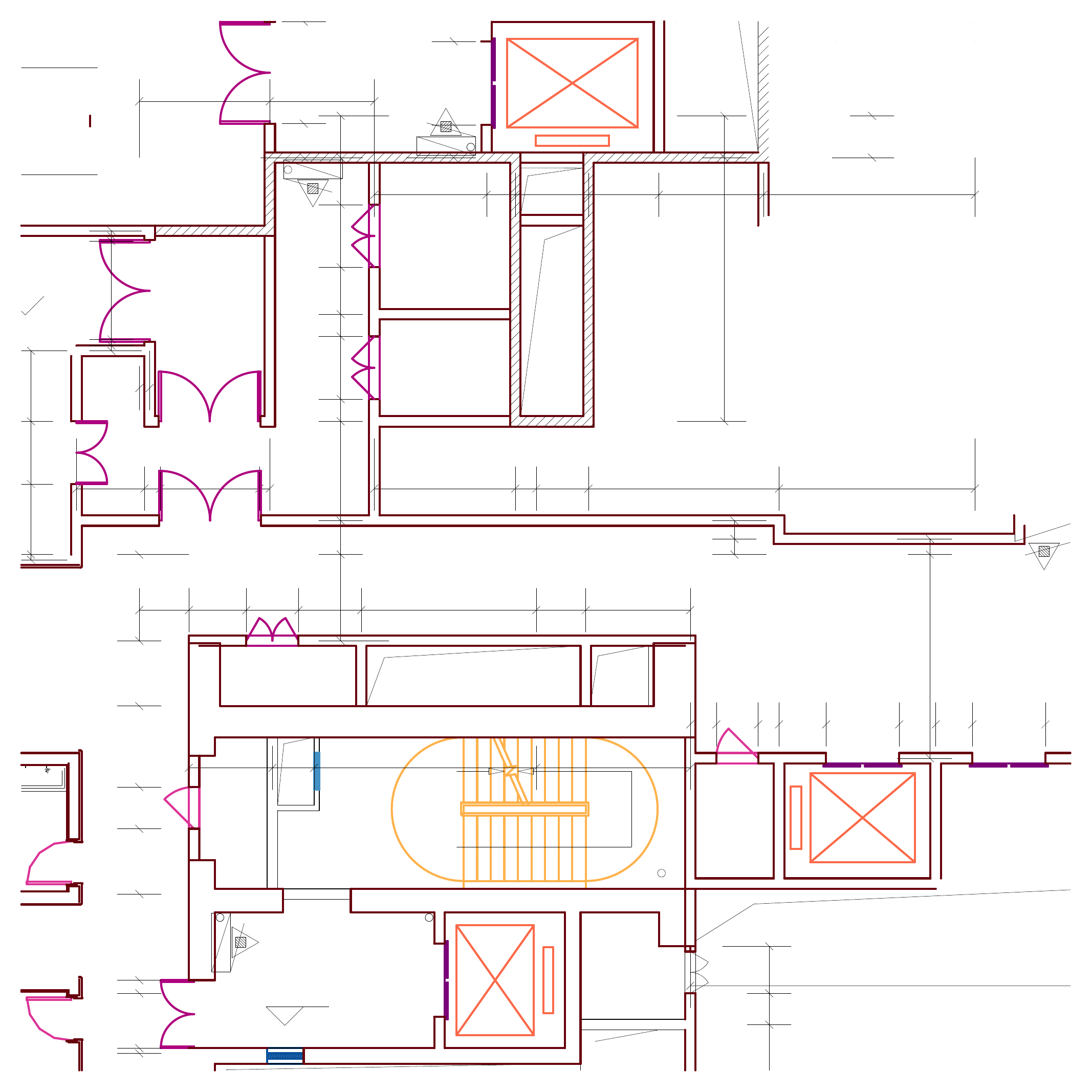} & \includegraphics[width=0.3\textwidth]{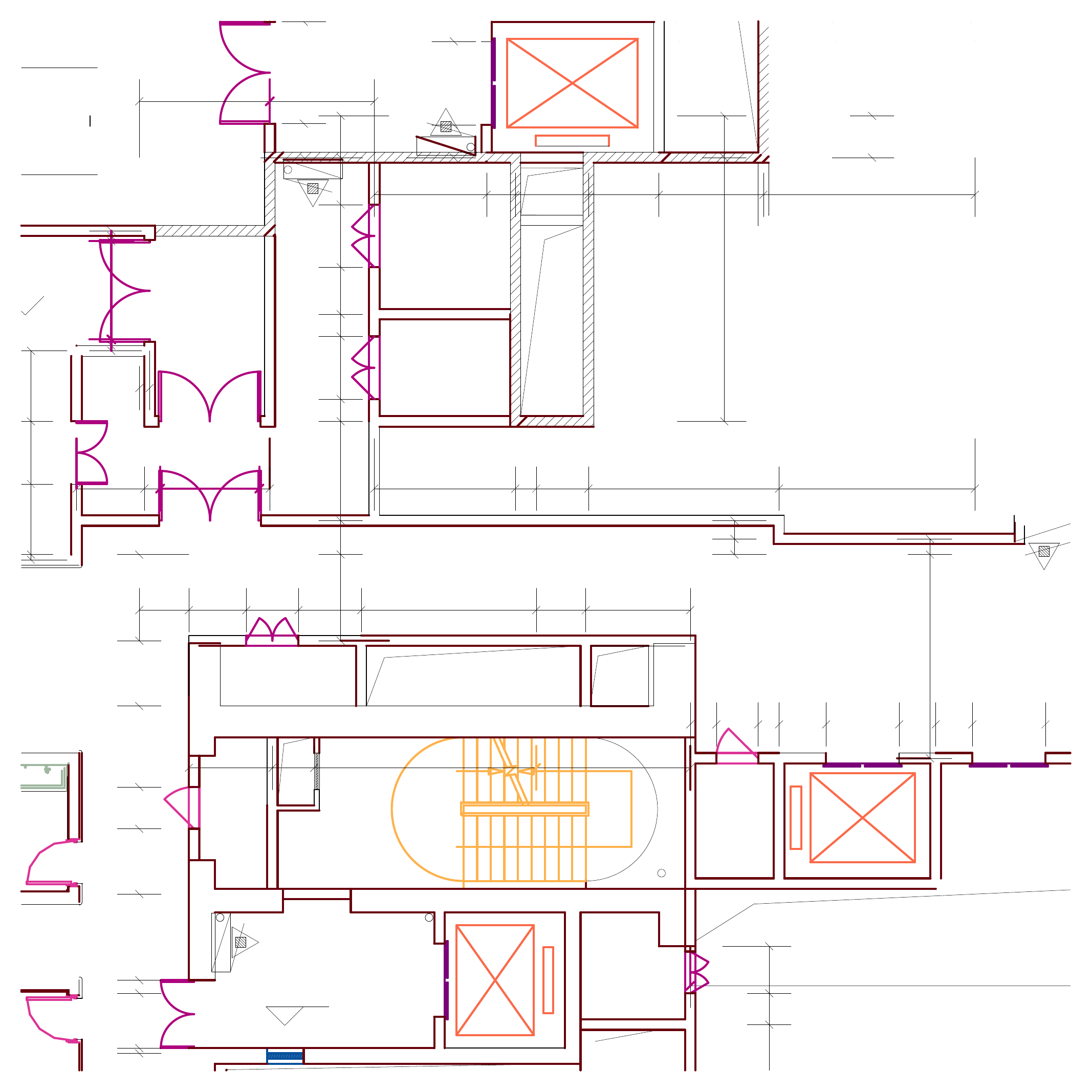} \\
    Raw Input & GT & Panoptic Prediction \\
\end{tabular}
\caption{Results of PanCADNet on FloorPlanCAD, see the main manuscript for annotation details. The images are part of our \textit{test} set of \textit{large shopping mall} CAD drawings.}\label{fig:pan_results_1}
\end{figure*}

\begin{figure*}
\centering
\begin{tabular}{c|c|c}
    \includegraphics[width=0.3\textwidth]{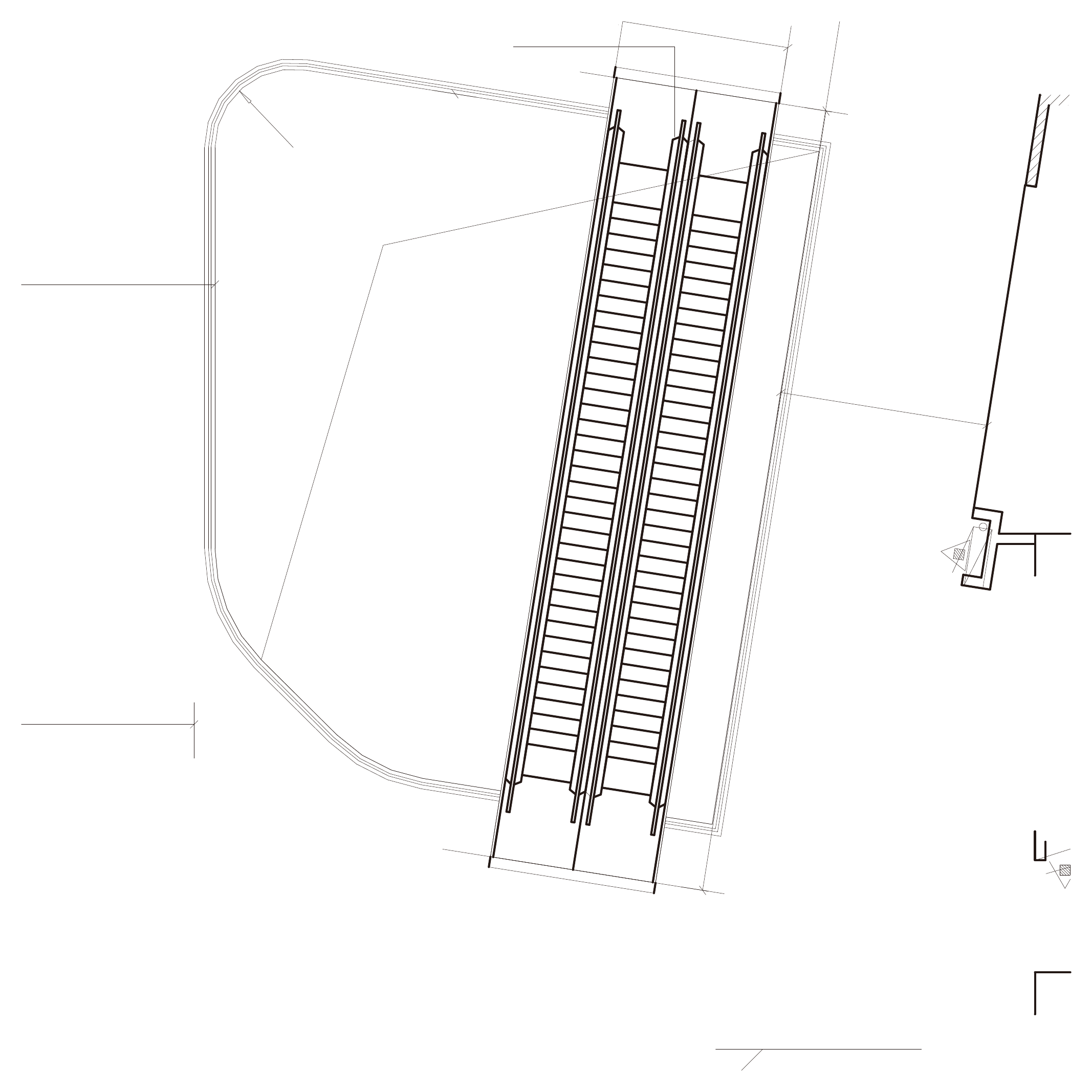} & \includegraphics[width=0.3\textwidth]{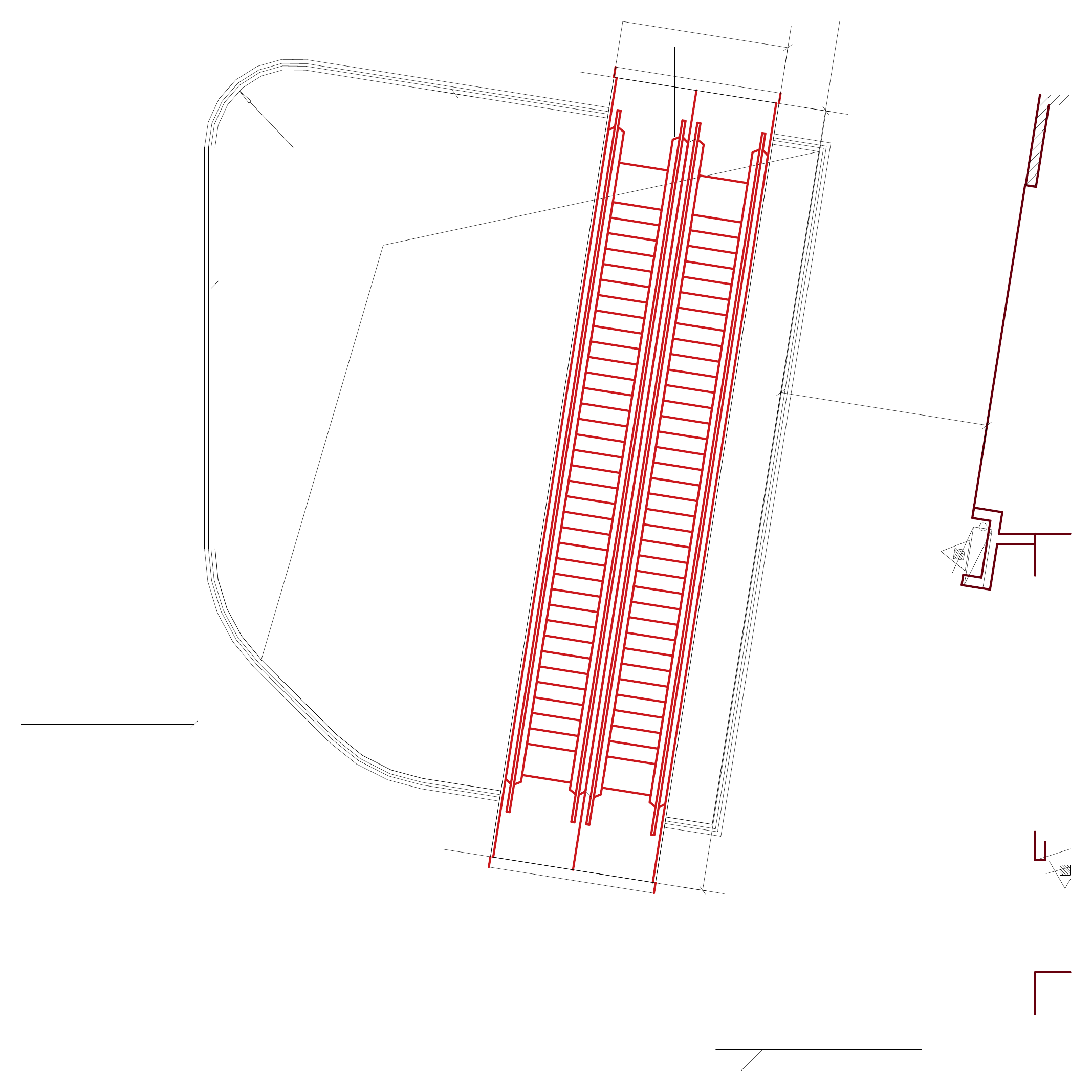} & \includegraphics[width=0.3\textwidth]{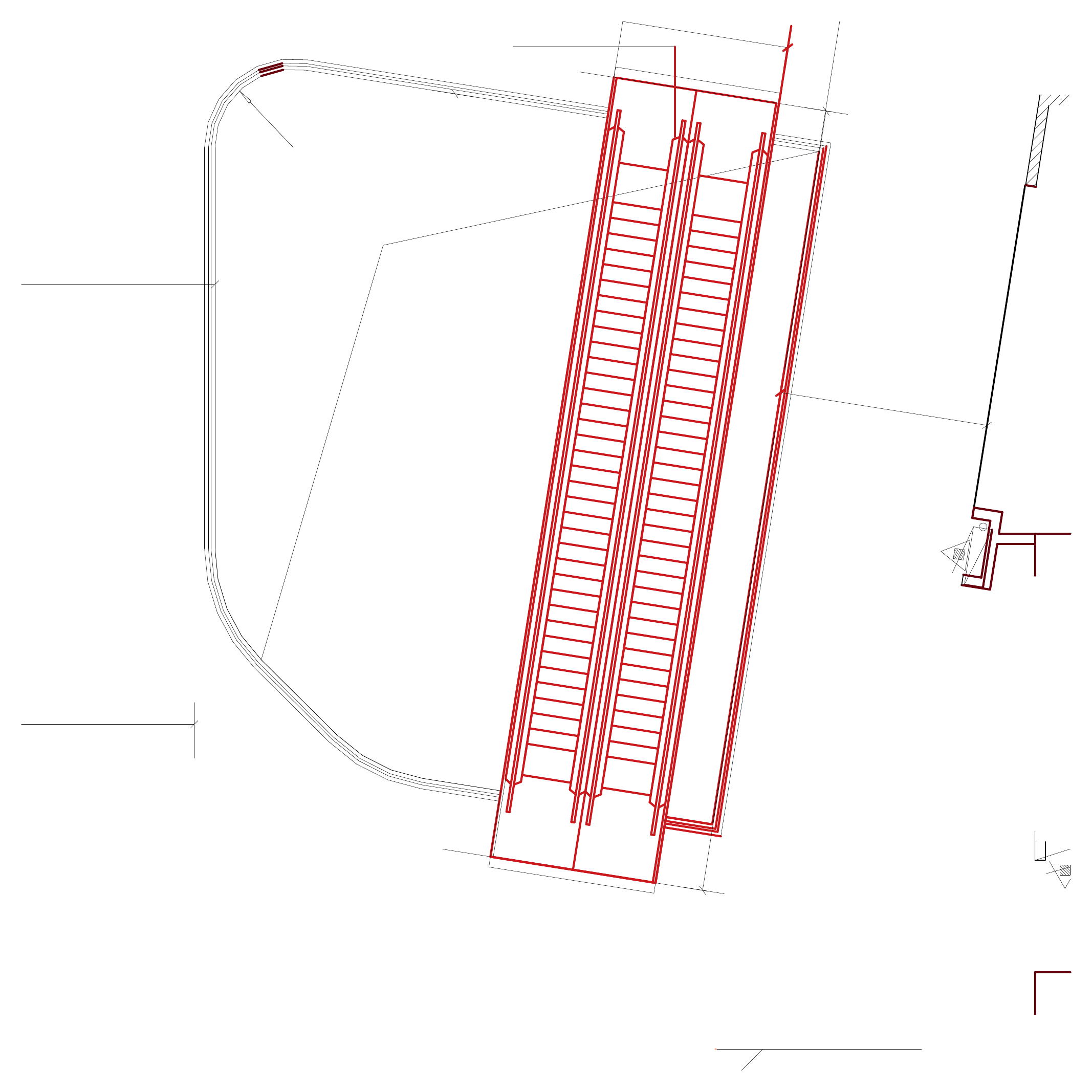} \\ \hline
    \includegraphics[width=0.3\textwidth]{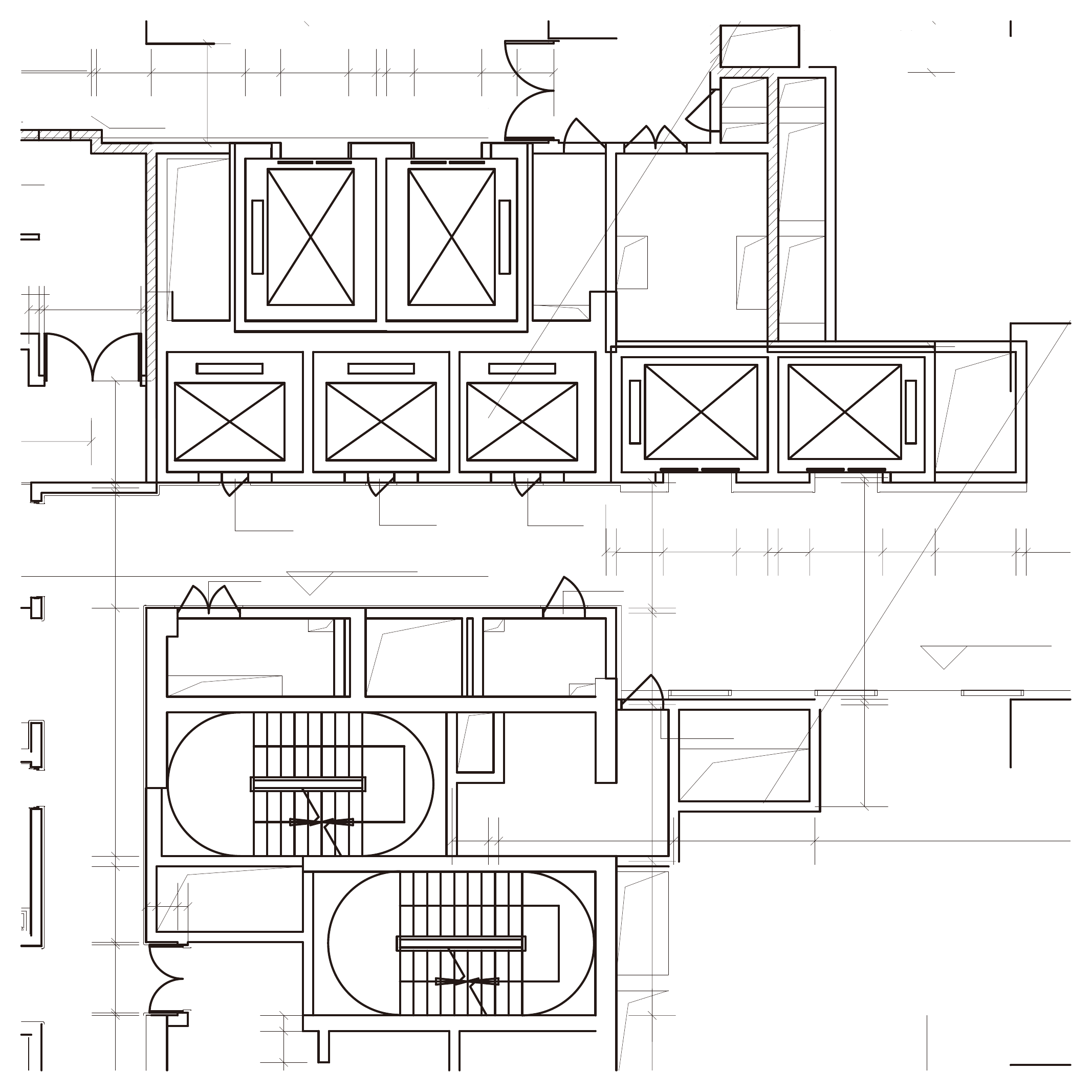} & \includegraphics[width=0.3\textwidth]{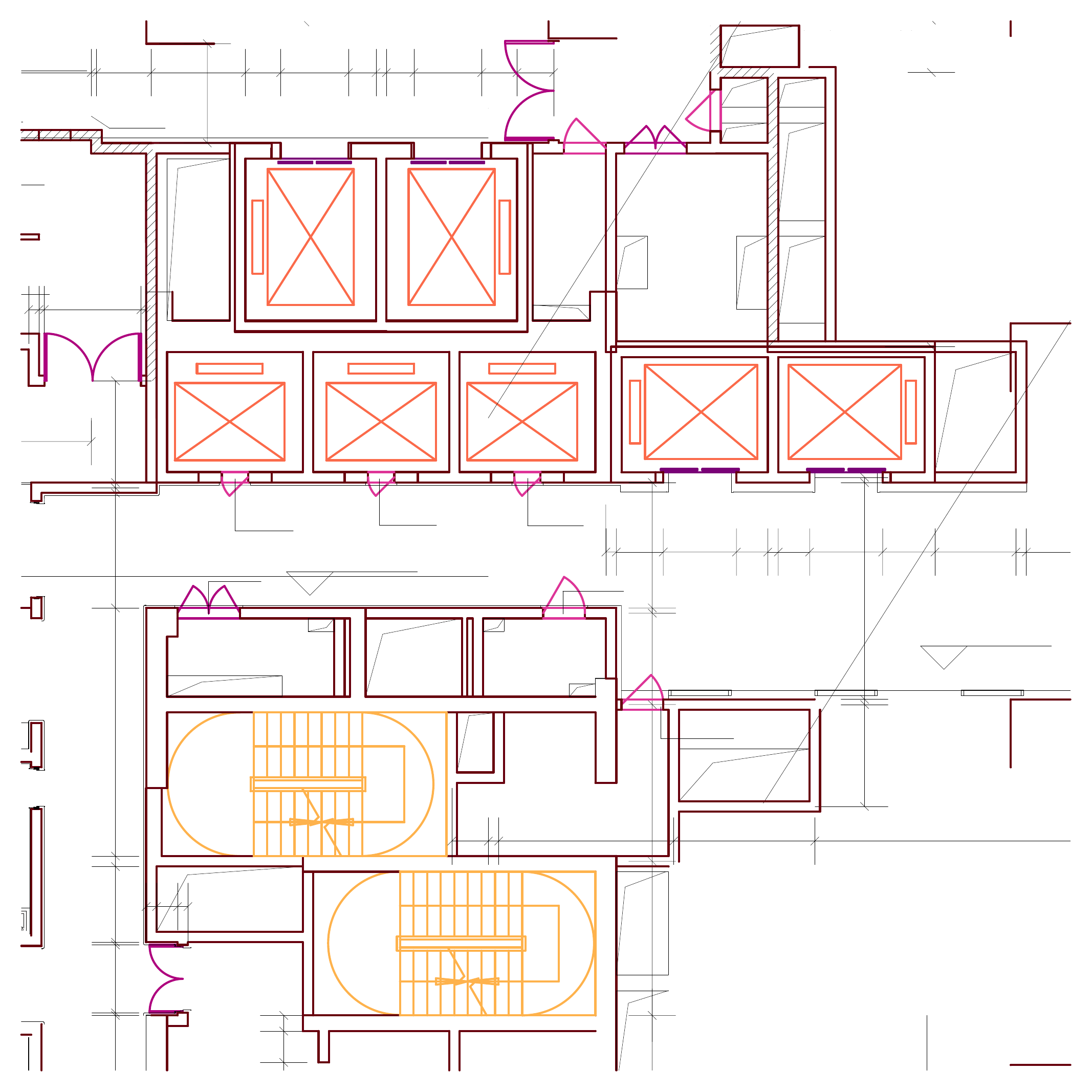} & \includegraphics[width=0.3\textwidth]{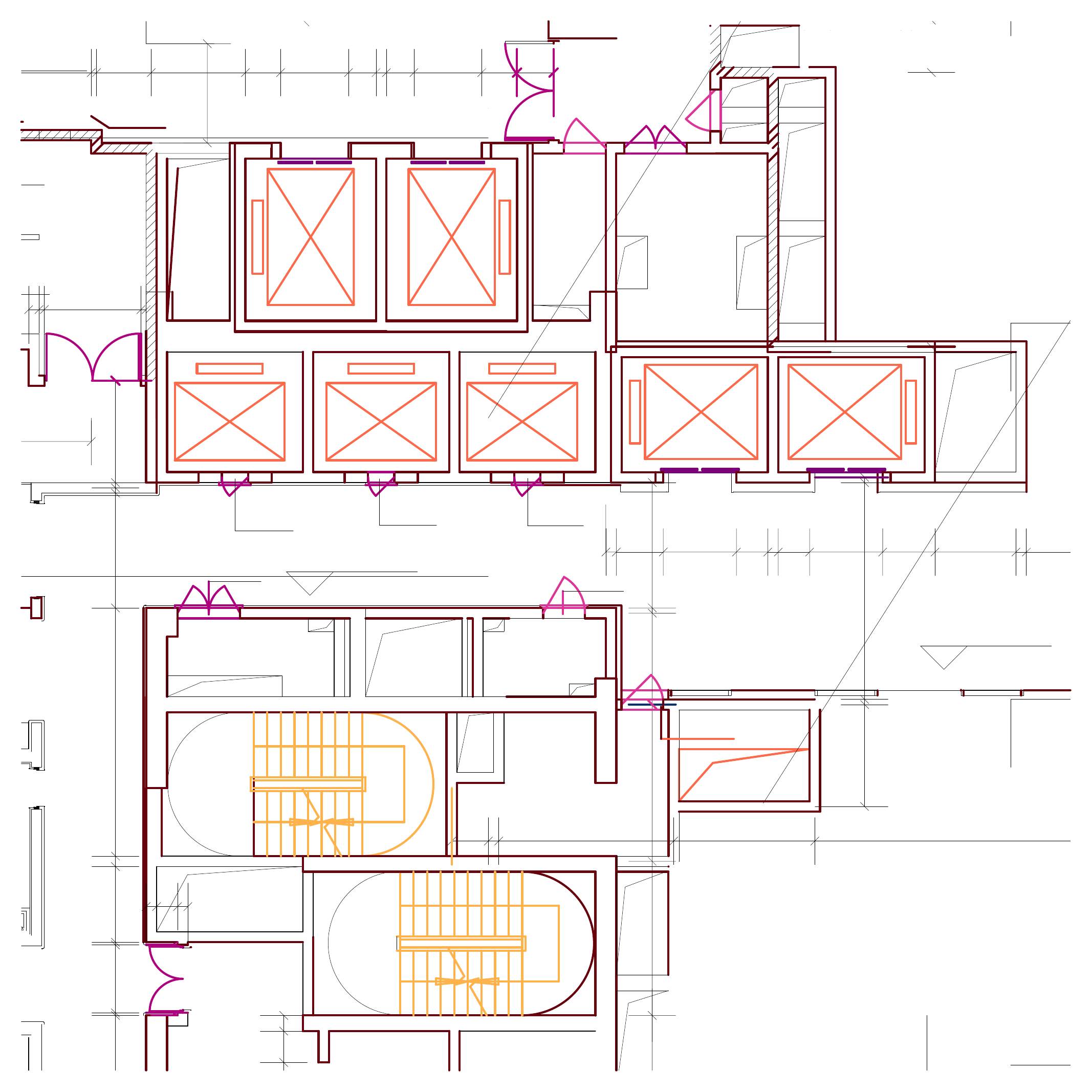} \\ \hline
    \includegraphics[width=0.3\textwidth]{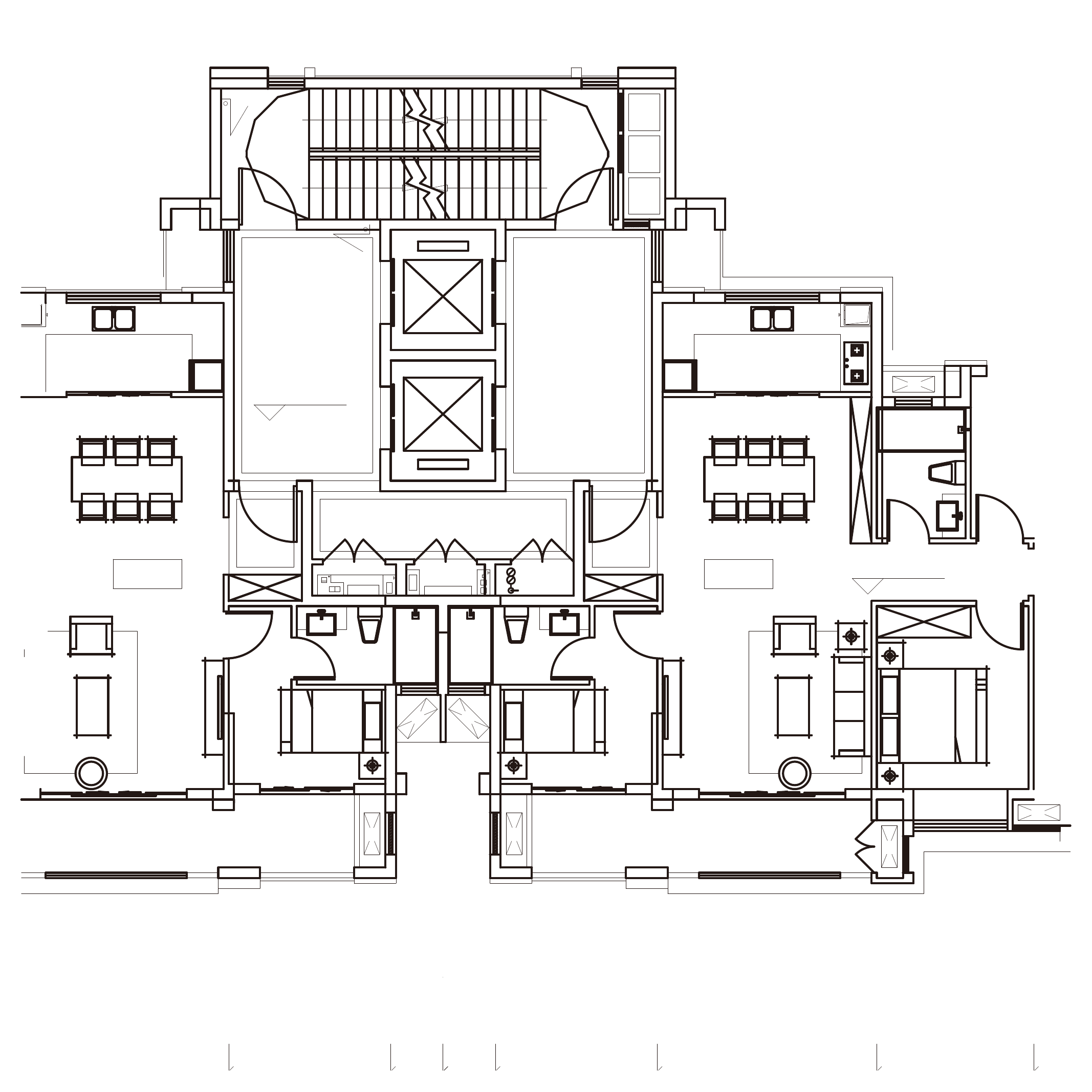} & \includegraphics[width=0.3\textwidth]{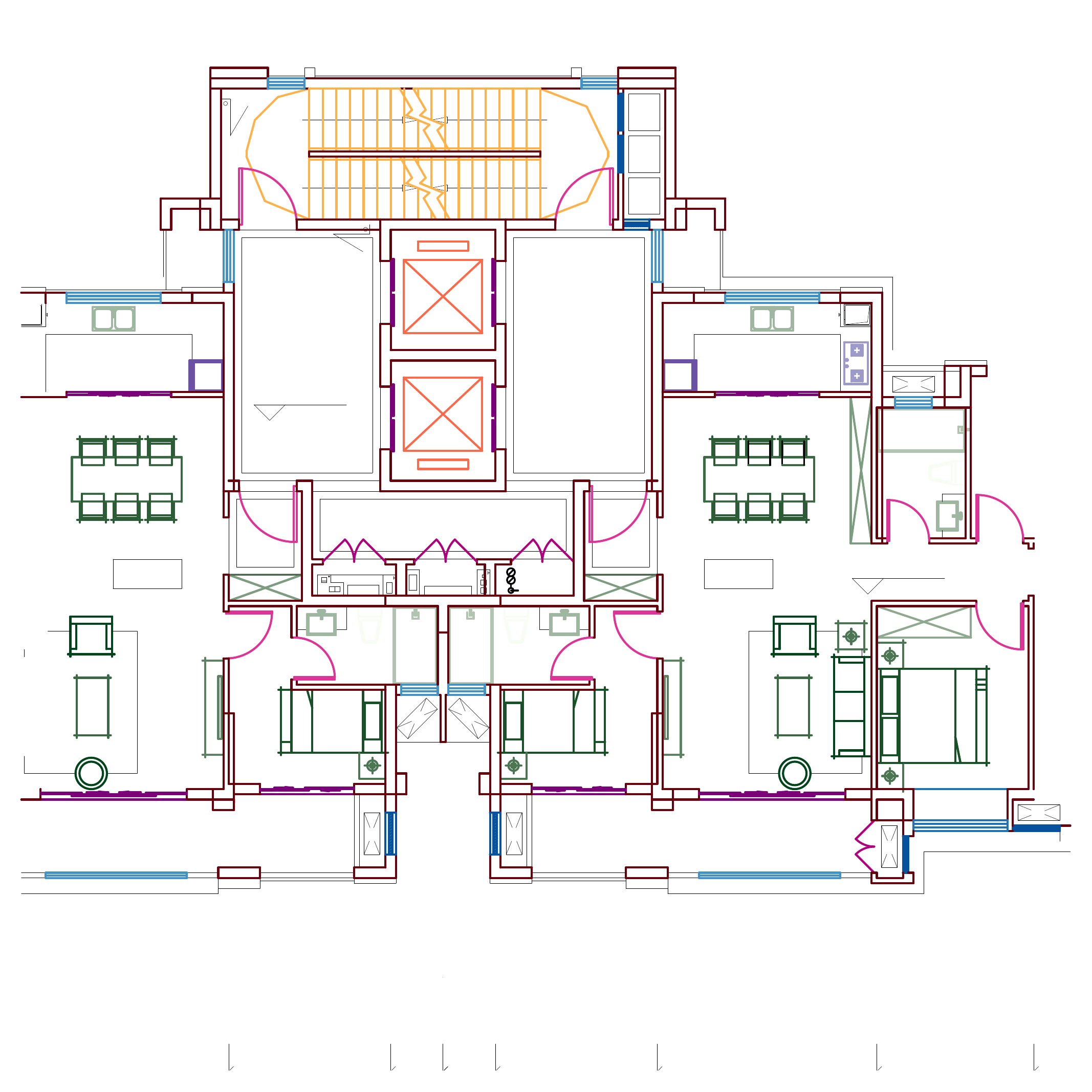} & \includegraphics[width=0.3\textwidth]{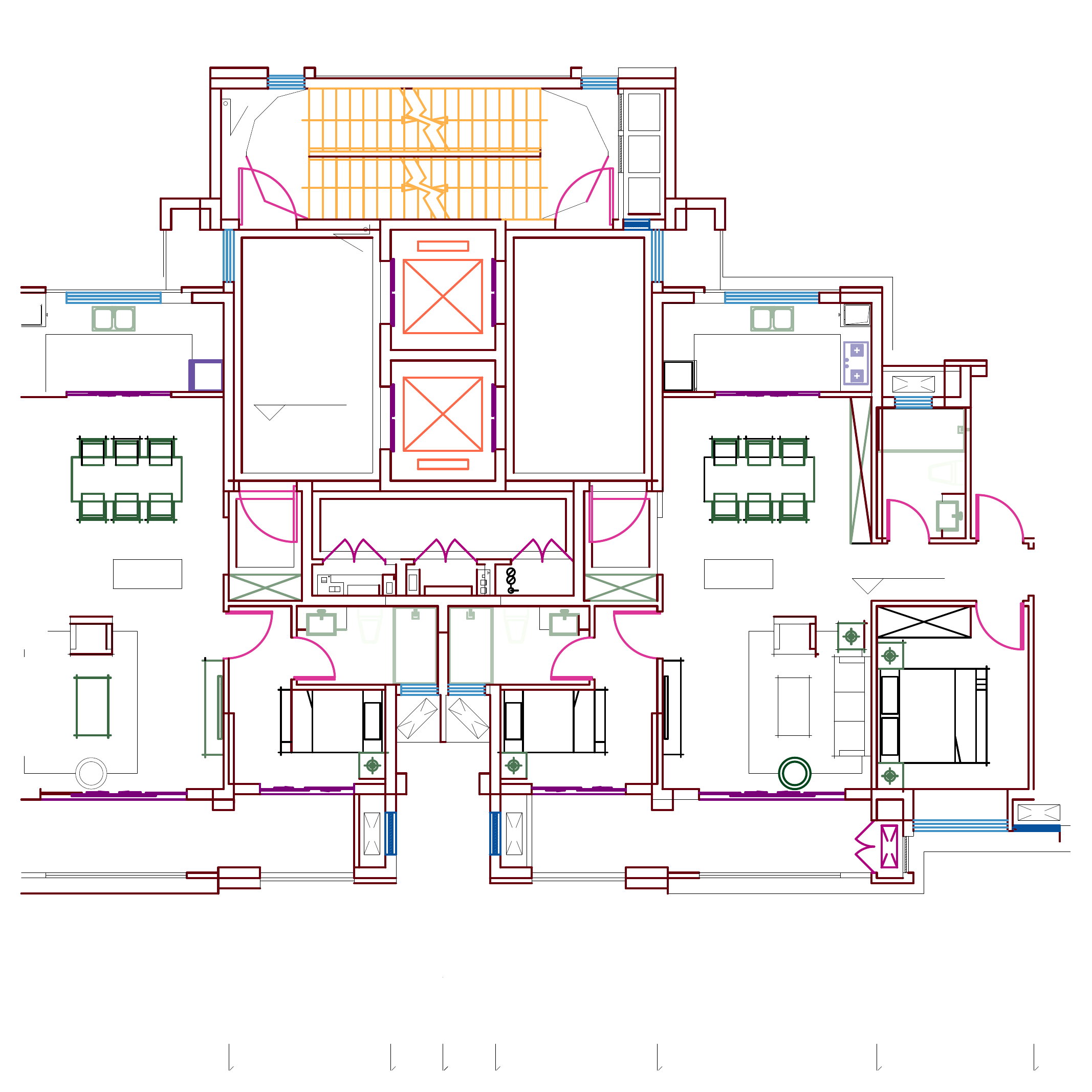} \\ \hline
    \includegraphics[width=0.3\textwidth]{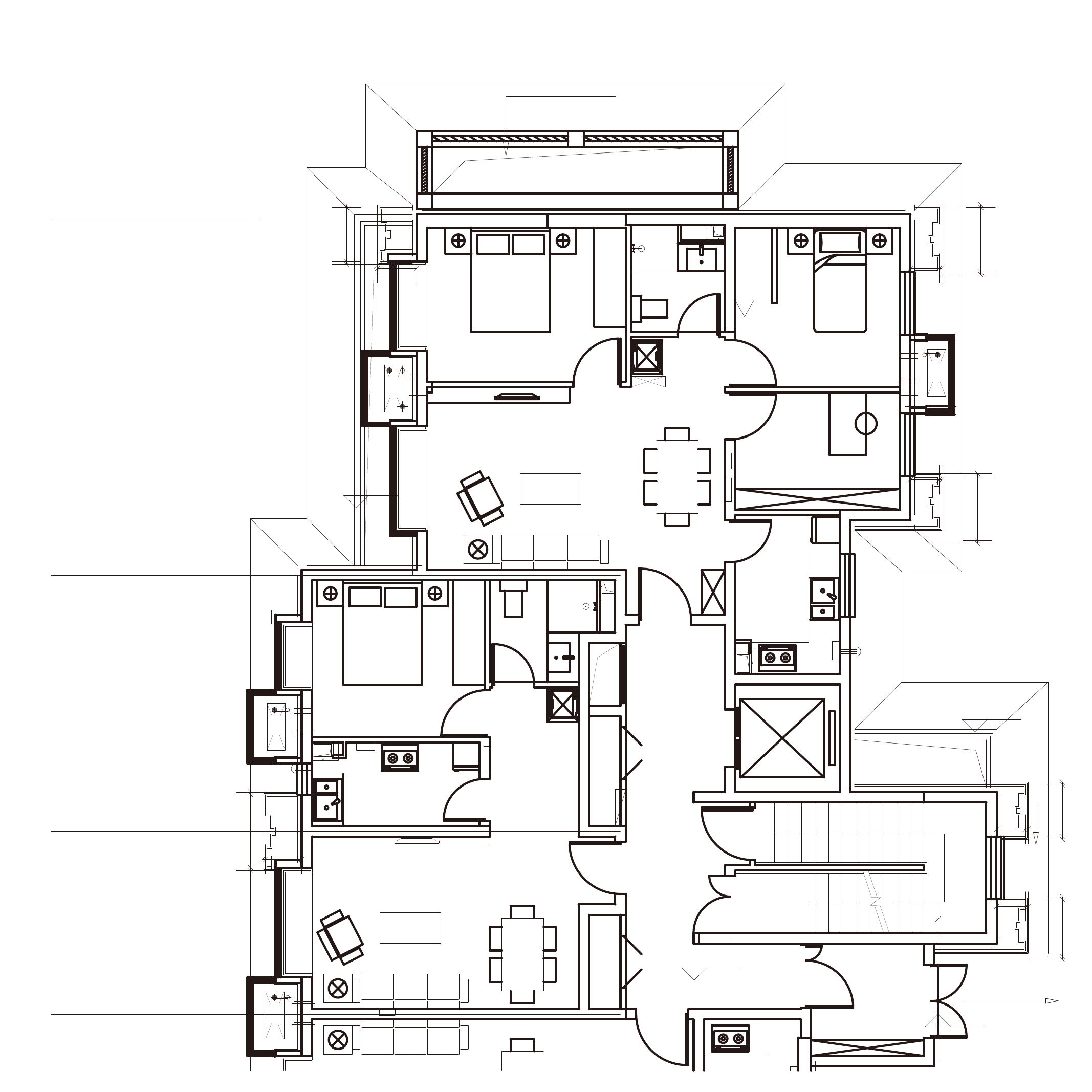} & \includegraphics[width=0.3\textwidth]{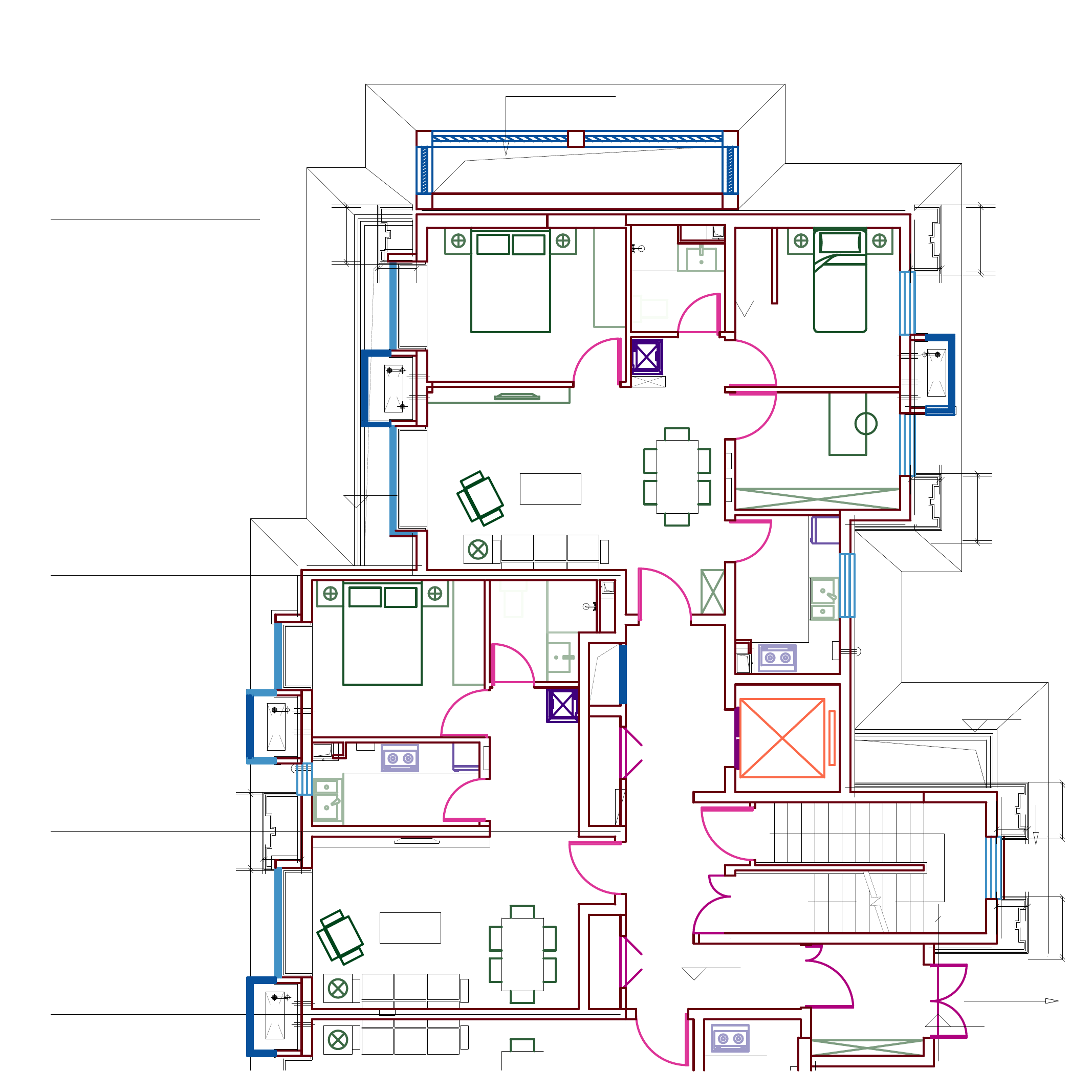} & \includegraphics[width=0.3\textwidth]{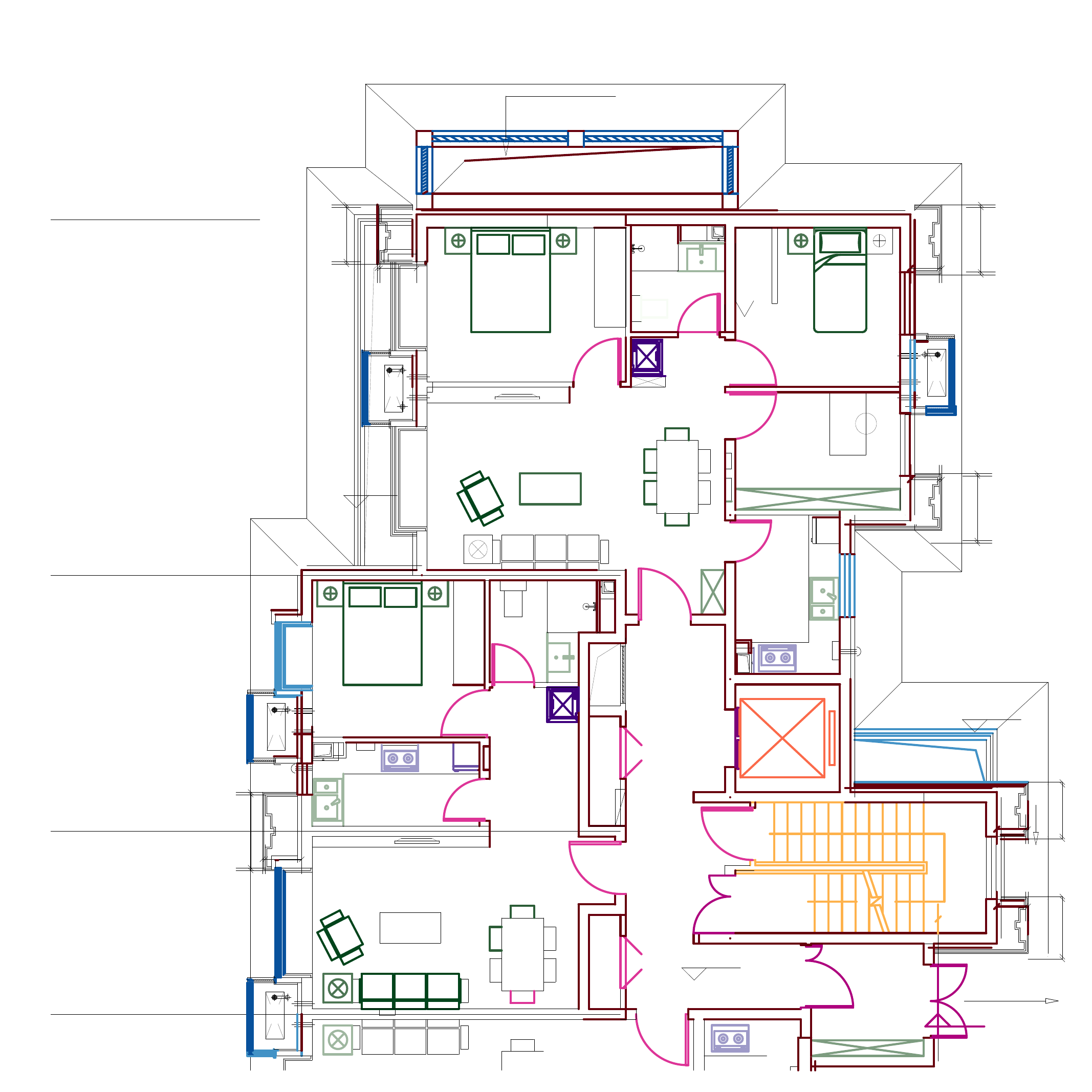} \\
    Raw Input & GT & Panoptic Prediction \\
\end{tabular}
\caption{Results of PanCADNet on FloorPlanCAD, see the main manuscript for annotation details. The images are part of our \textit{test} set of \textit{large shopping mall} and \textit{residential building} CAD drawings.}\label{fig:pan_results_2}
\end{figure*}

\begin{figure*}
\centering
\begin{tabular}{c|c|c}
    \includegraphics[width=0.3\textwidth]{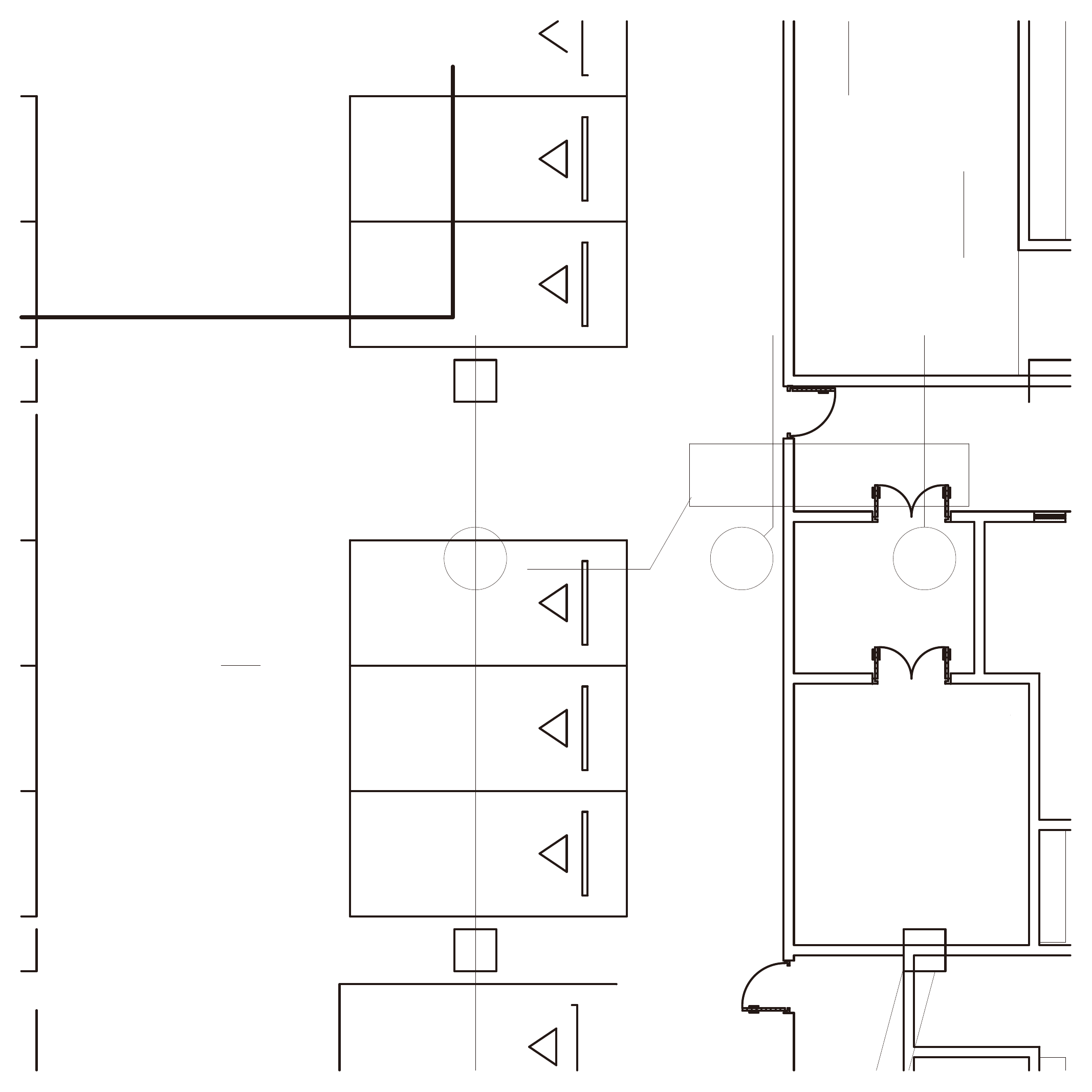} & \includegraphics[width=0.3\textwidth]{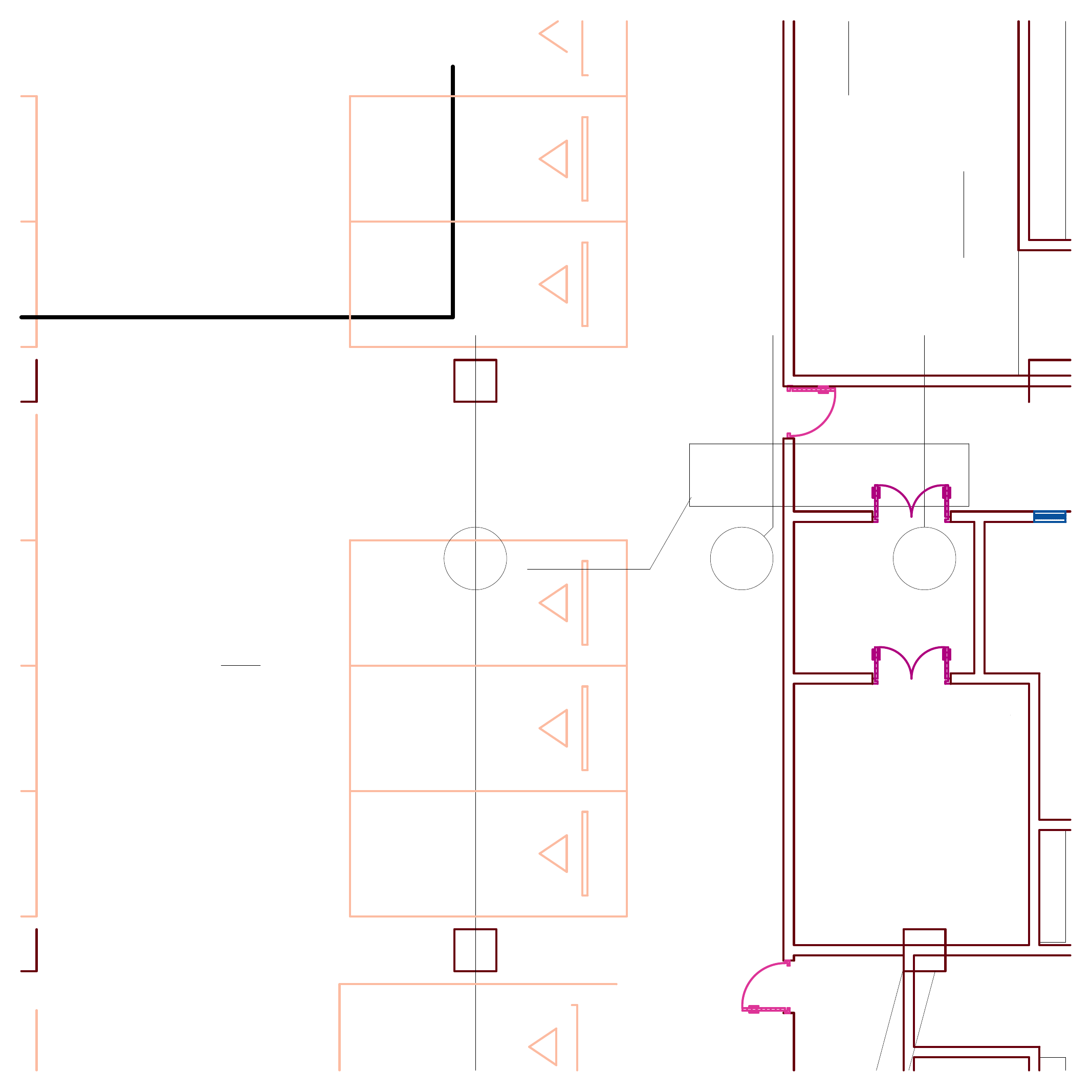} & \includegraphics[width=0.3\textwidth]{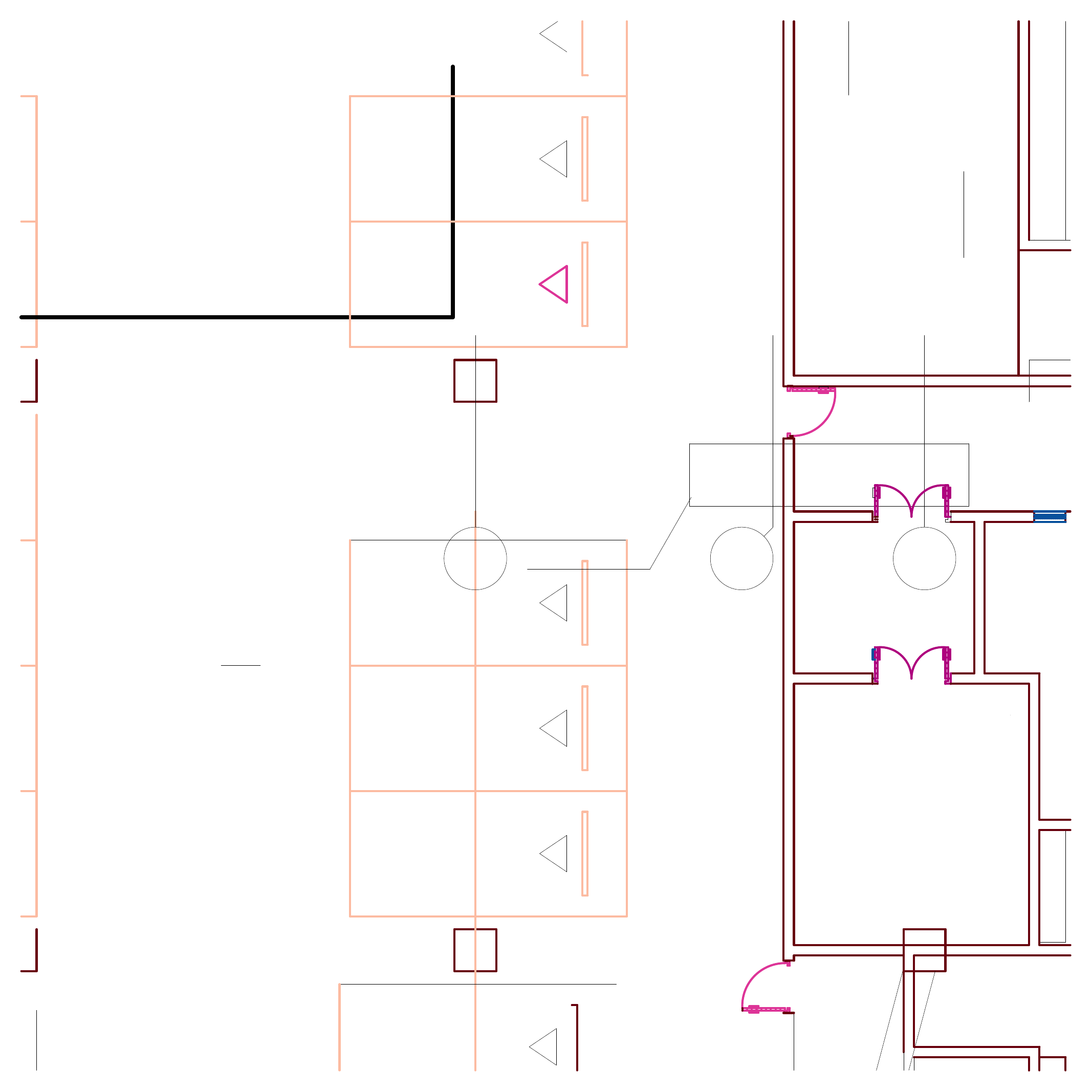} \\ \hline
    \includegraphics[width=0.3\textwidth]{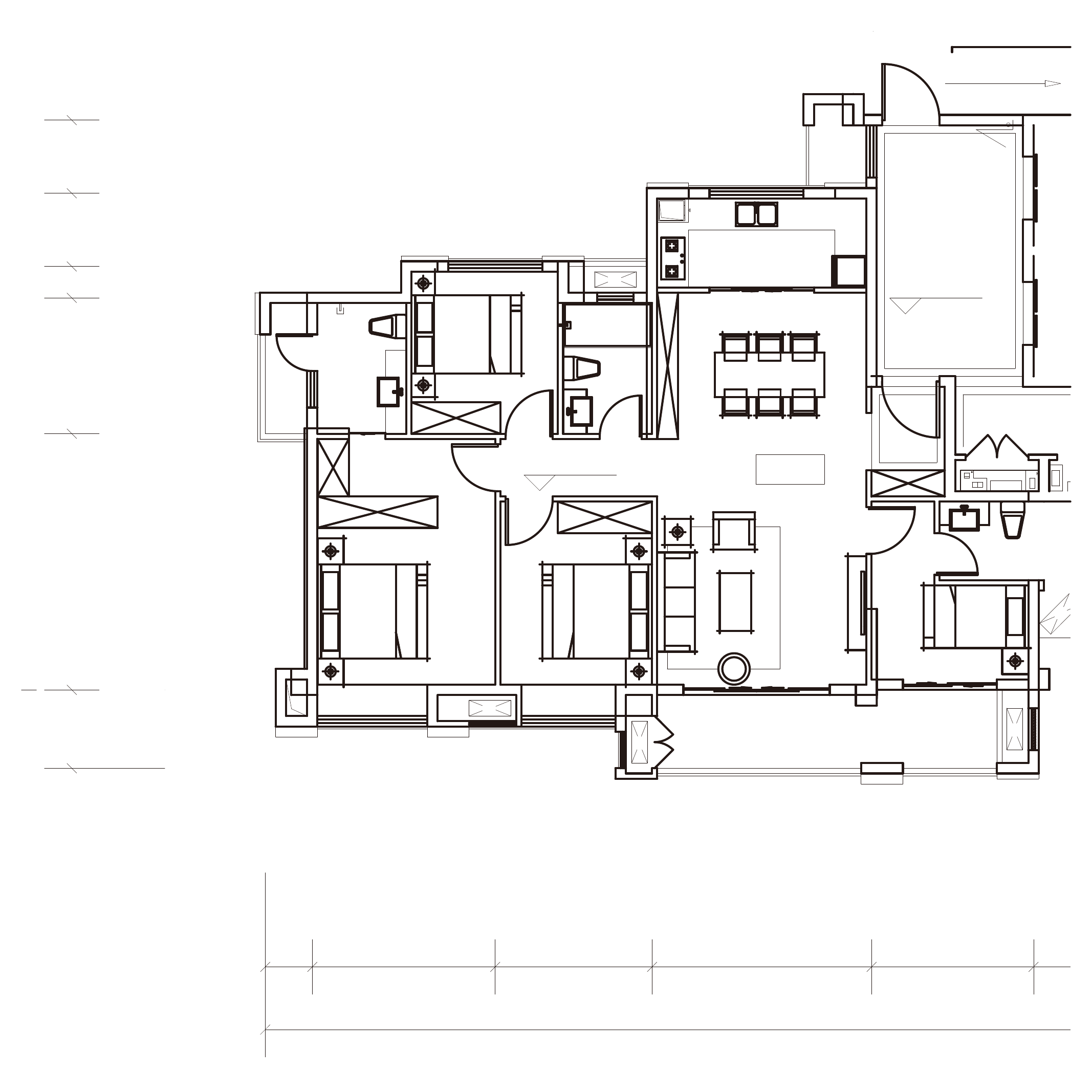} & \includegraphics[width=0.3\textwidth]{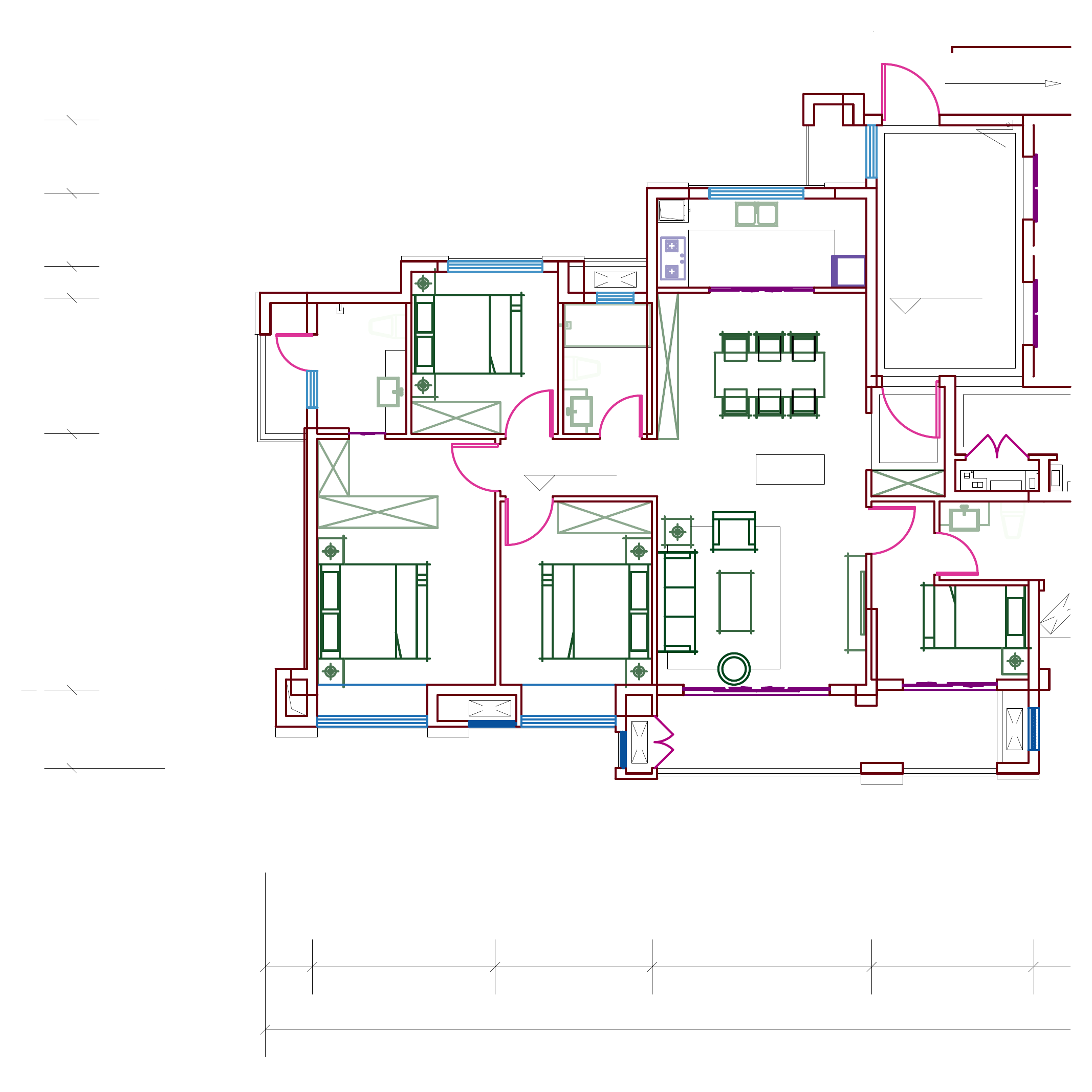} & \includegraphics[width=0.3\textwidth]{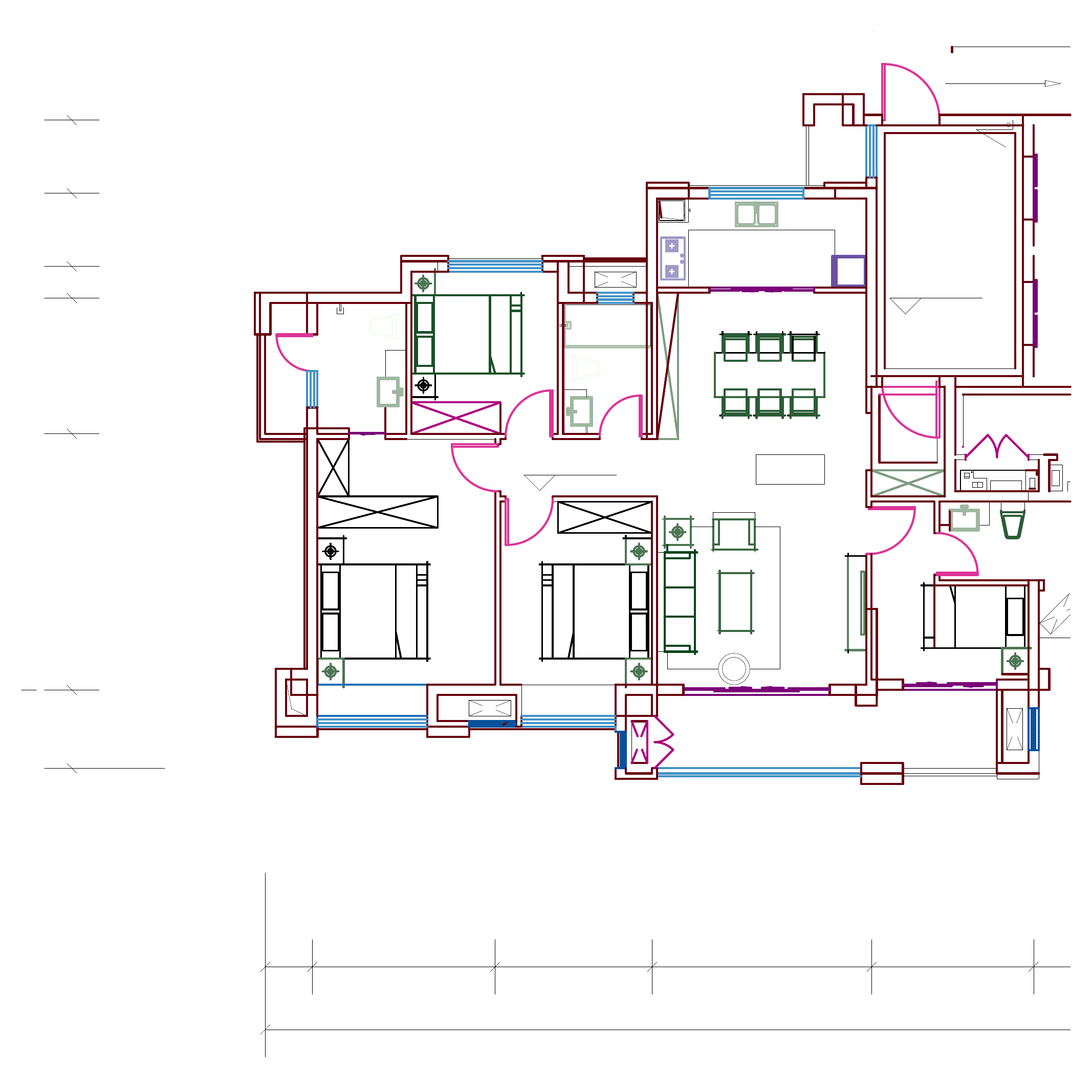} \\ \hline
    \includegraphics[width=0.3\textwidth]{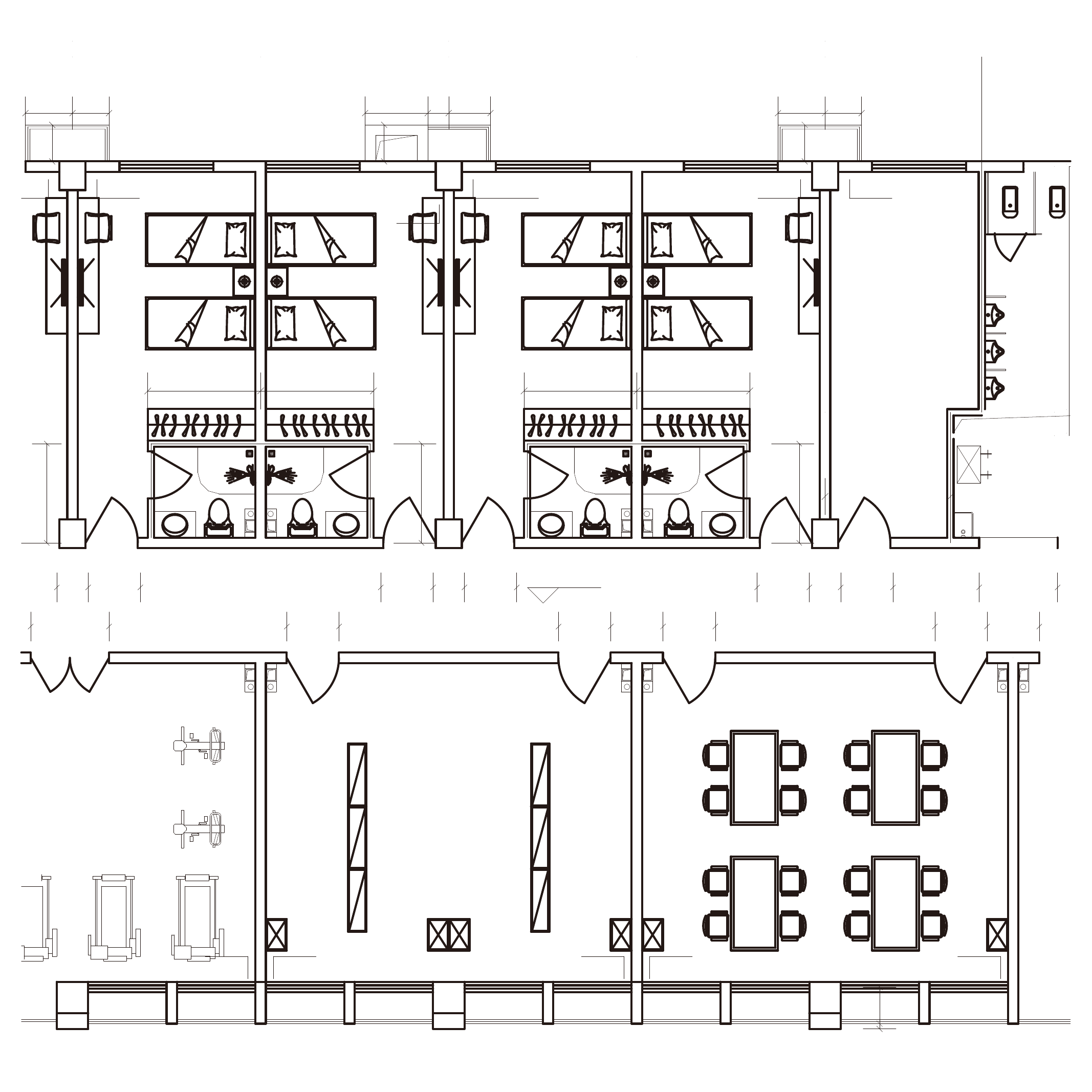} & \includegraphics[width=0.3\textwidth]{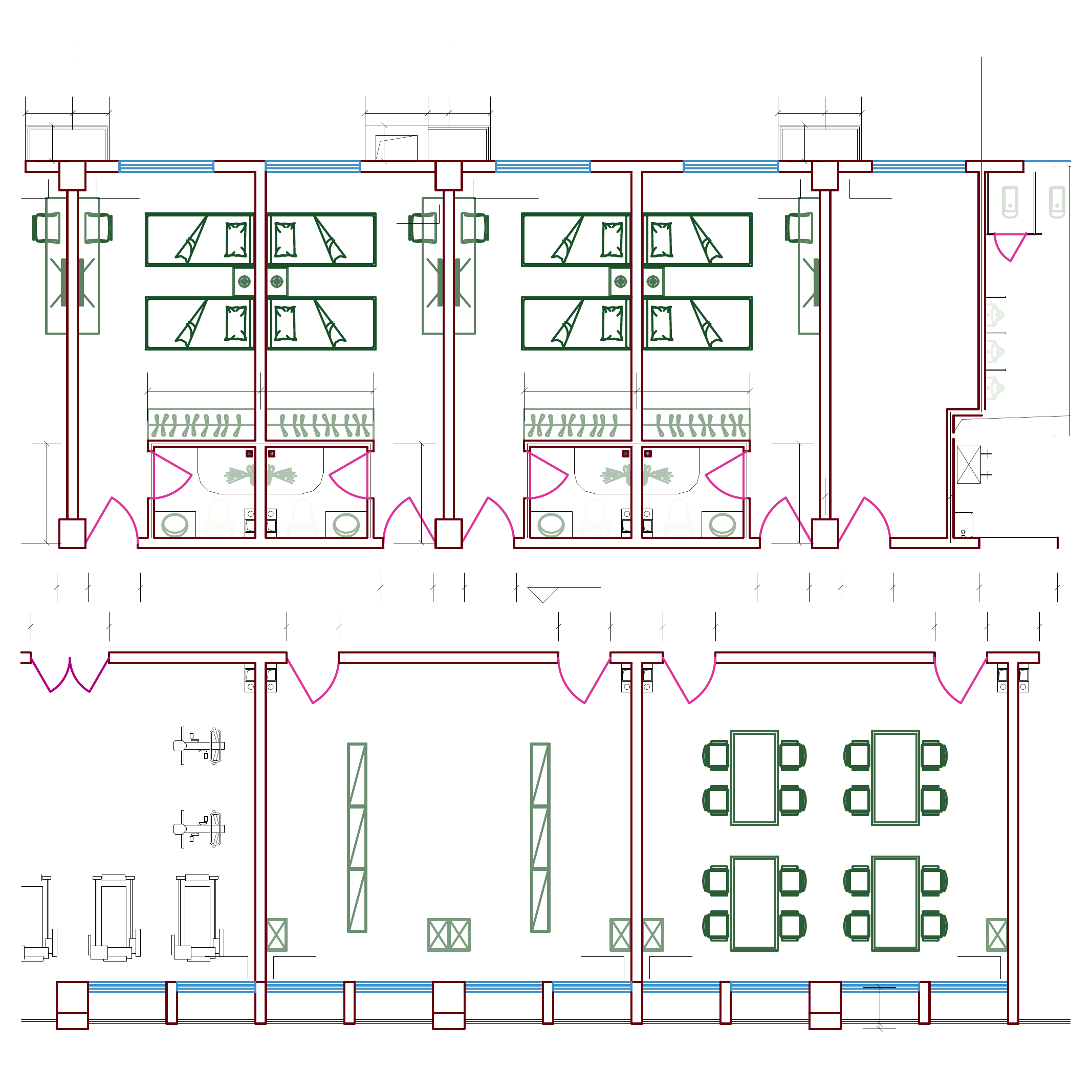} & \includegraphics[width=0.3\textwidth]{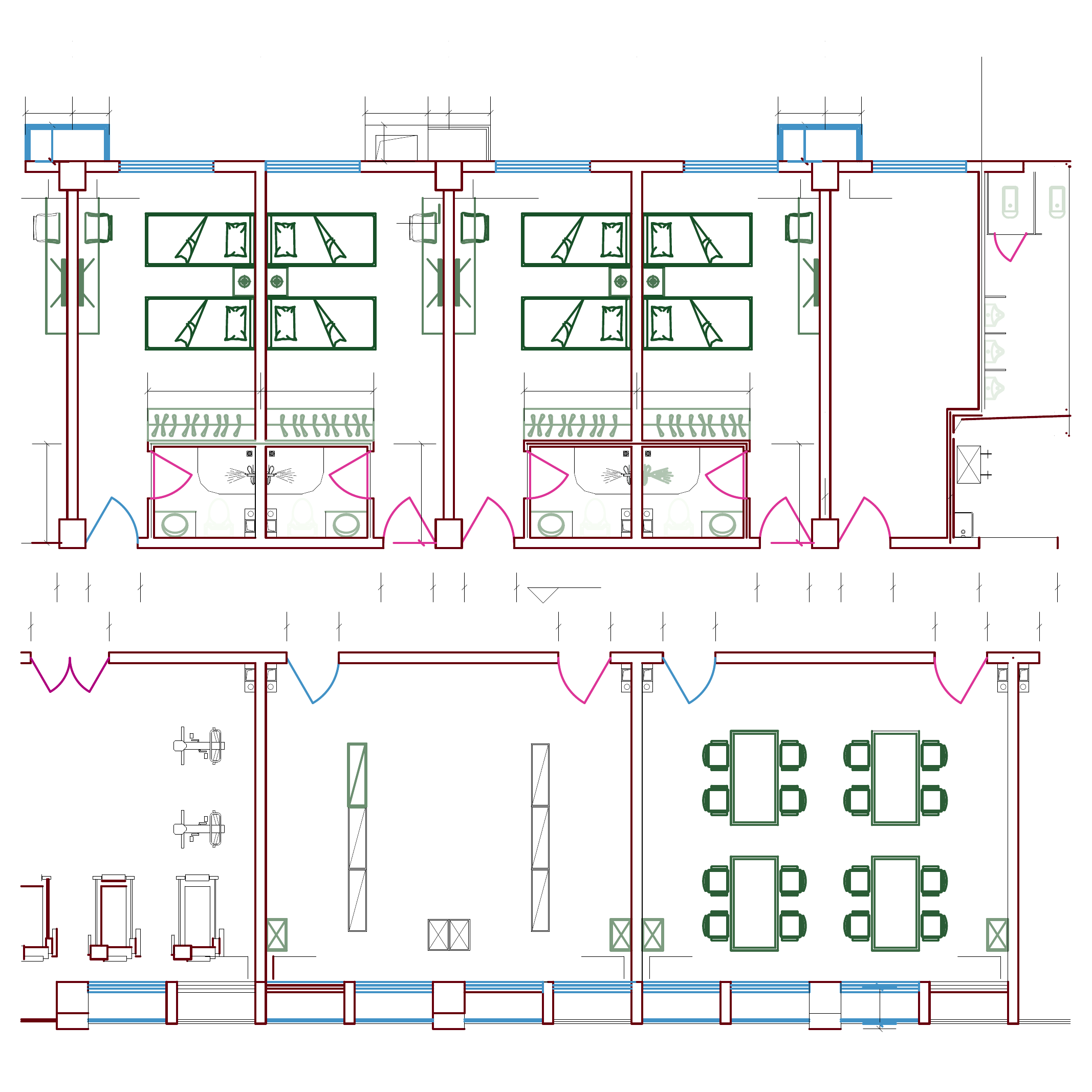} \\ \hline
    \includegraphics[width=0.3\textwidth]{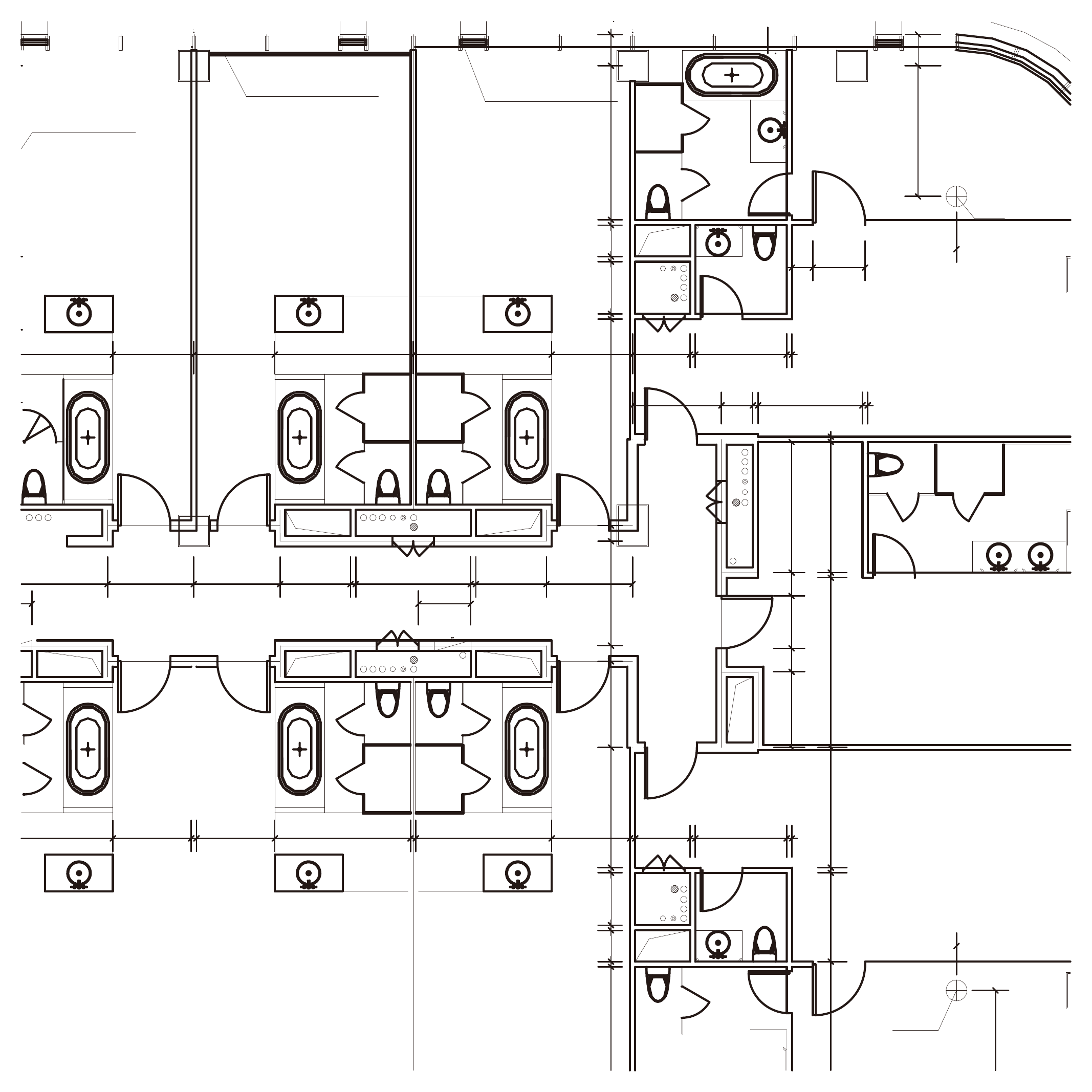} & \includegraphics[width=0.3\textwidth]{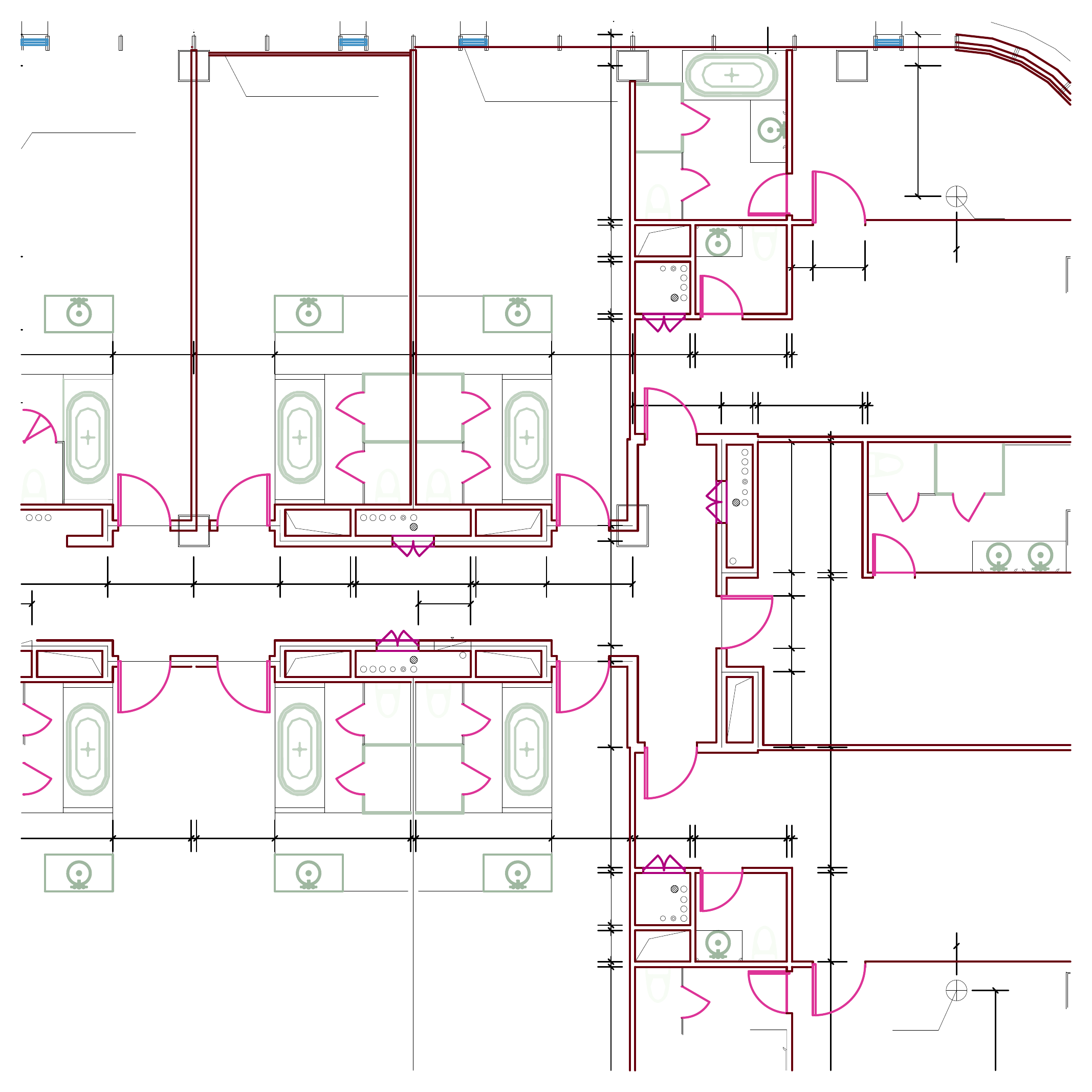} & \includegraphics[width=0.3\textwidth]{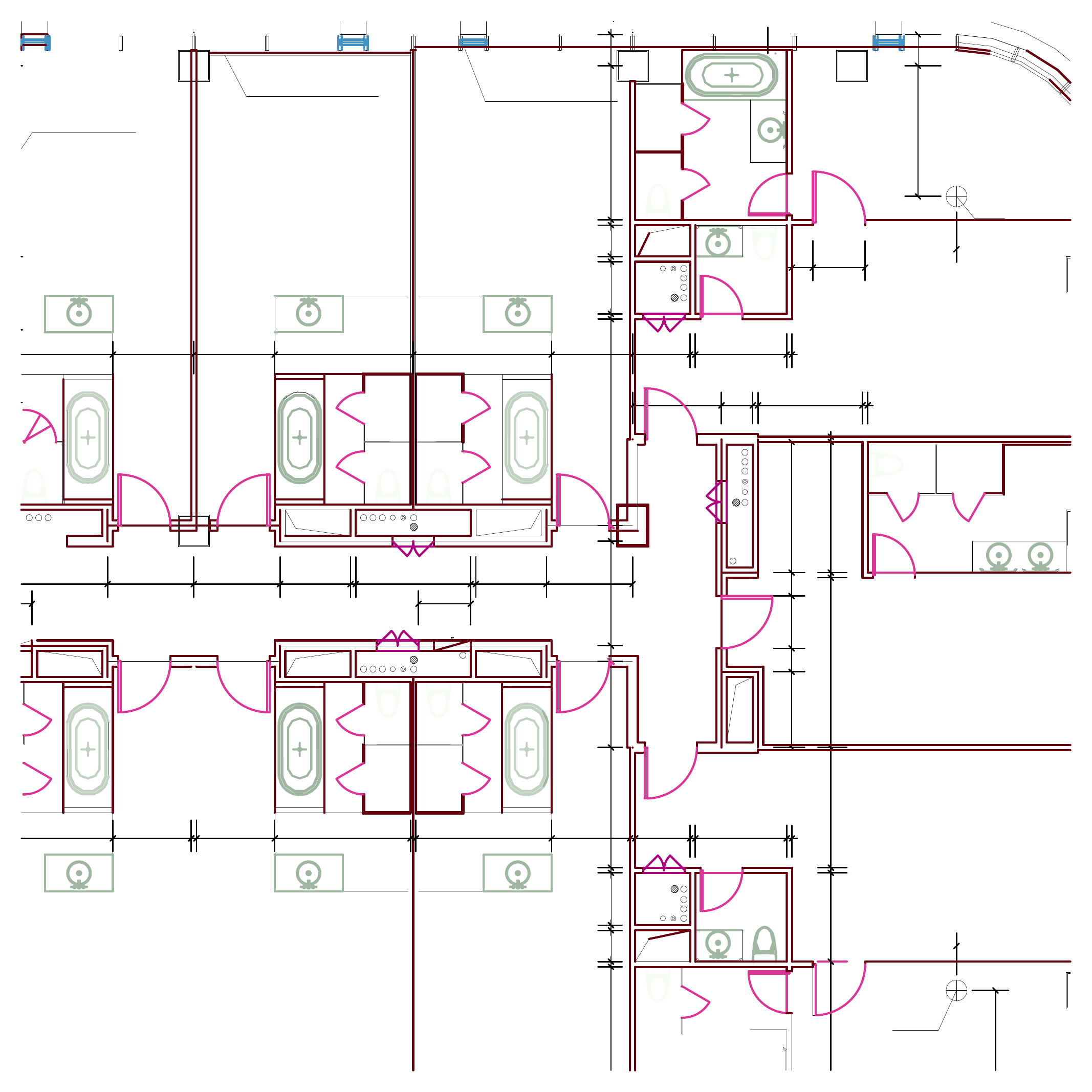} \\
    Raw Input & GT & Panoptic Prediction \\
\end{tabular}
\caption{Results of PanCADNet on FloorPlanCAD, see the main manuscript for annotation details. The images are part of our \textit{test} set of \textit{underground parking lot}, \textit{residential building} and \textit{hotel} CAD drawings. }\label{fig:pan_results_3}
\end{figure*}

In the fifth column of Table~\ref{tab:per_class}, we provide the detailed evaluation results of panoptic quality(PQ), segmentation quality(SQ) and recognition quality(RQ). Additional visualization results of PanCADNet on FloorPlanCAD dataset are shown in Figure~\ref{fig:pan_results_1},  Figure~\ref{fig:pan_results_2} and Figure~\ref{fig:pan_results_3}. The results for 2 stuff classes are obtained by GCN head while 28 thing classes are obtained by detection head~\cite{ren2016faster}.


\section{Limitations and Future Works}
The proposed PanCADNet benefits from the GCN head which takes the vectorized data as input, utilizing both the geometric feature and aligned CNN features, aggregating neighbour information by graph topology, resulting in a good results for the two key stuff classes (i.e. \textit{wall} and \textit{parking}). With the help of predicted bounding box using a detection head, we can distinguish each instance in thing classes easily. 

\begin{figure*}
\centering
\begin{tabular}{c| c| c}
    \includegraphics[width=0.3\textwidth]{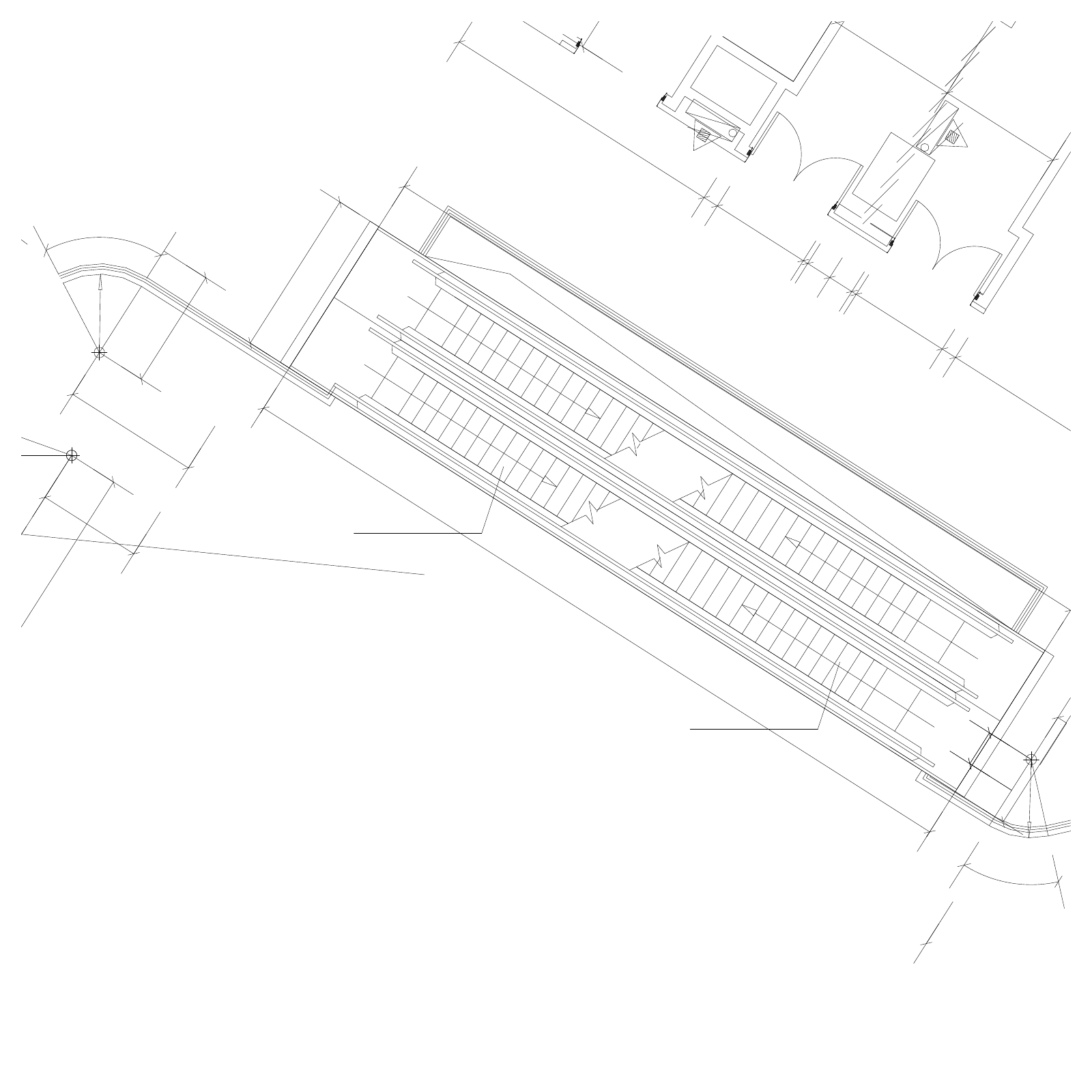} & \includegraphics[width=0.3\textwidth]{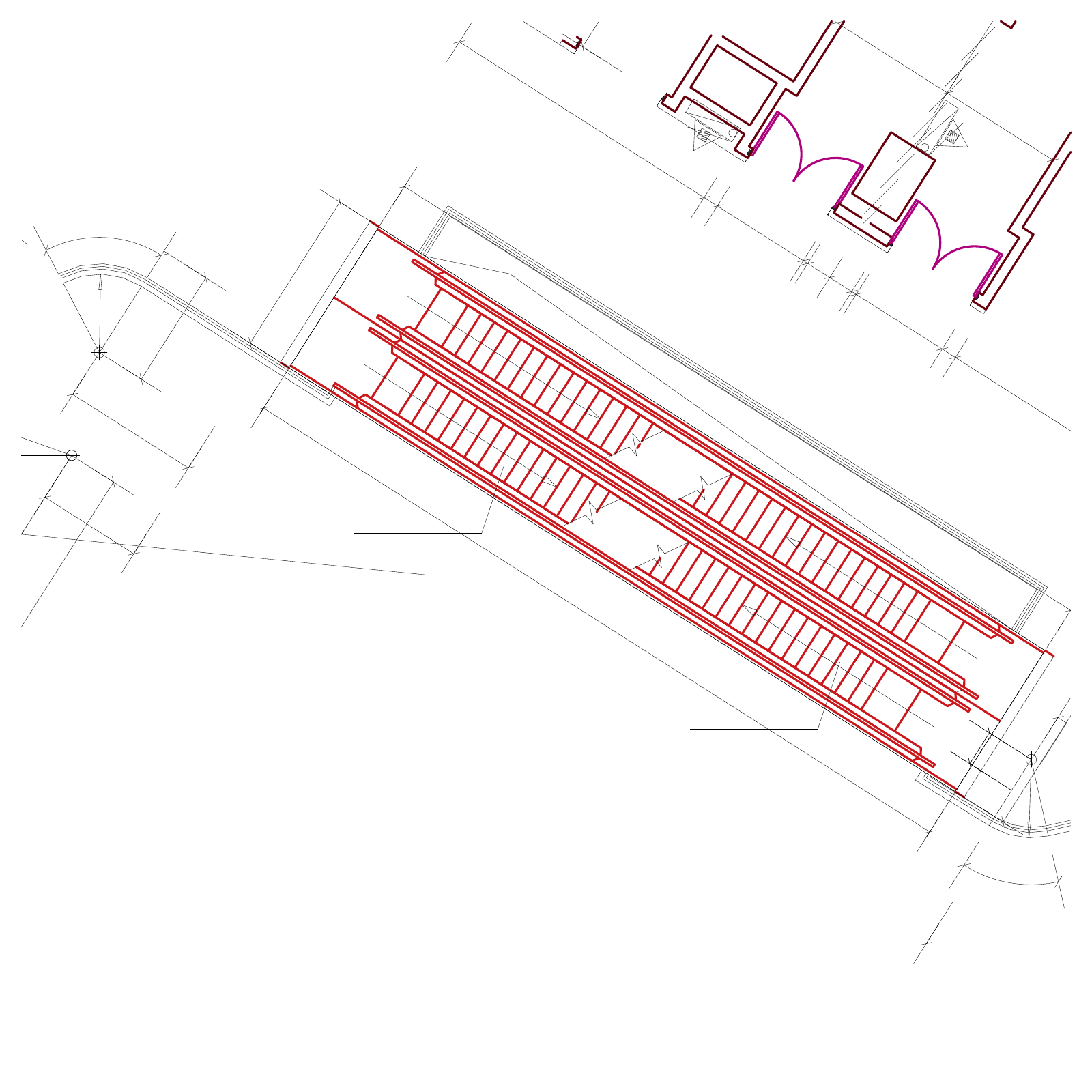} & \includegraphics[width=0.3\textwidth]{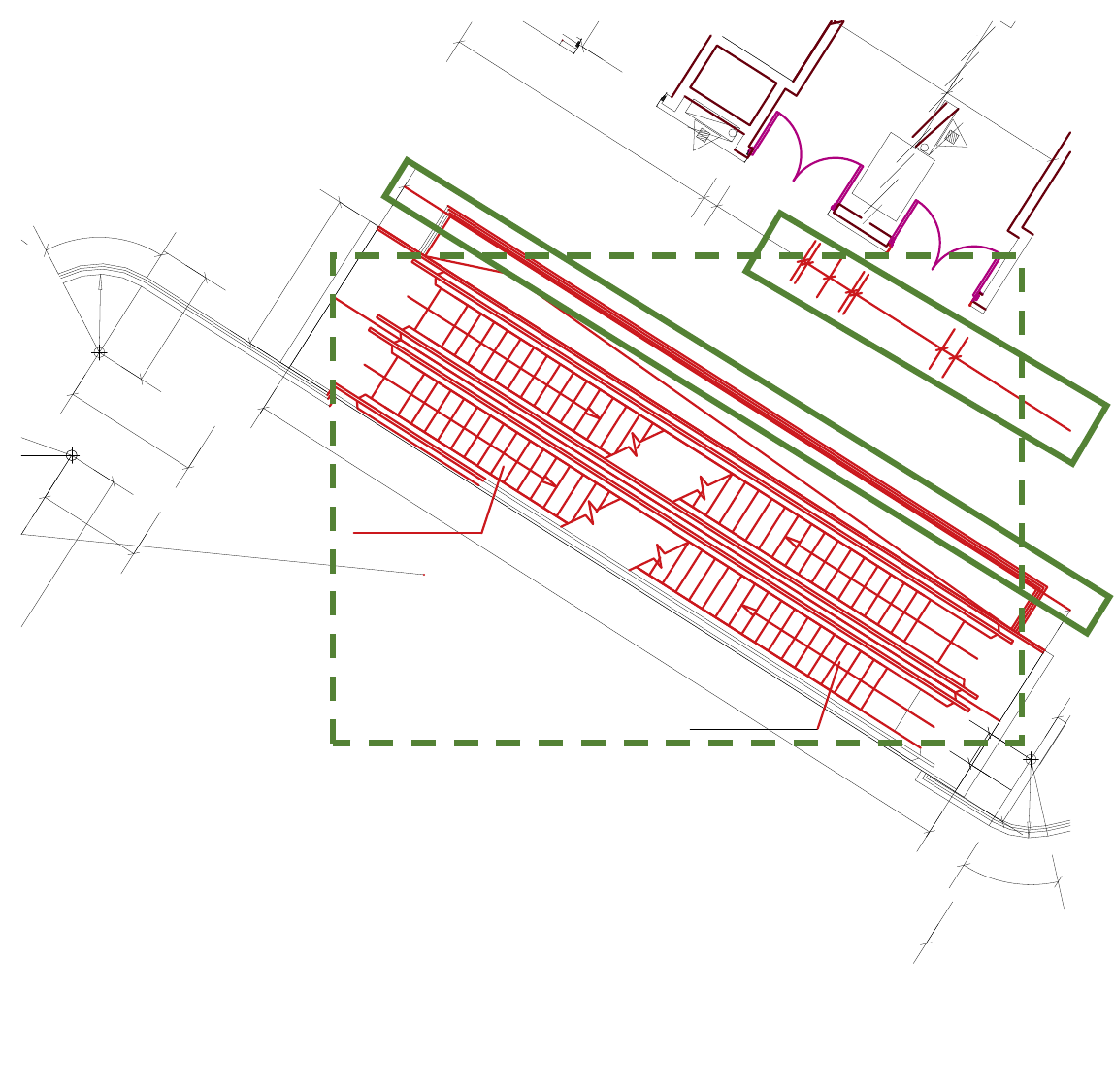} \\
    Raw Input & GT & Panoptic Prediction \\
\end{tabular}
\caption{A failure case of PanCADNet, where the predicted box of the stairs is shown using a dotted green box, the entities mis-classified by the predicted box are highlighted by solid green boxes.}\label{fig:failed_case}
\end{figure*}

Although the proposed method can solve the panoptic symbol spotting problem, some limitations still exist: as is pointed out in Figure~\ref{fig:failed_case}, some background elements may be mis-classified by the predicted box. 
The future works include generating instance proposals in vector space which can propose object instances in a more flexible way.

\clearpage 

    